\documentclass[10pt]{article}
\usepackage[paperwidth=210mm,paperheight=280mm,textwidth=176mm,textheight=236mm,centering,footskip=12mm]{geometry}
\usepackage[T1]{fontenc}
\usepackage{lmodern}
\usepackage{amsmath,amssymb,amsfonts,bm}
\usepackage{graphicx}
\usepackage[table]{xcolor}
\usepackage{booktabs,multirow,tabularx,array}
\usepackage{arydshln,makecell,adjustbox,nicefrac,placeins}
\usepackage{enumitem}
\usepackage{cite}
\usepackage[font=footnotesize,labelfont=bf,labelsep=quad]{caption}
\usepackage[hidelinks,pdfstartview=FitH,breaklinks]{hyperref}
\hypersetup{
  pdftitle={Deep learning reforms image matching: a survey and outlook},
  pdfauthor={Shihua Zhang; Zizhuo Li; Kaining Zhang; Yifan Lu; Yuxin Deng; Linfeng Tang; Xingyu Jiang; Hao Zhang; Jiayi Ma},
  pdfsubject={Image matching survey},
  pdfkeywords={image matching, deep learning, 3D vision, local features, feature matching, outlier filtering, pose estimation}
}
\renewcommand{\arraystretch}{1.3}
\setlist[itemize]{leftmargin=0pt,itemindent=2.1em,labelsep=0.5em,topsep=0pt,itemsep=0pt,parsep=0pt}
\makeatletter
\renewcommand\footnotesize{\@setfontsize\footnotesize{7}{8.5}}
\renewcommand\section{\@startsection{section}{1}{\z@}{-4.5ex \@plus -1ex \@minus -.2ex}{2.5ex \@plus .1ex}{\normalfont\fontsize{12}{14}\selectfont\bfseries}}
\renewcommand\subsection{\@startsection{subsection}{2}{\z@}{-2ex \@plus -2ex \@minus -.1ex}{2ex \@plus .2ex}{\normalfont\normalsize\bfseries}}
\renewcommand\subsubsection{\@startsection{subsubsection}{3}{\z@}{-3ex \@plus -1ex \@minus -.1ex}{1ex \@plus .1ex}{\normalfont\normalsize\itshape}}
\renewcommand\@biblabel[1]{#1~}

\makeatother
\providecommand{\mathbbm}[1]{\mathbb{#1}}

\definecolor{tabhl}{HTML}{FFF1E6}
\definecolor{tabfirst}{HTML}{ECBB97}
\definecolor{tabsecond}{HTML}{F5DAC5}
\definecolor{tabthird}{HTML}{FBF0E6}
\newcommand{\tabfirst}[1]{\cellcolor{tabfirst}\textbf{#1}}
\newcommand{\tabsecond}[1]{\cellcolor{tabsecond}#1}
\newcommand{\tabthird}[1]{\cellcolor{tabthird}#1}
\DeclareRobustCommand{\capfirst}[1]{{\setlength{\fboxsep}{1.2pt}\colorbox{tabfirst}{\textbf{#1}}}}
\DeclareRobustCommand{\capsecond}[1]{{\setlength{\fboxsep}{1.2pt}\colorbox{tabsecond}{#1}}}
\DeclareRobustCommand{\capthird}[1]{{\setlength{\fboxsep}{1.2pt}\colorbox{tabthird}{#1}}}
\begin{document}
\begin{center}
{\fontsize{16}{20}\selectfont\bfseries Deep learning reforms image matching:\\[2pt] a survey and outlook\par}
\vspace{4mm}
{\fontsize{11}{14}\selectfont
Shihua Zhang\textsuperscript{1,$\dagger$},
Zizhuo Li\textsuperscript{1,$\dagger$},
Kaining Zhang\textsuperscript{2},
Yifan Lu\textsuperscript{1},
Yuxin Deng\textsuperscript{1},\\
Linfeng Tang\textsuperscript{1},
Xingyu Jiang\textsuperscript{3},
Hao Zhang\textsuperscript{1},
Jiayi Ma\textsuperscript{1,*}\par}
\vspace{2mm}
{\fontsize{8}{10}\selectfont
\textsuperscript{1}Electronic Information School, Wuhan University, Wuhan 430072, China\\
\textsuperscript{2}School of Artificial Intelligence and Robotics, Hunan University, Changsha 410082, China\\
\textsuperscript{3}School of Artificial Intelligence and Automation, Huazhong University of Science and Technology, Wuhan 430074, China\par}
\vspace{2mm}
{\fontsize{8}{10}\selectfont
\textsuperscript{$\dagger$}These authors contributed equally to this work.\quad
\textsuperscript{*}Corresponding author: \href{mailto:jyma2010@gmail.com}{jyma2010@gmail.com}\par}
\end{center}
\vspace{1mm}
\begin{center}\textbf{Abstract}\end{center}
\begin{quote}\small
Image matching, which establishes correspondences between two images to recover 3D structure and camera geometry, is a cornerstone of computer vision and underpins a wide range of applications, including visual localization, 3D reconstruction, and simultaneous localization and mapping (SLAM). Traditional pipelines, composed of a detector-descriptor, a feature matcher, an outlier filter, and a geometric estimator, falter in challenging scenarios. Recent advances in deep learning have substantially improved both their robustness and accuracy. This survey reviews how deep learning has progressively transformed the classical image matching pipeline. Our taxonomy is aligned with the traditional pipeline and covers two aspects: i) replacing individual steps with learnable alternatives, including learnable detector-descriptors, outlier filters, and geometric estimators; and ii) merging multiple steps into end-to-end learnable modules, including middle-end sparse matchers, end-to-end semi-dense/dense matchers, and pose regressors. We first examine the design principles, advantages, and limitations of both aspects, and then benchmark representative methods on relative pose estimation, homography estimation, matching accuracy assessment, visual localization, and 3D reconstruction. Finally, we discuss open challenges and directions for future research. By systematically categorizing and evaluating learning-based methods, this survey offers a clear overview of how image matching is evolving and where further progress is needed. The project repository is available at \url{https://github.com/ZizhuoLi/awesome-image-matching-survey}.

\end{quote}

\section{Introduction}
Computer vision, which processes, analyzes, and interprets images captured by sensors such as cameras, is one of the main ways in which artificial intelligence perceives the environment. Image matching, which recovers the 3D relationships between 2D images, is a fundamental building block of many computer vision systems and enables robots to perceive the world comprehensively. It identifies the same textures or regions, typically represented as keypoints, across images taken from different viewpoints, and establishes correspondences (matches) between them to recover 3D structure and estimate the relative poses of the involved views and objects. Image matching underpins a wide range of applications, including image retrieval~\cite{datta2008image}, visual localization~\cite{sattler2018benchmarking}, 3D reconstruction~\cite{hartley2003multiple}, Structure from Motion (SfM)~\cite{pan2024global}, Simultaneous Localization and Mapping (SLAM)~\cite{engel2014lsd}, and novel view synthesis~\cite{kerbl20233d}.

\begin{figure}[!t]
	\centering
	\includegraphics[width=0.95\textwidth]{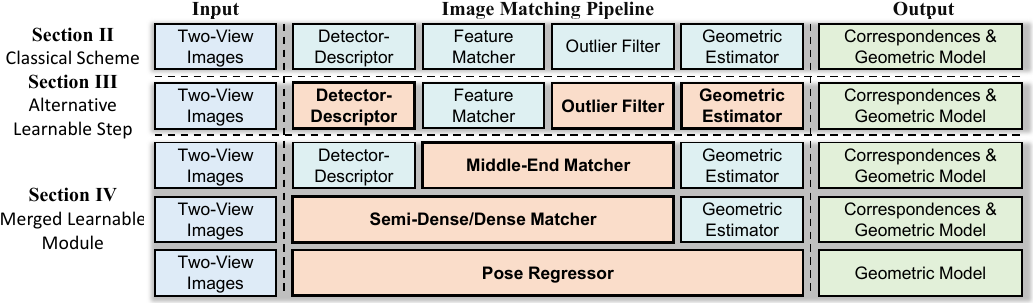}
	\caption{
		Taxonomy of image feature matching. The orange boxes mark the focus of this paper.
	}
	\label{fig:introduction}
\end{figure}

Research on image matching dates back to early studies on pattern recognition and human vision~\cite{marr1976cooperative}, which inspired template matching~\cite{brice1970scene} and cross-correlation~\cite{gruen1985adaptive}. The concept of ``interest points''~\cite{moravec1981rover} was later introduced to define distinctive feature points (keypoints). Matching such keypoints with local invariant descriptors~\cite{schmid1995matching} led to the standard feature-based matching scheme, which consists of a detector-descriptor, a feature matcher, an outlier filter, and a geometric estimator, and outputs both correspondences and a geometric model. This pipeline is illustrated in Figure~\ref{fig:introduction}(II) and briefly reviewed in Section~\ref{sec:conventional}.
While effective under mild conditions, it often fails under extreme illumination changes, large viewpoint changes, weak textures, repetitive patterns, or occlusions.

\begin{figure}[!t]
	\centering
	\includegraphics[width=\textwidth]{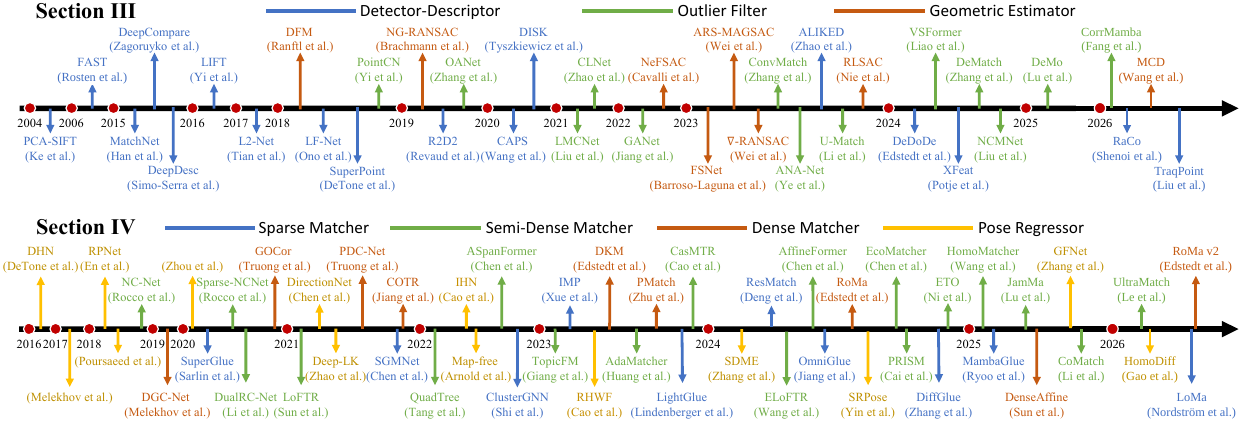}
	\caption{
		Timelines of representative methods for alternative learnable steps (Section~\ref{sec:alternative}) and merged learnable modules (Section~\ref{sec:merged}).
	}
	\label{fig:timeline}
\end{figure}

Learning-based approaches have since been developed to improve the robustness and accuracy of the classical pipeline.
A straightforward strategy replaces individual modules with learnable counterparts, namely learnable detector-descriptors for better feature representation, outlier filters for reliable matching under challenging conditions, and geometric estimators for robust model estimation (Figure~\ref{fig:introduction}(III)), while matching itself still relies on feature similarity. Another strategy merges consecutive stages into a unified module, which gives rise to the three representative paradigms shown in Figure~\ref{fig:introduction}(IV). Middle-end sparse matchers combine the feature matcher and the outlier filter, and establish correspondences directly in a learned feature space. Semi-dense/dense matchers further integrate the detector-descriptor into an end-to-end framework, avoiding the inconsistency between off-the-shelf detector-descriptors and the later stages. Pose regressors bypass explicit correspondences and directly regress the two-view transformation without iterative model fitting.
These two strategies are discussed in Sections~\ref{sec:alternative} and~\ref{sec:merged}, respectively.
Figure~\ref{fig:timeline} shows representative methods of each category over time.

This paper reviews how machine learning and deep learning techniques have progressively replaced components of the classical image matching pipeline, traces the evolution of individual modules and merged frameworks, and compares their strengths and weaknesses through extensive experiments on multiple tasks. Previous surveys in this field have mainly focused on specific stages of the pipeline. Some early reviews concentrate on the detector-descriptor stage, covering handcrafted~\cite{aanaes2012interesting,heinly2012comparative,awrangjeb2012performance} and learned methods~\cite{balntas2017hpatches,schonberger2017comparative}. Zitov\'a and Flusser~\cite{zitova2003image} give a broader overview of the entire registration pipeline, but their work predates learning-based approaches. Ma \emph{et al.}~\cite{ma2021image} are among the first to survey both handcrafted and learnable techniques along the full pipeline, but most merged modules appeared after their survey. More recent reviews~\cite{xu2024local,liao2024local} divide methods into detector-based and detector-free ones. This division does not show which pipeline stages a method replaces or merges, and these reviews give little or no coverage to learnable geometric estimators, pose regressors, and most learned outlier filters. Two doctoral theses also provide in-depth overviews of wide multiple baseline stereo~\cite{mishkin2021thesis} and of learned matching for 3D reconstruction~\cite{edstedt2025thesis}. Table~\ref{tab:survey_compare} summarizes how this survey differs from these works. %
In contrast, this work focuses on learning-based methods. It i) introduces a pipeline-aligned taxonomy that covers both alternative learnable steps and merged learnable modules (Figure~\ref{fig:introduction}), and ii) conducts unified experiments on relative pose estimation~\cite{yi2018learning}, homography estimation~\cite{detone2018superpoint}, matching accuracy assessment~\cite{truong2023pdc}, visual localization~\cite{sarlin2019coarse}, and 3D reconstruction~\cite{lindenberger2021pixsfm} for consistent comparisons across categories. Given the breadth of the field, we focus on methods that introduce a distinct technical idea or have influenced subsequent work. 

\begin{table}[!t]
	\centering
	\caption{Comparison with representative image matching surveys and evaluation studies.}
	\label{tab:survey_compare}
	\footnotesize
	\setlength{\aboverulesep}{0pt}
	\setlength{\belowrulesep}{0pt}
	\renewcommand{\arraystretch}{1.25}
	\begin{tabularx}{\textwidth}{>{\raggedright\arraybackslash}p{0.17\linewidth}>{\raggedright\arraybackslash}p{0.19\linewidth}>{\raggedright\arraybackslash}p{0.27\linewidth}>{\raggedright\arraybackslash}X}
		\toprule
		\textbf{Survey} & \textbf{Main scope} & \textbf{Pipeline coverage} & \textbf{Distinction from this survey} \\
		\midrule
		Balntas \emph{et al.}~\cite{balntas2017hpatches} and Sch\"onberger \emph{et al.}~\cite{schonberger2017comparative} & Handcrafted and learned local descriptors & Descriptor evaluation on patch-level tasks and in image-based 3D reconstruction & Evaluate early learned descriptors, mostly on handcrafted keypoints, before learned matchers emerged. \\
		Ma \emph{et al.}~\cite{ma2021image} & Handcrafted and learning-based image matching, including graph matching and point set registration & Detection, description, and matching with mismatch removal & Covers methods up to 2020, before most learned sparse, semi-dense, and dense matchers were proposed. \\
		Xu \emph{et al.}~\cite{xu2024local} & Deep local feature matching & Detector-based methods, including graph-based sparse matchers, and detector-free methods & Does not review outlier filters, learnable estimators, or pose regressors; results are collected from the original papers; methods up to 2023. \\
		Liao \emph{et al.}~\cite{liao2024local} & Local feature matching & Detector-based and detector-free methods & Groups methods by detector usage rather than by pipeline stage; methods up to 2023. \\
		\rowcolor{tabhl}
		\textbf{Ours} & Learning-based image matching & Learnable detector-descriptors, outlier filters, and geometric estimators; sparse, semi-dense, and dense matchers; pose regressors & Pipeline-aligned taxonomy covering methods up to 2026, with unified re-evaluation on five tasks. \\
		\bottomrule
	\end{tabularx}
\end{table}

We summarize our contributions as follows:
\begin{itemize}
	\item We present a comprehensive survey of learning-based image matching.\footnote{Project repository: \url{https://github.com/ZizhuoLi/awesome-image-matching-survey}.} Our taxonomy is aligned with the classical pipeline and shows how individual components are progressively replaced by learnable alternatives and how multiple stages are merged into unified modules.
	\item We analyze the key challenges of both alternative learnable steps and merged modules, discuss representative solutions, and trace the methodological evolution within each category.
	\item We conduct extensive experiments on multiple tasks to evaluate representative methods and identify unresolved issues of current learning-based methods.
	\item We discuss open problems and directions for future research.
\end{itemize}

\section{Classical Image Matching Scheme}
\label{sec:conventional}
The classical scheme illustrated in Figure~\ref{fig:introduction}(II) begins by detecting and describing keypoints in both images. In its early form~\cite{schmid1995matching}, interest points are matched with multi-scale, rotation-invariant local grayvalue invariants, with the aim of providing enough correct matches for subsequent robust geometric estimation. The popularity of this scheme is closely tied to 3D reconstruction from unstructured image collections~\cite{schonberger2016structure}. Since $N$ images yield $O(N^2)$ candidate pairs, it is much cheaper to extract keypoints and descriptors once per image and reuse them for every pair than to process each pair jointly. The detector identifies the spatial coordinates of keypoints, often together with their scale and orientation, while the descriptor encodes the local appearance around each keypoint. Handcrafted methods exploit image intensity and structural patterns to identify informative regions, including blob detectors~\cite{lowe2004distinctive}, corner detectors~\cite{harris1988combined,rosten2006machine}, and region detectors~\cite{matas2004robust,mikolajczyk2005comparison}. Among them, SIFT~\cite{lowe1999object,lowe2004distinctive} is one of the most widely used detector-descriptors. It detects keypoints as extrema in a difference-of-Gaussians (DoG) pyramid, assigns each keypoint a dominant orientation, and describes its neighborhood with histograms of local gradient orientations.
ORB~\cite{rublee2011orb} is widely used in industrial applications because of its efficiency. It detects keypoints with the FAST segment test~\cite{rosten2006machine}, ranks them by the Harris corner response~\cite{harris1988combined}, and describes them with an orientation-aware variant of BRIEF~\cite{calonder2010brief}. Later handcrafted binary descriptors further improve robustness to illumination changes by encoding local inter-patch invariance~\cite{lin2023illumination}.

The feature matcher then establishes correspondences between the descriptors of the two images. The most common strategy is nearest neighbor (NN) matching, which finds for each descriptor the most similar one in the other image under a distance such as the Euclidean distance. Mutual nearest neighbor (MNN) matching retains only reciprocal best matches. Since it removes NN matches that are not reciprocal, MNN can also be viewed as a simple outlier filter.
This process yields a set of putative correspondences. %

Due to the limited discriminability of descriptors, the putative set often contains many false matches (outliers), especially in challenging scenes.
Outlier filtering aims to extract the correct matches (inliers) from this set. For instance, the ratio test~\cite{lowe2004distinctive} discards ambiguous matches whose closest/second-closest distance ratio exceeds a threshold.
Other methods exploit the local consensus of inliers~\cite{ma2014robust,bian2017gms}. For example, VFC~\cite{ma2014robust} models the motion field with a function in a reproducing kernel Hilbert space and imposes motion smoothness through regularization.

Finally, the geometric estimator fits a transformation model to the correspondences. It often solves a set of linear equations, where the Direct Linear Transformation (DLT) with least squares gives an initial estimate and reweighted least squares further improves robustness~\cite{torr1997development}.
RANdom SAmple Consensus (RANSAC)~\cite{fischler1981random} provides a more reliable estimate in the presence of outliers. It generates model hypotheses from random minimal subsets, scores each hypothesis by its inlier count, and selects the highest-scoring one. Numerous variants have since been proposed to improve its speed and accuracy~\cite{raguram2012usac}. MSAC and MLESAC~\cite{torr2000mlesac} score hypotheses with truncated or likelihood-based residuals instead of the inlier count, LO-RANSAC~\cite{chum2003locally} and GC-RANSAC~\cite{barath2018graph} locally optimize promising hypotheses, and MAGSAC++~\cite{barath2020magsac++} marginalizes over the noise scale to reduce the dependence on a manually set inlier threshold.

The classical pipeline remains practical and effective, but its handcrafted components have limited representational capacity. Researchers have therefore turned to learning-based techniques, either replacing individual stages with learnable alternatives or merging several steps into end-to-end modules. The following sections review both directions, including their design principles, representative methods, and effect on the overall matching process.

\section{Alternative Learnable Step}
\label{sec:alternative}
Deep neural networks have reshaped the classical pipeline in two ways. This section covers the first, which replaces individual steps with learnable alternatives, namely learnable detector-descriptors (Section~\ref{subsec:learn_det_desc}), outlier filters (Section~\ref{subsec:learn_filter}), and geometric estimators (Section~\ref{subsec:learn_estimator}).

\subsection{Learnable Detector-Descriptor}
\label{subsec:learn_det_desc}
\begin{figure}[!t]
	\centering
	\includegraphics[width=0.66\textwidth]{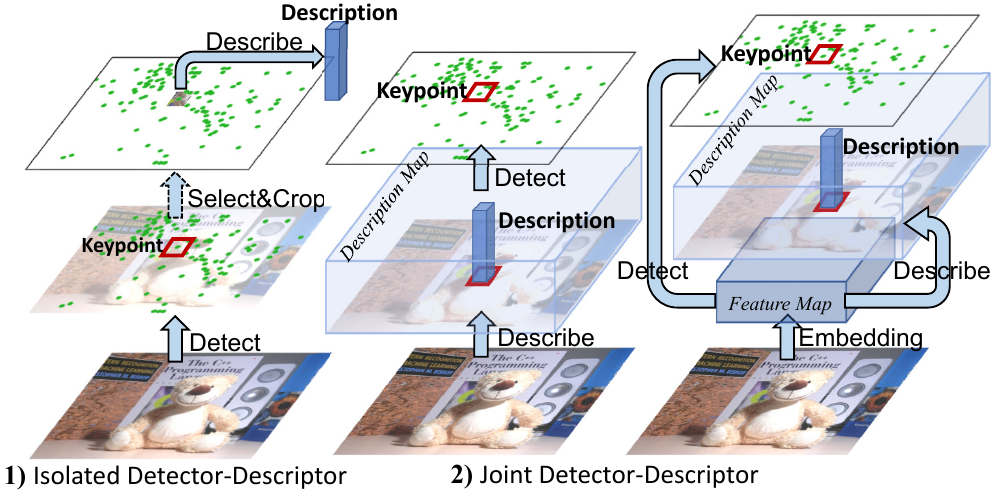}
	\caption{
		Frameworks of different learnable detector-descriptors.
	}
	\label{fig:detector_descriptor}
\end{figure}

Stepwise image matching relies on the detector-descriptor to provide keypoints as matching primitives. Because of its fundamental role in sparse matching, this stage was among the first to be replaced by learnable alternatives.
Figure~\ref{fig:detector_descriptor} illustrates representative frameworks, which are detailed below.

\subsubsection{Isolated Detector-Descriptor}
\label{subsubsec:isolate_det_desc}
Early learnable detectors and descriptors are developed in isolation, following the separate stages of traditional handcrafted pipelines~\cite{zitova2003image,ma2021image}.

\noindent \textbf{Learnable Detector.}
The primary concern of a detector is how to define keypoints and their locations in a 2D image, which determines the \underline{\emph{reliability}} of the detector. For example, keypoints should not lie on transient structures or in noise-prone regions. In addition, keypoints corresponding to the same physical structures or 3D locations should be detected consistently across views, a property known as \underline{\emph{repeatability}}. These two attributes have remained central to the development of learning-based detectors.

The earliest learnable detectors rely on classical machine learning.
Some of them aim at reliable corner detection.
Similar to the fast corner detector of Trajkovic and Hedley~\cite{trajkovic1998fast}, which relies on direct gray-value comparisons, FAST~\cite{rosten2006machine} compares the pixels on a circle around each candidate, but learns a decision tree from training images to accelerate this segment test. Kienzle \emph{et al.}~\cite{kienzle2006learning} instead learn an interest operator from human eye movements. For repeatability, FAST-ER~\cite{rosten2008faster} optimizes the FAST decision tree with simulated annealing, and Hartmann \emph{et al.}~\cite{hartmann2014predicting} learn a decision tree that selects matchable and robust keypoints for SfM. However, these early detectors are limited by shallow models and small training sets, and deep learning has since enabled more powerful detector learning.

Convolutional Neural Networks (CNNs) are widely used for learnable detection, as their local receptive fields suit the local nature of keypoints. An early work by Richardson and Olson~\cite{richardson2013learning} learns linear convolutional filters through random sampling and frequency-domain selection to minimize the pose error of stereo visual odometry. Later CNN-based detectors continue to emphasize reliability and repeatability.
Fully supervised methods use the outputs of off-the-shelf detectors~\cite{verdie2015tilde} or synthetic salient points~\cite{detone2017toward} as labels. TILDE~\cite{verdie2015tilde} is trained on images with extreme illumination changes for cross-condition repeatability, while MagicPoint~\cite{detone2017toward} augments data with homographic warps and heavy noise (brightness changes, shadows, blur, and Gaussian/speckle noise) to enhance robustness.
Although they predict good keypoints in familiar scenes, such fully supervised detectors rely on predefined keypoints and often perform poorly under unseen transformations or noise.
In a different line, Lenc \emph{et al.}~\cite{lenc2016learning} learn covariant detectors without manual labels by directly using the covariance constraint as the regression target. Building on this, Zhang \emph{et al.}~\cite{zhang2017learning} incorporate TILDE anchors to improve localization reliability, and Key.Net~\cite{barroso2019key,barroso2022key} combines handcrafted and learned filters for robustness across conditions. Quad-Networks~\cite{savinov2017quad} also avoid manual labels by learning to rank interest points consistently under image transformations. Instead of training a dedicated detector, ELF~\cite{benbihi2019elf} and D2D~\cite{tian2020d2d} derive keypoints from the feature maps or dense descriptors of pre-trained networks.
Later work improves detectors with more advanced designs. NeSS-ST~\cite{pakulev2023ness} adds a learnable scorer to select the most reliable Shi-Tomasi keypoints~\cite{shi1994good}. Rotation-equivariant networks~\cite{lee2022self} improve repeatability under rotation without relying on data augmentation, and MorseDet~\cite{barbaraniscale} obtains scale-free keypoints with differentiable persistent homology. DaD~\cite{edstedt2025dad} trains a descriptor-free detector with reinforcement learning and distills two complementary detectors, which respond to light and dark keypoints, respectively, into a single model. RaCo~\cite{shenoi2026raco} jointly learns a repeatable detector, a differentiable keypoint ranker, and a metric covariance estimator for keypoint uncertainty, and is trained on single images without covisible image pairs.

\noindent \textbf{Learnable Descriptor.}
Based on the detected keypoints, learnable descriptors assign each keypoint a representation (description) that distinguishes it from the others. Like detectors, descriptors face challenging conditions such as perspective and illumination changes, under which a reliable descriptor should produce stable descriptions for the same keypoint in different images. The \underline{\emph{discriminative}} power of descriptors and their \underline{\emph{invariance}} to image distortions and environmental changes are therefore key to their performance.

Among early machine-learning descriptors, PCA-SIFT~\cite{ke2004pca} applies Principal Component Analysis (PCA) to reduce a local gradient vector to a compact and robust description. In a similar vein, Cai \emph{et al.}~\cite{cai2010learning} use linear discriminant projection to improve discriminativeness while reducing dimensionality, and Brown \emph{et al.}~\cite{brown2010discriminative} optimize descriptor parameters via Linear Discriminant Analysis (LDA)~\cite{hua2007discriminant} and Powell minimization~\cite{winder2007learning}.
Invariance has also received attention. LDAHash~\cite{strecha2011ldahash} learns short binary strings in Hamming space from challenging data. Subsequent work introduces a boosted binary descriptor for faster description~\cite{trzcinski2013boosting} and a sparse spatial-pooling framework that uses L1 regularization to select pooling regions~\cite{simonyan2014learning}.

Siamese networks~\cite{bromley1993signature} then inspired the first deep learning descriptors. DeepCompare~\cite{zagoruyko2015learning} explores various architectures for comparing local patches, including Siamese, pseudo-Siamese, two-channel, and central-surround two-stream networks. MatchNet~\cite{han2015matchnet} jointly learns a feature network and a metric network with a cross-entropy loss over matching and non-matching patch pairs. Both methods rely on a learned metric network for match prediction, which places less emphasis on the discriminativeness of the descriptors themselves.
Building on Siamese CNNs, DeepDesc~\cite{simo2015discriminative} introduces a contrastive loss on L2-normalized features to pull matching pairs (positives) together and push non-matching pairs (negatives) apart, and employs hard negative mining to enhance discriminative power. Concurrently, TFeat~\cite{balntas2016learning}, PN-Net~\cite{balntas2016pnnet}, and TNet~\cite{kumar2016learning} adopt triplet losses that require the distance of a positive pair to be smaller than that of a negative pair. TNet further proposes a triplet Siamese network and adds a global loss that minimizes the overall classification error on the training set, improving invariance under challenging conditions.

L2-Net~\cite{tian2017l2} then adopts a considerably deeper CNN than earlier shallow patch networks such as TFeat and DeepDesc, which substantially improves descriptor quality. It is trained with progressive negative sampling, a compactness regularizer, and intermediate feature supervision. HardNet~\cite{mishchuk2017working} later shows that keeping the L2-Net architecture and mining the hardest negative within each batch achieves comparable or better performance without these auxiliary designs. Following this line, Zhang \emph{et al.}~\cite{zhang2017learning1} propose a global orthogonal regularization (GOR) term that encourages descriptors to spread uniformly over the feature space, and subsequent work further refines the loss function.
SOSNet~\cite{tian2019sosnet} adds a second-order similarity term to enforce consistency within and across descriptor pairs, HyNet~\cite{tian2020hynet} employs a hybrid similarity measure and a magnitude regularizer for more effective learning, and SDGMNet~\cite{deng2024sdgmnet} dynamically modulates the gradients of the triplet loss according to the statistics of training pairs. Keller \emph{et al.}~\cite{keller2018learning} show that common Siamese and triplet losses may yield inconsistently scaled descriptors, and address this with mixed-context losses and scale-aware sampling. In contrast to the widely used triplet loss, DOAP~\cite{he2018local} formulates a learning-to-rank objective that directly optimizes average precision.

Improving the invariance of descriptions to viewpoint and appearance changes has since become a focus. One direction introduces stronger geometric supervision. GeoDesc~\cite{luo2018geodesc} uses geometric constraints from multi-view reconstructions to mine hard training pairs by geometric error, and adds a geometric similarity loss that pulls together the descriptions of the same 3D point. CAPS~\cite{wang2020learning} uses epipolar and cycle-consistency losses as weak supervision from relative poses. It also designs a differentiable matching layer that models the matching probability distribution, and adopts a coarse-to-fine matching framework to refine descriptions progressively. These epipolar and cycle losses, the matching distribution formulation, and the coarse-to-fine design have motivated many later learnable matchers. Pultar \emph{et al.}~\cite{pultar2019leveraging} enrich the training data with patches from outdoor webcam sequences that cover illumination and seasonal changes. Instead of learning invariance, Toft \emph{et al.}~\cite{toft2020single} use single-image depth predicted by a CNN to rectify perspective distortions before extracting local features, which improves matching under strong viewpoint changes.
Another direction enriches descriptors with additional context. ContextDesc~\cite{luo2019contextdesc} augments off-the-shelf local descriptors with visual context from a global feature map and geometric context from the keypoint distribution, while MTLDesc~\cite{wang2022mtldesc} and AWDesc~\cite{wang2023awd} aggregate global and local context and incorporate spatial attention into descriptor learning. For efficient enhancement, FeatureBooster~\cite{wang2023featurebooster} boosts handcrafted or learned descriptors with a lightweight network that exploits keypoint geometry and the other keypoints in the same image, and Shrestha \emph{et al.}~\cite{shrestha2023residual} keep a handcrafted detector and learn descriptors as residuals over handcrafted ones. CrossFeat~\cite{schneider2026crossfeat} further learns to translate frozen descriptors across imaging modalities while preserving their geometric content.
Other methods design specialized network architectures. AffNet~\cite{mishkin2018repeatability} regresses affine transformations to normalize patches, Ebel \emph{et al.}~\cite{ebel2019beyond} enlarge the receptive field with log-polar patches to cover diverse scales, and GIFT~\cite{liu2019gift} applies group convolutions~\cite{cohen2016group} to rotated and rescaled image samples to encode transformation-equivariant features. Similarly, Lee \emph{et al.}~\cite{lee2023learning} employ steerable networks~\cite{weiler2019general} for explicit cyclic rotational equivariance instead of relying on data augmentation, whereas RoRD~\cite{parihar2021rord} learns rotation-robust descriptors through rotation augmentation and orthographic viewpoint projection. LISRD~\cite{pautrat2020online} jointly learns meta-descriptors at multiple regional scales and selects the level of invariance appropriate to each context, and SANDesc~\cite{durso2026sandesc} designs a compact attention-based U-Net descriptor that can be trained on top of different detectors. MATCHA~\cite{xue2025matcha} builds descriptors on vision foundation models, fusing diffusion and DINOv2 features into a single descriptor for geometric, semantic, and temporal matching.

\subsubsection{Joint Detector-Descriptor}
\label{subsubsec:joint_det_desc}
Isolated learnable detectors and descriptors perform well in normal scenes. However, under extreme conditions such as wide baselines, day-night changes, seasonal changes, or weakly textured scenes, their performance deteriorates sharply. One reason is that descriptors consider only local structures and rely heavily on low-level information while neglecting high-level features. Moreover, developing detectors and descriptors separately ignores the dependencies between them and leads to information loss and inconsistent optimization. Joint detector-descriptors address these issues by performing detection and description within a single end-to-end model.
We classify these methods by network structure into cascaded structures, where detection and description are performed sequentially; branched structures, where both are performed simultaneously from shared features; and decoupled structures, where the two modules are trained separately but designed to work together.

\noindent \textbf{Cascaded Structure.}
The cascaded design follows classical local features such as SIFT, whose covariant detectors select not only keypoint locations but also local regions that cover the same surface area in different images under geometric transformations, \emph{e.g.}, Harris corners under translation and rotation, and DoG or Hessian extrema additionally under scale changes~\cite{mikolajczyk2005comparison}. Each region is then normalized to a canonical patch before description~\cite{mikolajczyk2005performance}. LIFT~\cite{yi2016lift}, an early cascaded approach, mirrors this design with learnable modules. It uses a learnable detector to produce a score map and detects keypoints via a differentiable soft-argmax operation. It then crops keypoint neighborhoods for an orientation estimation module and finally extracts descriptions from patches rotated according to the estimated orientation using another learnable module. The three modules are trained progressively (descriptor, orientation, and then detector) for better convergence.
LF-Net~\cite{ono2018lf} implements a fully end-to-end pipeline with a Siamese structure. One branch extracts keypoints and descriptions differentiably. A learnable detection module identifies keypoints from a predicted score map, and a Spatial Transformer Network (STN)~\cite{jaderberg2015spatial} then crops local patches for a descriptor module. The other branch, which is non-differentiable and frozen, generates supervision targets.
Building on a similar methodology, RF-Net~\cite{shen2019rf} introduces receptive feature maps for more effective detection and a neighbor mask loss that facilitates patch selection and stabilizes descriptor training. S-TREK~\cite{santellani2023strek} also follows a detect-then-describe design, in which a translation- and rotation-equivariant detector is trained with a reinforcement-learning-inspired sequential procedure that rewards repeatability, and a lightweight descriptor is then learned for the detected keypoints.
In contrast, D2-Net~\cite{dusmanu2019d2} first computes dense descriptions for the full image and then identifies keypoints as local maxima (within and across channels) of these description maps using a soft local-maximum operation. ASLFeat~\cite{luo2020aslfeat} extends D2-Net with Deformable Convolution Networks (DCN)~\cite{dai2017deformable} to model local shapes, and improves keypoint localization by finding channel and spatial peaks on multi-level feature maps. ReDFeat~\cite{deng2022redfeat} introduces a mutual weighting strategy for the joint learning of cross-modal keypoint detection and description. Its successor SGPFeat~\cite{deng2026sgpfeat} further distills semantic priors from a vision foundation model and keypoint and descriptor priors from a visible-light local feature model, compensating for scarce and imperfectly aligned multi-modal training data.

\noindent \textbf{Branched Structure.}
Different from cascaded structures, which either crop patches based on score maps to generate descriptions or predict score maps from feature representations, branched structures use a shared backbone for both keypoint detection and feature description.
SuperPoint~\cite{detone2018superpoint}, an early example of this structure, introduces a self-supervised framework. Its detector, MagicPoint~\cite{detone2017toward}, is first trained on noise-contaminated synthetic shapes (quadrilaterals, triangles, lines, and ellipses), with ground-truth keypoints at junctions, line endpoints, and ellipse centers. A homographic adaptation strategy then generates pseudo-ground-truth keypoints on real images, on which the detector and a descriptor sharing the same backbone are trained jointly. Several works simplify or extend this self-supervised design. UnsuperPoint~\cite{christiansen2019unsuperpoint} learns from homographic image pairs without synthetic pre-training, SEKD~\cite{song2020sekd} lets the detector and descriptor iteratively improve each other, Tang \emph{et al.}~\cite{tang2019neural} use an outlier-rejection network as an auxiliary task to supervise keypoint learning, SiLK~\cite{gleize2023silk} reduces the framework to a simple, fully self-supervised pipeline, and ZippyPoint~\cite{kanakis2023zippypoint} adopts mixed-precision quantization and binary descriptors for fast inference. Bhowmik \emph{et al.}~\cite{bhowmik2020reinforced} further fine-tune such a network with rewards from a high-level task, namely relative pose estimation.
Following a similar branched architecture, R2D2~\cite{revaud2019r2d2} uses a modified L2-Net~\cite{tian2017l2} as its backbone and adds prediction heads for reliability and repeatability to the detector branch. HLDD~\cite{hu2025hldd} and LARKED~\cite{zhang2026larked} follow a similar reliability-aware design. HLDD uses a learned distinctiveness map to discard ambiguous points detected from a repeatability map and matches its hierarchically learned descriptors from coarse to fine, whereas LARKED applies non-maximum suppression directly to the learned reliability to sample keypoints with a lightweight network.
DISK~\cite{tyszkiewicz2020disk} also predicts score and feature maps from a shared U-Net, but trains them with reinforcement learning (RL)~\cite{sutton2018reinforcement} by formulating keypoint detection and matching as probabilistic processes. RIPE~\cite{kunzel2025ripe} extends this RL formulation to weak supervision, requiring only a binary label that indicates whether two images show the same scene.
ALIKE~\cite{zhao2022alike} also adopts a shared backbone and proposes a differentiable keypoint detection module for sub-pixel keypoint localization, with descriptors trained by a neural reprojection error loss. Its successor ALIKED~\cite{zhao2023aliked} introduces a sparse deformable description head that learns keypoint-specific deformable features. UPAL~\cite{costa2026upal} extends an ALIKED-style shared encoder with a line distance-field head, so that keypoints, descriptors, and line segments are obtained in a single forward pass.
SFD2~\cite{xue2023sfd2} embeds high-level semantics into detection and description. This encourages keypoints in reliable regions (\emph{e.g.}, buildings and traffic lanes) and suppresses them in unreliable areas (\emph{e.g.}, sky and cars), focusing computation on stable image content. SAMFeat~\cite{wu2025samfeat} distills category-agnostic semantics and edge cues from the Segment Anything Model into a lightweight shared-backbone network, so that SAM is not needed at inference. SDE2D~\cite{li2025sde2d} uses segmentation guidance during training to make descriptors more discriminative in weakly and repetitively textured indoor scenes, and Wang \emph{et al.}~\cite{wang2025semantic} fuse segmentation semantics into detection and description and further refine the network with semantic-guided and task rewards. Other designs revisit the representation used for detection and description, such as region-wise weighting~\cite{li2021rap}, salient-region features~\cite{jung2023salient}, low-level features~\cite{wang2022rethinking}, transformation-predictive representations~\cite{wang2023transformation}, and hierarchical detectors with contextual descriptors~\cite{he2023d2former}.

\noindent \textbf{Decoupled Structure.}
While most approaches favor joint learning for its performance benefits, others argue that decoupling detection and description mitigates training instability, since in a joint pipeline the failure of one component can impede the updates of both networks.
PoSFeat~\cite{li2022decoupling} addresses this with a delayed coupling strategy. The detection network remains decoupled until a discriminative and robust descriptor has been learned, which then enables reliable keypoint detection from the high-quality description map. SCFeat~\cite{sun2023shared} further improves this decoupled training with a shared coupling-bridge normalization between the description and detection networks.
DeDoDe~\cite{edstedt2024dedode} employs fully decoupled yet aligned detector and descriptor modules. Its detector learns keypoints directly from 3D consistency, using tracks from large-scale SfM reconstructions, while the descriptor is trained by maximizing a mutual nearest neighbor objective on these keypoints. DeDoDe~v2~\cite{edstedt2024dedode2} further applies non-maximum suppression to the detector's target distribution during training and adds data augmentations, improving the validity and robustness of keypoints. Steerers~\cite{bokman2024steerers} learns a linear transformation in the DeDoDe description space that emulates image rotations, making the descriptors rotation-equivariant, and AffSteer~\cite{bokman2024affine} extends this idea to affine transformations. RDD~\cite{chen2025rdd} follows the decoupling principle with a deformable-transformer descriptor network and a separate lightweight keypoint network trained after the descriptor, and complements MegaDepth with an air-to-ground training set. TraqPoint~\cite{liu2026traqpoint} adopts an RDD-style decoupled architecture, freezes its descriptor branch, and trains the keypoint network with track-aware policy gradients over multi-frame sequences, rewarding keypoints that remain consistent and distinctive along tracks.
XFeat~\cite{potje2024xfeat} also uses a decoupled structure, keeping a high image resolution while limiting the number of channels to balance accuracy and speed. Its detector is trained partly by distilling keypoints from ALIKE~\cite{zhao2022alike}, and a match refinement module refines keypoint locations based on local descriptions. LiftFeat~\cite{liu2025liftfeat} and SemLight~\cite{shi2026semlight} build on this lightweight design, but predict keypoints and descriptors from shared features and enrich the descriptors with auxiliary cues. LiftFeat fuses surface normals, learned from monocular-depth pseudo labels, into the 2D descriptors, and SemLight further distills DINOv3 semantics into a compact head and fuses appearance, geometric, and semantic cues through reliability-gated fusion.

\noindent \textbf{Application-Specific Features.}
Local features have also been tailored to specific domains and platforms, including texture images~\cite{zhang2018learningtexture}, retinal images~\cite{liu2022retinal}, low-light images~\cite{he2023darkfeat,wang2025derfeat}, underwater scenes~\cite{yang2023knowledge}, deformable objects~\cite{cadar2023deformable,potje2023dalf}, motion blur~\cite{zhao2024balf,gao2025motionfeat}, multi-robot SLAM~\cite{guo2023descriptor}, and task-aligned visual localization~\cite{liu2023taskaligned}. These methods mainly adapt the training data, supervision, or network capacity of general-purpose local features to the target domain.

\noindent \textbf{Keypoint and Match Refinement.}
Despite diverse strategies for supervising keypoint selection, what makes a ``good'' keypoint remains contested, and keypoints from different detectors vary considerably in localization accuracy. Instead of designing yet another detector, a separate line of work refines the keypoints or matches produced by existing detectors. GMM-IKRS~\cite{santellani2024gmm} is a non-learned, interpretable framework that refines and scores keypoints from arbitrary detectors. It applies image warpings, reprojects the detected points to the original image, and fits a robust Gaussian mixture model to refine stable keypoints while estimating scores for robustness and localization accuracy. Keypt2Subpx~\cite{kim2024learning} refines matched keypoints to sub-pixel accuracy after detection, description, and matching. It predicts the offsets of each matched pair from local image patches and descriptors and is trained with an epipolar objective, so that existing detectors can be improved by a lightweight detector-specific refiner. XRefine~\cite{schmid2026xrefine} removes the dependence on detector-specific descriptors and scores by refining matched keypoints from image patches alone with cross-attention. A single refiner is thus applicable to arbitrary detectors and extends to multi-view feature tracks.

\subsection{Learnable Outlier Filter}
\label{subsec:learn_filter}
\begin{figure}[!t]
	\centering
	\includegraphics[pagebox=mediabox,trim={507.2bp 324.5bp 364.5bp 66.9bp},clip,width=0.465\textwidth]{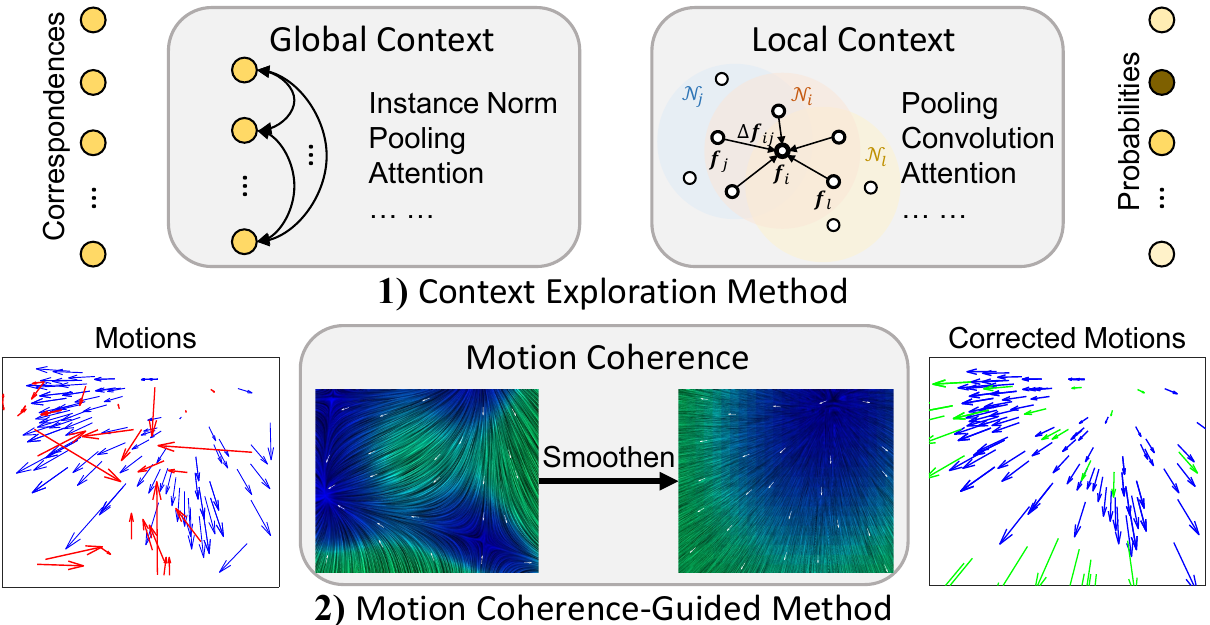}\hfill
	\includegraphics[pagebox=mediabox,trim={487.4bp 172.6bp 349.9bp 219.7bp},clip,width=0.495\textwidth]{Figure/outlier_filter.pdf}
	\caption{
		Frameworks of different learnable outlier filters.
	}
	\label{fig:outlier_filter}
\end{figure}

After keypoint detection and description, matching approaches establish correspondences between keypoints with similar descriptions. However, under extreme viewpoint changes, weak textures, or heavy occlusions, this putative correspondence set often contains many false correspondences (outliers). Outlier filters are therefore needed to identify the true correspondences (inliers) and improve the quality of the final correspondences.
Learning-based outlier filters formulate this task as binary classification of correspondences into inliers and outliers~\cite{qi2017pointnet,ma2019lmr}. Most filters take only the coordinates of putative correspondences as input, \emph{i.e.}, an $N\times4$ matrix of $(x,y,x',y')$. Due to the sparse and unordered nature of this input, most methods use Multi-Layer Perceptrons (MLPs) as the backbone, which process each correspondence individually. We group existing methods into two families according to how they exploit the relations among correspondences. The first family, which we term context exploration methods, aggregates global and local context from other correspondences to classify each correspondence. The second family, motion coherence-guided methods, builds on the observation that correct correspondences form a coherent motion field that is smooth except at discontinuities~\cite{ma2014robust,bian2017gms}, and explicitly estimates or regularizes this field before separating inliers from outliers.
Figure~\ref{fig:outlier_filter} illustrates the frameworks of these learnable outlier filters.

\subsubsection{Context Exploration Method}
\label{subsubsec:context_explore_outlier_filter}
Learning-based outlier filters originate from PointNet~\cite{qi2017pointnet}, which introduces an MLP-based backbone for classifying and segmenting irregular point clouds. Building on this PointNet-like structure, PointCN~\cite{yi2018learning}, the earliest work in this area, classifies inliers and outliers with a simple MLP backbone and uses context normalization (\emph{i.e.}, instance normalization) to capture global context.
Following this context exploration paradigm, more advanced context modules have been proposed to extract reliable context from both global and local areas.
Early modules are implemented with MLPs and pooling-like blocks.
For instance, OANet~\cite{zhang2019learning,zhang2020oanet} proposes an order-aware network with permutation-invariant Differentiable Pooling (DiffPool) and Differentiable Unpooling (DiffUnpool) layers. DiffPool clusters correspondences and represents each cluster with a compact embedding, from which local context is obtained. An order-aware filtering block at the bottom of the network then captures global context with context normalization, as in PointCN.
PointACN~\cite{sun2020acne} incorporates learnable weights into context normalization, using a weighted normalization supervised by inlier labels to reduce the impact of outliers on global context aggregation.
T-Net~\cite{xiao2024t} proposes a T-shaped network that uses the outputs of all layers to extract more robust global context, and introduces a permutation-equivariant context squeeze-and-excitation block to capture context along the channel dimension.
Beyond pooling-based schemes, some methods enhance local context within $k$-nearest neighbors ($knn$s) and derive global context progressively.
LMCNet~\cite{liu2021learnable} searches $knn$s in the coordinate space of raw correspondences to find spatially consistent neighbors, and applies max pooling within these neighbors to obtain local context.
Beyond spatial neighbors, NMNet~\cite{zhao2019nm} introduces a compatibility-specific mining strategy to find more reliable neighbors, based on the idea that compatible correspondences should be consistent in their local affine transformations. It then progressively aggregates local information into global context.
Other methods search $knn$s in the feature space. CLNet~\cite{zhao2021progressive} proposes an annular convolutional layer that retains detailed structural information while capturing local context among feature-space neighbors, and connects the local neighbors into a global graph whose embedding is computed with a graph convolutional network~\cite{kipf2017semi}.
MS$^2$DGNet~\cite{dai2022ms2dg} also constructs graphs in the feature space, extracting local context by max pooling within local areas and global context by context normalization.
NCMNet~\cite{liu2023progressive} progressively searches consistent neighbors in the coordinate space, the feature space, and a global graph space to capture both local and long-range relations among correspondences.
MGNet~\cite{luanyuan2024mgnet} constructs local graphs from implicit and explicit aspects and combines them into a global graph, on which a graph soft degree attention exploits all correspondences at once.
The attention mechanism~\cite{vaswani2017attention} has also been used to capture global and local context. GANet~\cite{jiang2022learning} applies fully connected attention to all correspondences to propagate long-range information.
ANA-Net~\cite{ye2023learning} introduces attention in attention to model second-order attentive context, extracting additional consistent context from the attention map.
U-Match~\cite{li2023u} integrates full attention into a U-Net-like structure to explore context and geometric cues hierarchically with graph pooling and unpooling~\cite{gao2019graph}. Its extended version~\cite{li2024u} further restructures the U-Net-like network to aggregate multi-level local features.
BCLNet~\cite{miao2024bclnet} also uses attention to capture local context.
VSFormer~\cite{liao2024vsformer} embeds visual cues into correspondences to identify inliers reliably in challenging scenes.
LeCoT~\cite{dai2026lecot} revisits the pruning backbone with spatial-channel fusion Transformer blocks and uses intermediate predictions to guide later stages, and CLCT~\cite{dai2026clct} models complementary local consensus within a Transformer. CorrMamba~\cite{fang2026corrmamba} replaces attention with a selective state-space model and learns a causal ordering of the unordered correspondences through Gumbel-Softmax, so that Mamba can mine information from likely inliers at low cost.
Beyond architecture design, Liao \emph{et al.}~\cite{liao2026scalable} pre-train correspondence pruners with masked inlier reconstruction in a masked-autoencoder framework, so that representations are learned without interference from outliers and transfer better across scenes.

\subsubsection{Motion Coherence-Guided Method}
\label{subsubsec:coherence_guided_outlier_filter}
Context exploration methods learn the relations among correspondences implicitly for classification, without explicitly modeling the coherence and smoothness of the motion field that conventional methods exploit~\cite{ma2014robust,bian2017gms}. Motion coherence-guided methods instead explicitly estimate or regularize the underlying motion field and identify inliers according to their consistency with it.
LMCNet~\cite{liu2021learnable} is the first to model motion coherence within a network. It derives a closed-form solution under a graph model and replaces specific terms of the motion smoothness regularization with learnable features.
Instead of this explicit smoothness constraint, ConvMatch~\cite{zhang2023convmatch} exploits motion coherence by converting unordered sparse motion vectors into a regular dense motion field, which it then smooths with a CNN
to enforce local consistency among neighboring motion vectors. Its extended version~\cite{zhang2024convmatch} refines the CNN backbone and proposes a bilateral convolution to preserve real discontinuities.
This idea is also used in DeMo~\cite{lu2025demo}, which applies reproducing kernel Hilbert space-based regularization~\cite{ma2014robust} with learnable kernels to model motion consensus, and further emphasizes it in both spatial and channel dimensions to distinguish discontinuities and avoid over-smoothing.
Inspired by Fourier expansion, DeMatch~\cite{zhang2024dematch} decomposes the motion field and retains its smooth ``low-frequency'' part, which achieves implicit regularization and naturally yields piecewise smoothness even under large disparities. Its extension DeMatch++~\cite{zhang2025dematchpp} further aggregates local context within each decomposed sub-field while enhancing global information across sub-fields.
CorrNeXt~\cite{li2025corrnext} strengthens ConvNet-style motion-field modeling by injecting 2D relative positional knowledge and decomposing large-kernel convolutions for efficient context aggregation, SC-Net~\cite{lin2026scnet} refines the motion field with spatial and cross-channel context to avoid over-smoothing under large disparities, and GeoMoE~\cite{le2026geomoe} decomposes the motion field into heterogeneous sub-fields guided by inlier probabilities and models them with a mixture of experts.

Although these outlier filters perform well in high-outlier scenarios, they often fail to generalize to matches produced by unseen detector-descriptors, even though they use only coordinates as input. In addition, when applied to nearly clean match sets, they risk over-rejecting valid correspondences. Future work should therefore target descriptor-agnostic filters that adapt to both simple and challenging scenarios.

\subsection{Learnable Geometric Estimator}
\label{subsec:learn_estimator}
\begin{figure}[!t]
	\centering
	\includegraphics[width=0.72\textwidth]{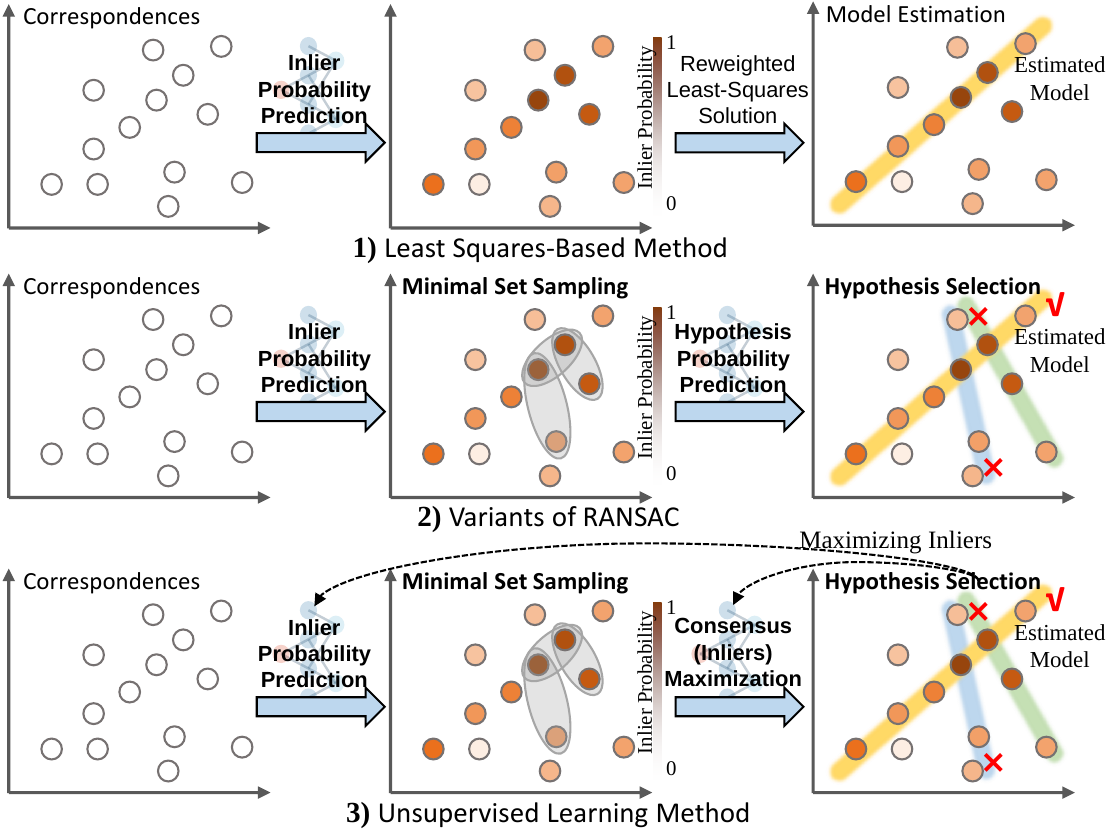}
	\caption{
		Frameworks of different learnable geometric estimators.
	}
	\label{fig:estimator}
\end{figure}

After outlier filtering, a geometric estimator is usually embedded into the pipeline to provide accurate transformation models for subsequent tasks. Traditional estimators such as DLT and RANSAC~\cite{fischler1981random} suffer from limited robustness or efficiency, which motivates learnable approaches that adapt least-squares solvers, improve RANSAC, or use unsupervised learning for better generalization. The boundary between learnable estimators and outlier filters is not strict. Many outlier filters also predict inlier weights and feed them into a weighted eight-point or DLT solver, while NG-RANSAC~\cite{brachmann2019neural} predicts per-correspondence sampling weights that resemble the inlier probabilities of a filter. We regard a method as an estimator when its learnable components operate inside model estimation, \emph{e.g.}, in solving, sampling, or hypothesis scoring, and as a filter when its main output is a pruned set of correspondences or inlier labels.
Figure~\ref{fig:estimator} illustrates the frameworks of these estimators.

\subsubsection{Least Squares-Based Method}
DFM~\cite{ranftl2018deep} is an early learnable estimator for the fundamental matrix. It solves a sequence of reweighted least-squares problems~\cite{torr1997development}, in which a PointNet-like network~\cite{qi2017pointnet} predicts correspondence weights from side information and from the residuals with respect to the previous model. After several iterations, DFM produces a reliable estimate of the fundamental matrix.

\subsubsection{Variants of RANSAC}
RANSAC remains the standard among conventional robust estimators.
It repeatedly samples minimal subsets to generate hypotheses, selects the hypothesis with the most inliers under an error threshold, and refits the final model using those inliers.
Learnable variants replace parts of RANSAC with learned components to accelerate and improve it.
DSAC~\cite{brachmann2017dsac} makes RANSAC differentiable by scoring hypotheses generated from uniform samples with a network and replacing deterministic hypothesis selection with a probabilistic one, so that the expected task loss can be minimized end-to-end.
Although designed for camera localization with 2D-3D scene coordinate prediction~\cite{miao2024survey}, DSAC's differentiable sampling and selection have inspired subsequent methods that learn minimal-set sampling and hypothesis selection.

\noindent \textbf{Learnable Minimal Set Sampling.}
Following DSAC, NG-RANSAC~\cite{brachmann2019neural} brings differentiable RANSAC to image matching by sampling minimal sets from a learned inlier distribution, which a PointCN-like network~\cite{yi2018learning} predicts from the putative correspondences, instead of sampling uniformly. Hypotheses generated from these sets are scored by inlier counting as in RANSAC, while the sampling network is trained by minimizing the expected task loss or, in a self-supervised manner, by maximizing the inlier count.
ARS-MAGSAC~\cite{wei2023adaptive} extends this framework by updating the predicted weights with a Bayesian rule, which decreases the inlier probabilities of the minimal set when an iteration fails to meet the RANSAC termination criterion.
It also adds a loss term that incorporates the orientation and scale provided by the detector.
BANSAC~\cite{piedade2023bansac} follows a related Bayesian idea without training. It models sample consensus as a dynamic Bayesian network that iteratively updates the inlier probability of each correspondence from the residuals of previous hypotheses, and uses these probabilities for both weighted sampling and a new stopping criterion.
Unlike the above methods, which differentiably sample geometric models over an entire hypothesis pool as in DSAC, $\nabla$-RANSAC~\cite{wei2023generalized} uses Gumbel softmax~\cite{jang2017categorical} to sample a good minimal set based on inlier scores from a lightweight network. For geometric matching, it also incorporates a relative pose error loss on rotation and translation and an average symmetric epipolar error loss over the residuals of all inliers.
Since these samplers are trained on correspondences from a specific matcher, their accuracy drops on matches from other front-ends. MCD~\cite{wang2026mcd} addresses this by training NG-RANSAC on ground-truth correspondences corrupted by a diffusion-style forward process with Monte Carlo-sampled noise levels, which improves generalization to unseen matchers without changing inference.

\noindent \textbf{Learnable Hypothesis Selection.}
As in DSAC, learning to select a good hypothesis is another option.
MQ-Net~\cite{barath2022learning} evaluates each hypothesis by computing the residuals of all correspondences, building a histogram over error levels, and feeding this histogram into a neural network that predicts a quality score. The same work introduces MF-Net, which analyzes the underlying motion to reject degenerate minimal sets early and thus improves efficiency.
The contemporaneous NeFSAC~\cite{cavalli2022nefsac} uses an MLP to assess each minimal sample before model estimation and scoring, so that unpromising samples are discarded early. It outputs a weighted average of confidence scores from several branches, including binary flags for outlier-free and non-degenerate configurations, and rejects motion-inconsistent and poorly conditioned samples.
FSNet~\cite{barroso2023two} evaluates hypotheses without explicit correspondences by processing the two images directly. Given a candidate geometric model, it aggregates image features along epipolar lines with an epipolar cross-attention block and predicts the relative rotation and translation errors.

\subsubsection{Unsupervised Learning Method}
Most of the aforementioned estimators rely on ground-truth transformations for supervision.
Unsupervised methods remove this dependency to improve robustness and generalization.
Probst \emph{et al.}~\cite{probst2019unsupervised} frame estimation as consensus maximization for polynomial transformations defined by a basis of linearly independent equations. The objective is to maximize the number of inliers while preserving the dimension of the polynomial space. In practice, this is implemented by maximizing inlier weights predicted by a PointNet-like network~\cite{qi2017pointnet} while minimizing the weighted sum of the singular values of the inliers' Vandermonde matrix~\cite{hu2012fast}.
To handle high outlier ratios, the network is first pre-trained on synthetic data and then trained on real data.
Truong \emph{et al.}~\cite{truong2021unsupervised} instead present an end-to-end unsupervised RL framework~\cite{sutton2018reinforcement} for consensus maximization. It iteratively minimizes the maximum residual and removes points from the feasible region (the basis set). The action of the RL agent is to remove a basis point, and the state records the status of each data point, \emph{i.e.}, whether it belongs to the basis and whether it has been removed. The reward encourages finding as many inliers as possible below a residual threshold. The method uses Q-learning~\cite{mnih2013playing}, where a DGCNN~\cite{wang2019dynamic} predicts rewards and is optimized by minimizing the temporal difference error. The final model is derived from the remaining points. An extended version~\cite{truong2022unsupervised} further explores alternative reward functions.
RL is also integrated with RANSAC in RLSAC~\cite{nie2023rlsac}, where the action is sampling the minimal set. The state comprises information on each data point, including its residual, its membership in the minimal set, and its usage history. The reward is the inlier ratio under the predicted model, so that maximizing the accumulated reward amounts to consensus maximization. A DGCNN-based policy network outputs inlier weights, and the points with the highest scores form a hypothesis.

However, most current learnable estimators focus on the essential or fundamental matrix and cannot fit arbitrary models as flexibly as traditional methods such as RANSAC. Moreover, their robustness across different matching pipelines and diverse scenarios remains underexplored, with only initial attempts to decouple their training from a specific matcher~\cite{wang2026mcd}.

\section{Merged Learnable Module}
\label{sec:merged}

Instead of replacing a single step, merged learnable modules integrate several consecutive steps of the classical pipeline into one network. According to the steps they merge, we review middle-end sparse matchers (Section~\ref{subsec:sparse_matcher}), end-to-end semi-dense and dense matchers (Sections~\ref{subsec:semi_matcher} and~\ref{subsec:dense_matcher}), and pose regressors (Section~\ref{subsec:pose_regressor}).

\subsection{Middle-End Sparse Matcher}
\label{subsec:sparse_matcher}

After keypoints are detected and described with off-the-shelf methods~\cite{lowe2004distinctive,detone2018superpoint}, tentative correspondences are formed by NN or MNN matching. Due to the limited discriminability of descriptors, these matches often contain many outliers. Outlier filters can remove some false matches, but their performance is bounded by the inliers present in the initial candidate set, and they treat visual descriptions and spatial coordinates separately, ignoring their interaction. These limitations motivate learnable sparse matchers, which jointly exploit visual and geometric information to overcome the bottlenecks of vanilla NN matching.

\begin{figure}[!t]
	\centering
	\includegraphics[width=0.66\textwidth]{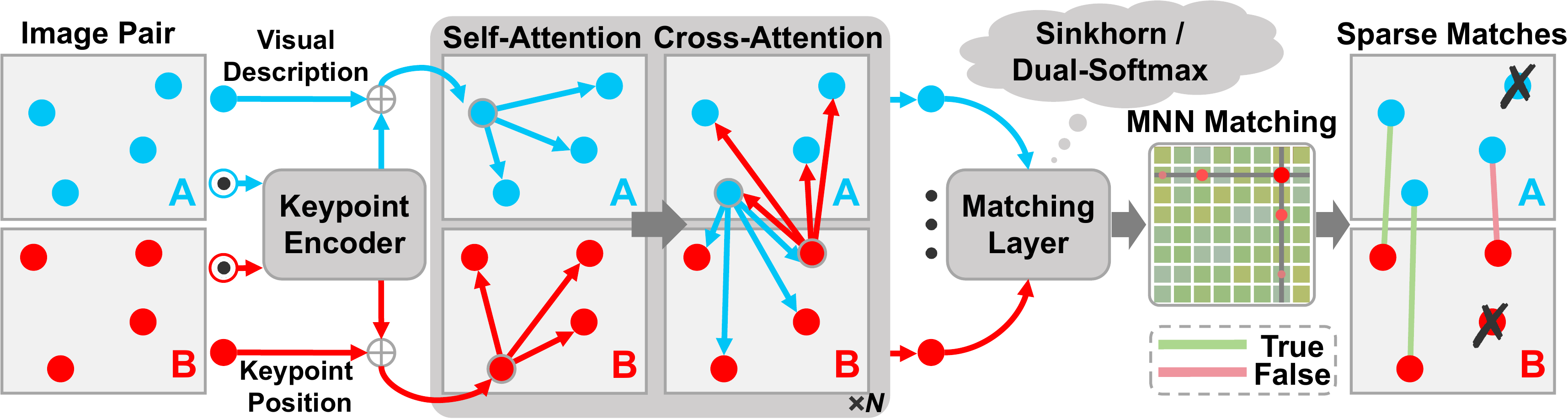}
	\caption{
		Framework of middle-end sparse matchers.
	}
	\label{fig:sparse_matcher}
\end{figure}

To this end, several studies~\cite{sarlin2020superglue,chen2021learning,shi2022clustergnn,lindenberger2023lightglue,jiang2024omniglue} formulate sparse feature matching as an assignment problem solved by an attention-based Graph Neural Network (GNN)~\cite{wu2020comprehensive}, as shown in Figure~\ref{fig:sparse_matcher}.
SuperGlue~\cite{sarlin2020superglue} exemplifies this approach. It builds fully connected intra- and inter-image graphs over keypoints and their descriptions, applies self- and cross-attention~\cite{vaswani2017attention} to reason jointly about spatial and visual cues, and applies the Sinkhorn algorithm~\cite{cuturi2013sinkhorn} to the resulting score matrix to produce matches. SuperGlue has become a standard baseline for learned sparse matching, but the $O(N^{2})$ cost of attention over $N$ keypoints limits its use in latency-sensitive applications.
Many works have therefore sought to improve \underline{\emph{efficiency}}. SGMNet~\cite{chen2021learning} first selects $K$ reliable seed matches with an NN matcher and then applies a sparse GNN in which attention is computed only between the seeds and all keypoints. This reduces the complexity from $O(N^{2})$ to $O(NK)$.
ClusterGNN~\cite{shi2022clustergnn} uses learnable hierarchical clustering to partition the keypoints into $K$ subgraphs and performs message passing only within each subgraph, which removes redundant connections and improves efficiency and scalability.
Rather than interleaving self- and cross-attention, ParaFormer~\cite{lu2023paraformer} performs both in parallel with shared cross-attention scores to reduce redundancy, and uses a wave-based positional encoding that unifies descriptions and positions through amplitude and phase. Its variant ParaFormer-U adopts a U-Net-like architecture with graph pooling to select informative keypoints and graph unpooling for reconstruction~\cite{gao2019graph}, further improving efficiency.
IMP~\cite{xue2023imp} jointly solves feature matching and relative pose estimation with a pose-consistency loss, so that matches and pose reinforce each other iteratively. Its accelerated variant EIMP adaptively prunes keypoints with low match potential, based on predicted matches, pose, and attention scores, without compromising accuracy.
Similarly, LightGlue~\cite{lindenberger2023lightglue} uses a matchability predictor to score each keypoint and a confidence classifier to decide when to stop inference. It prunes keypoints with low matchability and proceeds to deeper layers only if not enough keypoints are confident. Once a confident state is reached, it computes matches from an assignment matrix weighted by unary matchability. This adaptive mechanism adjusts both the depth and the width of the network to the difficulty of each image pair. LightGlue also uses Rotary Position Encoding (RoPE)~\cite{su2024roformer} to capture relative spatial context.
MaKeGNN~\cite{li2024learning} dynamically samples two compact sets of $K$ well-distributed keypoints with high matchability scores as message bottlenecks, so that each keypoint communicates only with matchable keypoints within and across images. The attention complexity is thus reduced to $O(NK)$.
MambaGlue~\cite{ryoo2025mambaglue} combines Mamba~\cite{gu2024mamba} and Transformer~\cite{vaswani2017attention} layers through a MambaAttention mixer, which selectively captures local and global context and achieves high accuracy with low latency.

Other methods focus on matching \underline{\emph{accuracy}}. SAM~\cite{lu2023scene} generates two group descriptions per image to represent overlapping and non-overlapping regions, captures scene-aware context between group and keypoint descriptions via self- and cross-attention, assigns matchable keypoints to the overlapping group, and derives the final matches by fusing group- and keypoint-level score matrices. Its extension SceneGlue~\cite{du2026sceneglue} combines implicit parallel attention with an explicit visibility Transformer that separates visible and invisible regions, without requiring scene-level annotation.
ResMatch~\cite{deng2024resmatch} recasts the GNN pipeline as an iterative process of matching and filtering by formulating self- and cross-attention as residual functions over the spatial and visual correlations between intra- and inter-image features. It injects relative positional similarity into self-attention and raw visual descriptions into cross-attention, so that matching and filtering are learned jointly. Its sparse variant sResMatch restricts the attention of each keypoint to neighbors selected by residuals, improving efficiency while retaining competitive accuracy. Similarly, CFM~\cite{liu2026cfm} couples a keypoint learning module with a correspondence learning module so that matching and match filtering reinforce each other progressively, and SGAT~\cite{zhang2026sgat} strengthens attention toward salient and distinctive singular points with high cross-view matching potential, which is especially helpful in low-texture and occluded regions.
OmniGlue~\cite{jiang2024omniglue} targets out-of-distribution generalization by using the DINOv2 foundation model~\cite{oquab2024dinov2} to select potential matches, so that each keypoint aggregates context only from these candidates. This suppresses irrelevant keypoints and focuses on matchable regions. OmniGlue also disentangles positional and appearance cues in attention, reducing its reliance on geometric priors and improving cross-domain transferability.
SemaGlue~\cite{zhang2025matching} instead improves generalization by integrating semantic priors into visual descriptions. It extracts semantic context with a pre-trained segmentation model (SegNeXt~\cite{guo2022segnext}), models channel-wise relationships between semantic and geometric features, and enriches local descriptions with the semantic representations. MapGlue~\cite{wu2025mapglue} extends graph-based sparse matching to multimodal remote sensing images with dual graph-guided attention, and releases a large-scale map-visible dataset for training and evaluation.
DiffGlue~\cite{zhang2024diffglue} embeds a diffusion model~\cite{croitoru2023diffusion} into sparse matching and uses its generative prior to progressively guide the assignment matrix toward the optimum. Specifically, it introduces assignment-guided attention, which resembles cross-attention but uses the assignment matrix as the attention map, thereby injecting correspondence priors into the GNN.
Match Any Keypoints~\cite{xu2026matchany} reduces the dependence on a specific detector-descriptor by sampling random positions as keypoints during training, replacing descriptors with multi-level sparse embeddings, and aggregating context from neighboring match proposals with a proposal-guided Transformer. LoMa~\cite{nordstrom2026loma} revisits local feature matching from a data-driven perspective by scaling data mixtures, model capacity, and compute, and introduces HardMatch, a set of 1000 manually annotated challenging image pairs, to counter benchmark saturation.

However, the performance of learnable sparse matchers is inherently bounded by the quality of the detected keypoints, and robust, repeatable detection remains challenging, particularly in low-texture scenes.

\subsection{End-to-End Semi-Dense Matcher}
\label{subsec:semi_matcher}

\begin{figure}[!t]
	\centering
	\includegraphics[width=0.90\textwidth]{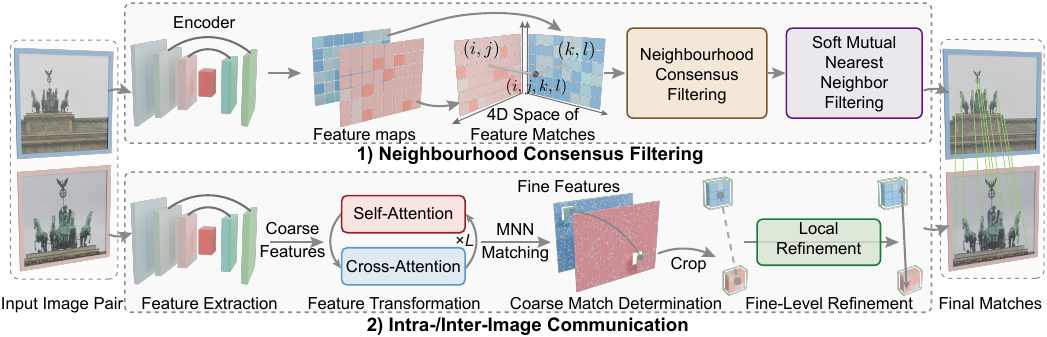}
	\caption{
		Frameworks of different end-to-end semi-dense matchers. Image refers to~\cite{he2025matchanything}.
	}
	\label{fig:semidense_matcher}
\end{figure}

Semi-dense matchers bypass explicit keypoint detection and establish matches directly from the raw image pair. They first match coarse patches instead of detected keypoints and then refine the matched positions, producing a large set of accurate correspondences. Depending on how coarse matches are established, they fall into neighborhood consensus filtering-based and intra-/inter-image communication-based matchers.

\subsubsection{Neighbourhood Consensus Filtering}
Early semi-dense methods used CNNs to process a 4D correlation volume, which encodes the matching space by recording the correlation scores between all feature pairs. This volume enables neighborhood consensus filtering, in which CNNs detect spatially consistent patterns, propagate context from confident matches to their neighbors, and select reliable correspondences, as illustrated in Figure~\ref{fig:semidense_matcher}.

As a pioneering semi-dense matcher, NC-Net~\cite{rocco2018neighbourhood} first extracts coarse feature maps, constructs a 4D correlation volume that enumerates all potential matches between an image pair, and applies 4D convolutions to regularize this volume and enforce neighborhood consensus. Final correspondences are then extracted by soft mutual nearest neighbor filtering, which ensures local and cyclic consistency.
Despite its encouraging performance, three limitations hinder its practical deployment: i) excessive memory usage due to the full 4D correlation volume; ii) high inference latency caused by 4D convolutions; and iii) poor localization due to low-resolution feature maps. To address these, Sparse-NCNet~\cite{rocco2020efficient} i) sparsifies the 4D correlation volume by retaining only the top-$K$ correspondences per feature; ii) replaces 4D convolutions with submanifold sparse convolutions for efficient neighborhood consensus filtering; and iii) employs a two-stage relocalization module to achieve sub-pixel accuracy.
Patch2Pix~\cite{zhou2021patch2pix} addresses the low matching resolution of such correspondence networks in a detect-to-refine manner. It takes their patch-level matches as proposals, regresses pixel-level correspondences within the local regions defined by these proposals under epipolar supervision, and jointly rejects outliers with confidence scores.
DualRC-Net~\cite{li2020dual,li2023dualrc} employs a dual-resolution, coarse-to-fine architecture for high-resolution images. It first extracts coarse- and fine-resolution feature maps. From the coarse features, it constructs a full 4D correlation volume refined by 4D convolution-based neighborhood consensus filtering. The filtered volume then guides the selection and reweighting of local regions in the fine-resolution feature map, from which the final correspondences are obtained. This design improves matching reliability and localization accuracy while avoiding the prohibitive cost of 4D convolutions on high-resolution features.
Building on DualRC-Net, DualRC-L~\cite{li2023dualrc} replaces standard 4D convolutions with sparse ones~\cite{rocco2020efficient}.
EDCNet~\cite{he2023efficient} further introduces a Psconv operator that approximates 4D convolutions on coarse features with linear complexity, and generates image-pair-specific 2D convolutions by weighting predefined prototype filters to improve robustness under illumination and viewpoint changes.

\subsubsection{Intra-/Inter-Image Communication}
Compared with neighborhood consensus filtering-based methods, which are constrained by limited receptive fields and search spaces, intra-/inter-image communication-based methods use the Transformer architecture~\cite{vaswani2017attention} to model long-range dependencies and achieve better performance. These methods typically comprise four stages: i) local feature extraction; ii) coarse feature transformation; iii) coarse-level match determination; and iv) fine-level match refinement, as illustrated in Figure~\ref{fig:semidense_matcher}.

As the pioneering work in this paradigm, LoFTR~\cite{sun2021loftr} uses a ResNet-FPN~\cite{lin2017feature} backbone to extract coarse features at \nicefrac{1}{8} resolution and fine features at \nicefrac{1}{2} resolution.
The coarse features are processed by $N$ layers of interleaved linear self- and cross-attention~\cite{katharopoulos2020transformers} with sinusoidal positional encoding~\cite{parmar2018image}, which efficiently makes them more distinctive. The transformed coarse features are correlated and normalized by dual-softmax to form an assignment matrix $\mathcal{S}$, from which coarse matches $\mathcal{M}_c$ are selected by MNN. Fixed-size patches around $\mathcal{M}_c$ are then cropped from the fine feature map and processed by attention, correlation, and expectation to regress sub-pixel matches $\mathcal{M}_f$.

Encouraged by LoFTR, many follow-up works have been proposed, mainly improving stages ii) and iv) to enhance matching \underline{\emph{accuracy}}. MatchFormer~\cite{wang2022matchformer} adopts an extract-and-match framework that interleaves self- and cross-attention to perform local feature extraction and transformation simultaneously.
ASpanFormer~\cite{chen2022aspanformer} introduces a global-local attention mechanism for multi-scale context interaction across image pairs, where the span of local attention adapts to intermediate flow and uncertainty estimates. Building on ASpanFormer, AffineFormer~\cite{chen2024affine} regularizes intermediate flow with affine consistency, fuses global and local context based on uncertainty, and incorporates a spatial softmax loss~\cite{wang2020learning} for better supervision.
3DG-STFM~\cite{mao20223dg} distills knowledge from an RGB-D teacher to an RGB student to transfer depth cues. In contrast, CSE~\cite{wang2023guiding} explicitly incorporates 3D geometry by fitting quadrics to monocular depth estimates~\cite{ranftl2021vision} to derive a curvature similarity map that is invariant to translation, rotation, and scaling, and combines it with the assignment matrix to guide coarse match selection. SEM~\cite{chang2023sem} instead exploits two-view epipolar geometry, iteratively restricting attention and matching to regions near the estimated epipolar lines. MASt3R~\cite{leroy2024grounding} grounds matching in 3D. It builds on DUSt3R~\cite{wang2024dust3r}, which regresses the pointmaps of two views in a common coordinate frame, and augments it with a dense local feature head trained with an additional matching loss; correspondences are then obtained by fast reciprocal matching of these features. Semantic priors from vision foundation models and language have also been introduced. SigMa~\cite{fang2026sigma} fuses CNN features with vision-foundation-model features through a cross-domain adapter and uses semantic similarity to guide matching, IMD~\cite{liu2025mindgap} aligns foundation-model embeddings with matching through diffusion features and cross-image interaction prompting, and TextFM~\cite{zheng2026textfm} uses textual embeddings from vision-language models as instance-level queries during coarse matching. TopicFM+~\cite{giang2024topicfm+} estimates a score map within each cropped patch to find a more reliable location for fine-level refinement instead of using the fixed patch center. CasMTR~\cite{cao2023improving} adds cascade matching at \nicefrac{1}{4} and \nicefrac{1}{2} resolutions to progressively increase and refine correspondences in both views, and applies a training-free non-maximum suppression detector as post-processing to retain keypoints in structurally informative regions. CasP~\cite{chen2025casp} also adopts a cascade, but decomposes coarse matching into two phases, where one-to-many priors from the first phase restrict the search range for one-to-one matches in the second. AdaMatcher~\cite{huang2023adaptive} unifies co-visible area estimation and context interaction. It predicts co-visible areas and uses a many-to-one assignment to identify patch-level correspondences within them, from which it estimates the inter-view scale ratio for alignment before sub-pixel regression. CoMatch~\cite{li2025comatch} further estimates covisibility dynamically to condense tokens and suppress messages from non-covisible regions, and refines matches bilaterally to sub-pixel accuracy.
To address scale differences, PATS~\cite{ni2023pats} divides the source image into equal patches and aligns them to target patches in a many-to-many fashion under visual similarity constraints, using an iterative scale-adaptive patch subdivision that refines correspondences from coarse to fine. ASTR~\cite{yu2023adaptive} handles scale discrepancies by adjusting the patch cropping size during fine-level refinement according to depth estimated from coarse matches and camera intrinsics. To enforce local consistency, namely that the matches of adjacent pixels remain close to each other across views, ASTR iteratively applies spot-guided attention to aggregate cross-view information from high-correlation regions identified in the coarse correlation matrix. Jin \emph{et al.}~\cite{jin2026entropy} observe that the MNN layer over-excludes matches under scale differences; they infer the scale ratio from the coarse score matrix and reformulate fine matching as cascaded flow refinement with a gradient loss for local consistency. Similar to LightGlue~\cite{lindenberger2023lightglue}, PRISM~\cite{cai2024prism} prunes irrelevant coarse-level features by maximizing inter-image dependency to focus on matchable regions. It further integrates feature similarity and matchability into a unified correlation matrix for precise coarse match proposals, and uses hierarchical aggregation to handle scale discrepancies. HomoMatcher~\cite{wang2025homomatcher} addresses the limited precision and continuity of prior point-to-patch methods with a lightweight homography estimation network for patch-to-patch alignment. It exploits geometric constraints to improve sub-pixel accuracy and allows matches to be inferred at arbitrary locations within aligned patches, supporting keypoint continuity and match densification.

Other methods improve matching \underline{\emph{efficiency}} while retaining competitive accuracy. QuadTree~\cite{tang2022quadtree} builds hierarchical token pyramids for coarse feature transformation and retains only the top-$K$ tokens with the highest attention scores at each level, progressively focusing on relevant regions and reducing the complexity of attention from quadratic to linear. From the perspective of latent topic modeling, TopicFM~\cite{giang2023topicfm} groups semantically similar tokens into topics to enable efficient message passing within each topic. Its extension~\cite{giang2024topicfm+} further proposes an efficient variant that merges tokens with context-aware topic embeddings instead of applying in-topic self- and cross-attention. Similarly, EcoMatcher~\cite{chen2025ecomatcher} uses coarse features as clustering centers, assigns similar features to each center to form clusters, and uses these clusters to guide efficient context interaction. Efficient LoFTR~\cite{wang2024efficient} redesigns LoFTR~\cite{sun2021loftr} with a lightweight RepVGG~\cite{ding2021repvgg} backbone, self-/cross-attention on aggregated tokens to reduce redundant computation, removal of dual-softmax during inference, and a two-stage correlation layer that handles positional variance in refinement. These changes substantially improve efficiency while maintaining competitive accuracy. ETO~\cite{ni2024etoefficient} approximates the continuous correspondence function in coarse matching by organizing tokens at \nicefrac{1}{8} resolution into groups, each associated with a homography hypothesis. This allows coarse feature transformation at \nicefrac{1}{32} resolution, greatly reducing the number of tokens processed by the Transformer. ETO+~\cite{ni2025etoplus} then revisits the refinement stage, correcting coarse matches with a hybrid Transformer-CNN interaction module before high-resolution refinement and allocating more capacity to coordinate regression. To address the quadratic complexity of Transformer-based methods, JamMa~\cite{lu2025jamma} proposes a linear-complexity matcher based on Mamba~\cite{gu2024mamba}. It employs a JEGO scanning-merge strategy, in which a joint scan enables high-frequency cross-view interaction, an efficient scan shortens the sequence, and a tailored scan path with local aggregators captures global omnidirectional features. Built on JamMa, PyraMatch~\cite{li2026pyramatch} replaces its scanning strategy with a multi-head pyramid scan. Lu \emph{et al.}~\cite{lu2026pmhi} further use progressively estimated matchability to guide hybrid feature interaction, balancing accuracy and latency. SLiM~\cite{choo2026slim} also adopts state-space modeling but focuses on the cost of coarse matching, combining a hybrid Conv-Mamba backbone with training-free norm-based filtering for sparse correlation and recurrent coordinate refinement. Also targeting coarse matching, Toida and Hotta~\cite{toida2026covisibility} jointly train a lightweight covisibility predictor to prune non-covisible tokens before coarse matching, reducing runtime and memory at high resolutions. UltraMatch~\cite{le2026ultramatch} avoids building the full token-to-token matching matrix. Instead, a lightweight transport path router ranks candidate target blocks for each source block and keeps only a few of them, and a sparse global dual-softmax matches tokens along these routes while preserving global competition. With structural reparameterization and a compact fine-matching head, UltraMatch runs faster than SuperPoint+LightGlue and handles 6K-resolution inputs on a single consumer GPU, and its router roughly halves the latency of other semi-dense matchers. EDM~\cite{li2025edm} instead revisits a CNN-based design, combining a deeper but thinner backbone, a correlation injection module, and a lightweight bidirectional axis-based regression head that predicts sub-pixel matches without high-resolution heatmaps. SURE~\cite{li2026sure} further makes such axis-wise offset regression evidential, so that each match carries aleatoric and epistemic uncertainty that can be used to filter unreliable correspondences.

Beyond accuracy and efficiency, the implicit keypoint selection of semi-dense matchers makes keypoint locations depend on the paired image, which fragments tracks across multiple views. Detect by Track~\cite{sekikawa2026det} addresses this at inference time by steering the fine-level matching of an off-the-shelf matcher, such as EDM or JamMa, from the sub-pixel location of an existing track, which produces connected multi-view tracks without retraining. Training data also limit semi-dense matchers. MatchGS~\cite{chen2026matchgs} corrects the geometry of 3D Gaussian Splatting to render accurate correspondence labels for novel views, improving their zero-shot performance. XoFTR~\cite{tuzcuoglu2024xoftr} adapts LoFTR to visible-thermal matching through masked image modeling pre-training on visible-thermal pairs, fine-tuning with pseudo-thermal augmentation, and a scale-aware refinement stage.

Despite these improvements, achieving a good balance between accuracy and efficiency remains an open challenge for semi-dense matchers.

\subsection{End-to-End Dense Matcher}
\label{subsec:dense_matcher}

\begin{figure}[!t]
	\centering
	\includegraphics[width=0.92\textwidth]{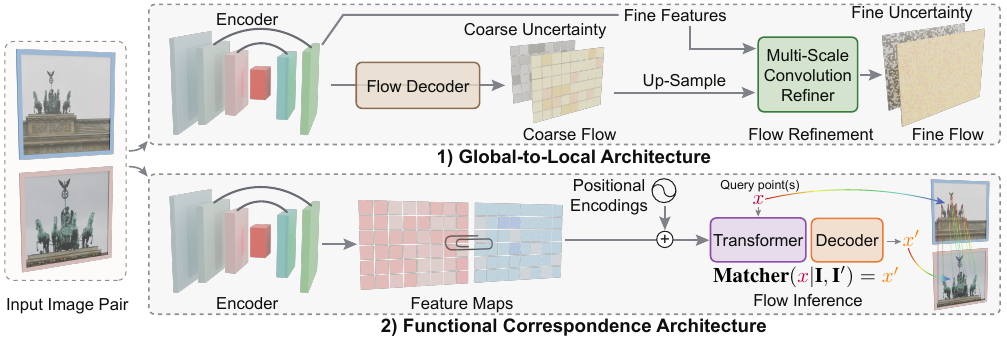}
	\caption{
		Frameworks of different end-to-end dense matchers. Image refers to~\cite{he2025matchanything}.
	}
	\label{fig:dense_matcher}
\end{figure}
Dense matchers regress a dense flow field between two views by processing a correlation volume of local or global pairwise similarities of deep features, unifying local feature matching and optical flow~\cite{dosovitskiy2015flownet}. The frameworks of dense matchers are shown in Figure~\ref{fig:dense_matcher}.

\subsubsection{Global-to-Local Architecture}
\label{subsubsec:global_to_local}
The global-to-local architecture can be traced back to the CNN architecture of Rocco \emph{et al.}~\cite{rocco2017convolutional}, which regresses parametric geometric transformations from feature correlations. As a precursor, DGC-Net~\cite{melekhov2019dgc} introduces a coarse-to-fine image warping approach for large displacements and appearance changes. It builds a feature pyramid and computes a global correlation volume at the coarsest level to predict an initial dense correspondence map. It then iteratively warps the source features with the current estimate, combines them with the reference features, and decodes a finer correspondence map. To handle ill-posed regions such as occlusions, it adds a matchability decoder that predicts pixel-wise confidence. DGC-Net requires a fixed input resolution of 240$\times$240 to keep the shape of the correlation volume constant, which limits its performance on high-resolution images.

To address this issue, GLU-Net~\cite{truong2020glu} introduces an adaptive-resolution architecture with two subnetworks, L-Net and H-Net. Given an image pair downsampled to a fixed size, L-Net computes a global correlation at the coarsest level and refines the flow with local correlations at finer levels. The resulting flow is upsampled and fed as an initial estimate to H-Net, which operates at the original resolution and further refines the flow with local correlation layers to produce sub-pixel dense correspondences. GLU-Net can thus handle both large and small displacements at arbitrary resolutions.
GOCor~\cite{truong2020gocor} replaces the feature correlation layer with an online optimization module to resolve ambiguities in repetitive or homogeneous regions. At inference, it minimizes a flexible term that enforces self-similarity in the reference image together with a regularization term that imposes spatial smoothness priors on the query image. Through iterative optimization, GOCor produces globally optimized correlation volumes that account for similar regions and matching constraints.
RANSAC-Flow~\cite{shen2020ransac} is a two-stage dense flow regression framework. It first performs coarse alignment with multiple homographies estimated by RANSAC~\cite{fischler1981random}. A self-supervised network then refines the alignment by predicting a dense flow and a matchability mask from local correlations. Trained with photometric and forward-backward consistency losses, RANSAC-Flow benefits from the RANSAC pre-alignment, which reduces the sensitivity of photometric losses to large appearance changes.
To address the poor generalization of dense matchers trained on synthetic warps and the limitations of unsupervised photometric losses under large appearance changes, WarpC~\cite{truong2021warp} proposes an unsupervised warp consistency objective. Given a real image pair $(\textbf{I}, \textbf{J})$, $\textbf{I}$ is warped to $\textbf{I}'$ with a random flow $\textbf{W}$ to form a triplet $(\textbf{I}, \textbf{I}', \textbf{J})$. The loss compares the flow composed along the path $\textbf{I}'\rightarrow\textbf{J}\rightarrow\textbf{I}$ with the direct flow $\textbf{I}'\rightarrow\textbf{I}$.

PDC-Net~\cite{truong2021learning} proposes a probabilistic framework that jointly estimates a dense flow field and a pixel-wise confidence map (\emph{i.e.}, flow uncertainty). It models the predictive distribution as a constrained mixture model that captures both accurate flow and outliers, and predicts its parameters from contextual cues in the correlation volume. To handle extreme viewpoint changes, PDC-Net adopts a multi-scale inference strategy that refines predictions based on uncertainty. It also introduces a self-supervised data pipeline that generates complex synthetic motions for uncertainty learning.
PDC-Net+~\cite{truong2023pdc} improves robustness in real-world scenes by augmenting training data with independently moving objects and introducing an injective criterion to mask out occluded regions that violate one-to-one ground-truth flow.
DKM~\cite{edstedt2023dkm} also models coarse flow probabilistically, but introduces a kernelized global matcher that combines a Gaussian process regressor for coarse flow with a CNN decoder that predicts flow coordinates and uncertainty. For local refinement, it applies depth-wise convolutions to stacked feature maps. To ensure both reliability and spatial coverage for pose estimation, DKM combines flow uncertainty with kernel density estimates to sample balanced correspondences.
PMatch~\cite{zhu2023pmatch} combines a LoFTR-style~\cite{sun2021loftr} encoder with a DKM-inspired warp refiner and is pre-trained with a paired masked image modeling pretext task to learn versatile visual features. It uses a global matcher based on the expectation over the correlation volume for robustness in texture-less regions, and adds a homography loss that regularizes planar surfaces locally.
Building on DKM, RoMa~\cite{edstedt2024roma} combines pre-trained coarse features from DINOv2~\cite{oquab2024dinov2} with specialized CNN fine features to form a precisely localized feature pyramid. It uses a Transformer-based embedding decoder that predicts anchor probabilities instead of regressing coordinates, which better represents the multimodal distributions in coarse matching, and improves the loss with regression-by-classification followed by robust regression. RoMa v2~\cite{edstedt2026romav2} redesigns the matching architecture and loss, trains on a curated and more diverse data distribution, decouples coarse matching and refinement into two training stages, and reduces refinement memory with a custom CUDA kernel. RoMa-$\Omega$~\cite{nordstrom2026romaomega} analyzes what feed-forward 3D models know about matching and finds that their features perform poorly for zero-shot matching, especially in later layers, but provide strong representations when a full matcher is trained on top. Replacing the DINO backbone of RoMa v2 with such a model further improves robustness. SPIDER~\cite{shao2026spider} observes that matches from 3D foundation models such as MASt3R tend to concentrate on dominant planar regions, and combines a RoMa-style 2D warp head with a MASt3R-style 3D descriptor head on a shared backbone. MV-RoMa~\cite{lee2026mvroma} extends pairwise dense matching to one source image and multiple co-visible targets, using pairwise matches as geometric priors for multi-view feature interaction and producing consistent tracks for SfM. Cai \emph{et al.}~\cite{cai2025evidential} replace the certainty estimation of dense matchers with evidential learning to obtain more robust confidence under challenging conditions. REDI-Match~\cite{ge2026redimatch} distills a non-equivariant foundation model into a lightweight rotation-equivariant encoder to handle severe in-plane rotations in dense matching.
To extract accurate affine correspondences from dense ones, DenseAffine~\cite{sun2025learning} extends DKM with a two-stage framework. The first stage uses a Sampson distance-based loss~\cite{hartley2003multiple} to improve epipolar consistency. The second stage estimates local affine transformations, decomposed into scale, orientation, and residual shape, and supervises them with an Affine Sampson Distance loss.

Optical flow methods, such as RAFT~\cite{teed2020raft}, GMFlow~\cite{xu2022gmflow}, and UniMatch~\cite{xu2023unimatch}, also estimate dense correspondences from global correlations followed by local refinement and can be applied to image matching, although they are mostly trained and evaluated on video frames with small displacements. UFM~\cite{zhao2025ufm} explicitly bridges the two tasks by training a unified model for co-visible pixels on both optical flow and wide-baseline matching data. PanMatch~\cite{zhang2025panmatch} similarly casts stereo matching, optical flow, and feature matching as 2D displacement estimation with features from large vision models. CRFT~\cite{liu2026crft} adapts recurrent flow refinement to cross-modal registration, learning a modality-independent feature flow and refining it at full resolution with discrepancy-guided attention.

Dense matchers are computationally expensive in both matching and the subsequent estimation. ArgMatch~\cite{deng2025argmatch} reduces the cost of refinement through adaptive refinement gathering, which combines a content-aware offset estimator, locally consistent match rectification, and locally consistent upsampling. DMS~\cite{astermark2025dms} targets the estimation stage and summarizes dense matches before RANSAC, reducing the estimation time by one to two orders of magnitude with little loss in accuracy.

\subsubsection{Functional Correspondence Architecture}
\label{subsubsec:functional_correspondence}
Instead of relying on correlation layers to capture local or global matching priors, COTR~\cite{jiang2021cotr} employs a functional correspondence network that takes a stitched image pair and a query coordinate in one image as input and directly regresses the corresponding coordinate in the other image with a Transformer~\cite{vaswani2017attention}. During inference, COTR recursively crops patches around the previous prediction and feeds them back into the network for refinement, forming a multi-scale pipeline that yields accurate matches. Its functional form allows flexible queries, either specific keypoints for sparse correspondences or all image coordinates for a dense flow field. However, the recursive refinement requires re-extracting features at each iteration, which is computationally expensive, and the cycle-consistency check used to reject outliers further doubles the computation.
ECO-TR~\cite{tan2022eco} accelerates COTR by organizing Transformer blocks in a stage-wise manner to progressively regress coordinates and uncertainty scores from feature-level crops of a multi-scale feature extractor. To support batch processing, it introduces a query clustering strategy that groups similar queries into shared patches.
UniCorrn~\cite{goswami2026unicorrn} also predicts the correspondences of query points, and shares one Transformer across 2D-2D, 2D-3D, and 3D-3D matching through a dual-stream decoder that separates appearance and positional features and iteratively refines the location of each query.

Despite these improvements, dense matchers remain much slower than sparse matchers and are often several times slower than semi-dense ones, which limits their use on high-resolution images.

\subsection{Pose Regressor}
\label{subsec:pose_regressor}

\begin{figure}[!t]
	\centering
	\includegraphics[pagebox=mediabox,trim={825.5bp 586.7bp 1086.1bp 118.7bp},clip,width=0.477\textwidth]{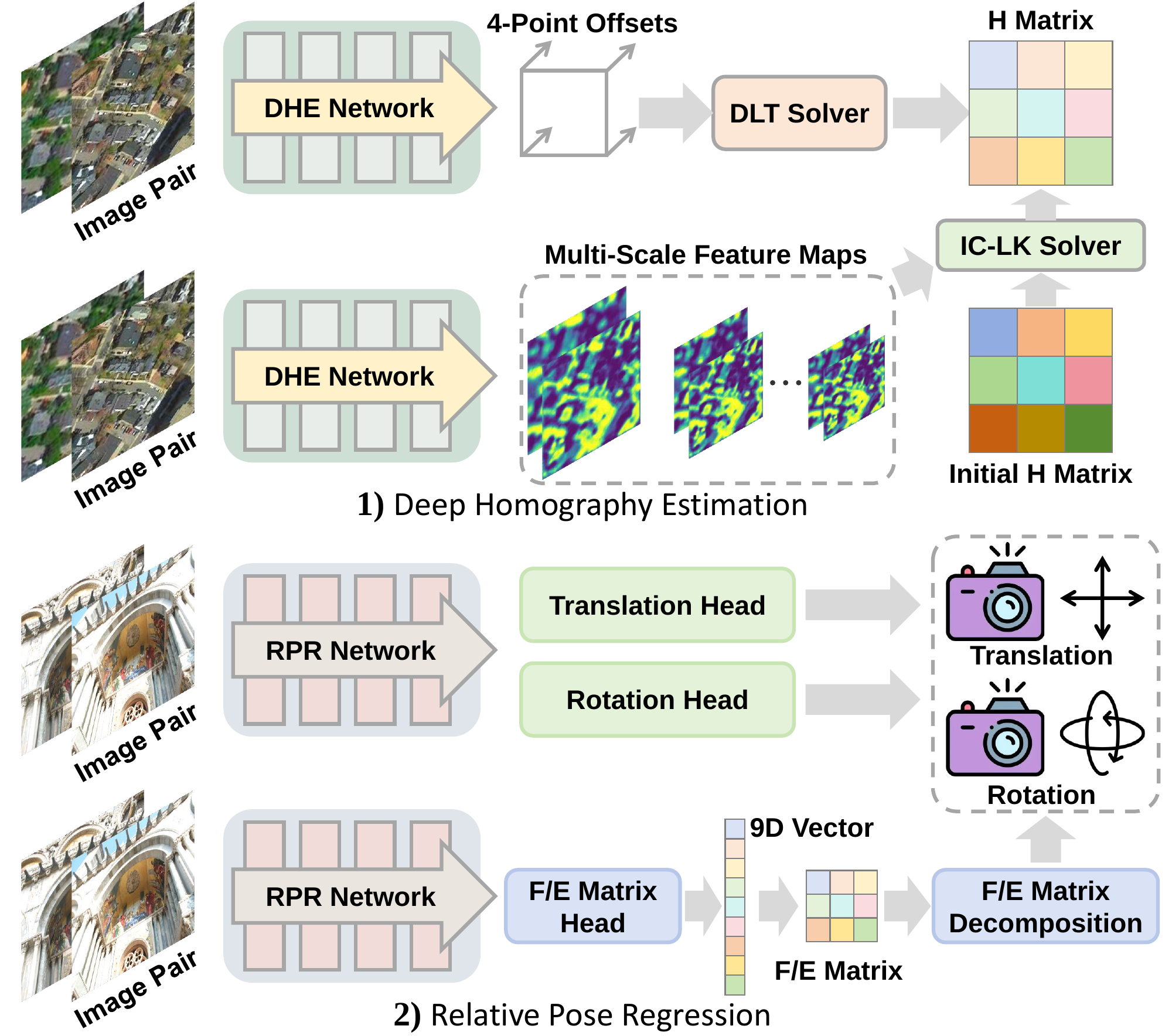}\hfill
	\includegraphics[pagebox=mediabox,trim={825.5bp 168.5bp 1074.2bp 555.6bp},clip,width=0.483\textwidth]{Figure/pose_regressor.pdf}
	\caption{
		Frameworks of different pose regressors.
	}
	\label{fig:pose_regressor}
\end{figure}

The image matching pipeline typically ends with pose estimation from established correspondences, which provides the geometric relationship between two views for downstream tasks. Some learning-based methods bypass correspondences and directly regress the pose from the image pair. These pose regressors can be categorized into Deep Homography Estimation (DHE) and Relative Pose Regression (RPR), as shown in Figure~\ref{fig:pose_regressor}.

\subsubsection{Deep Homography Estimation}

A homography is a general planar projective transformation represented by an $8$-degree-of-freedom (DoF) matrix $\mathbf{H} \in \mathbb{R}^{3 \times 3}$, normalized by fixing its last element to $1$.
DHE methods are trained either with supervision on synthetic data generated by perturbing image corners~\cite{cao2022iterative}, or without supervision through similarity losses~\cite{nguyen2018unsupervised,liu2022content}. Supervised approaches are more accurate but generalize worse to real images, whereas unsupervised ones generalize better but are harder to train. To bridge this gap, GFNet~\cite{zhang2025adapting} and DMHomo~\cite{li2024dmhomo} adopt more realistic data generation within supervised frameworks. Jiang \emph{et al.}~\cite{jiang2026supervised} extend this idea to both small- and large-baseline supervised homography learning with an iterative generation-training loop, and HomoGAN~\cite{liu2025homogan} enforces a coplanarity constraint with a GAN so that unsupervised estimation focuses on the dominant plane, extending it to mesh-grid local homographies. Other studies target cross-modal or degraded inputs. SSHNet~\cite{yu2025sshnet} splits unsupervised cross-modal homography estimation into two supervised sub-problems, namely homography estimation and modality transfer, trained with split optimization. HomoDiff~\cite{gao2026homodiff} interleaves multi-modal denoising and homography estimation within a diffusion process for noisy inputs, You \emph{et al.}~\cite{you2026generalized} synthesize unaligned training pairs with diverse textures and colors from single images to generalize to unseen modalities, and Liu \emph{et al.}~\cite{liu2026adverse} reformulate robust alignment under adverse degradations as progressive energy minimization. We categorize DHE methods by their homography parameterization: i) 4-point offset regression and ii) direct homography matrix parameterization.

\noindent \textbf{4-Point Offset Regression.}
Directly regressing the elements of $\mathbf{H}$ is unstable because its rotation and translation components have different scales. DHN~\cite{detone2016deep} instead regresses the offsets of the four image corners, which are then converted to $\mathbf{H}$ by DLT. As the first end-to-end homography network, DHN has inspired later work that improves regression accuracy under various conditions~\cite{nguyen2018unsupervised,le2020deep,cao2022iterative,cao2023recurrent}. For example, MHN~\cite{le2020deep} adopts a multi-scale cascade for coarse-to-fine estimation, which handles large deformations more effectively. IHN~\cite{cao2022iterative} replaces the cascade with iterative refinement within a single network. RHWF~\cite{cao2023recurrent} further combines homography-guided image warping with a focus Transformer in this recurrent framework to improve accuracy.

\noindent \textbf{Homography Matrix Parameterization.}
Some studies~\cite{zhao2021deep,zhang2024sparse} use the homography matrix $\mathbf{H}$ as the parameterization without regressing it directly. Instead, $\mathbf{H}$ is treated as an optimization variable and estimated with the IC-LK solver~\cite{baker2004lucas} to achieve feature-metric alignment between planes. These approaches focus on improving efficiency and convergence. For example, DeepLK~\cite{zhao2021deep} learns single-channel feature maps for faster optimization.
SDME~\cite{zhang2024sparse} learns features for both sparse and dense estimation within a multi-task network and uses a dedicated training strategy to achieve higher accuracy.

Other parameterizations have also been explored. Liu \emph{et al.}~\cite{liu2022unsupervised} parameterize a homography as a weighted combination of $8$ pre-computed flow fields, with a network predicting the weights, but the method is accurate only for small baselines. GFNet~\cite{zhang2025adapting} introduces a grid flow representation for flexibility with high-resolution inputs, at the cost of departing from the intrinsic $8$ DoF of a homography. Going beyond a single homography, EDFFDNet~\cite{zhu2025edffdnet} uses free-form deformation with exponential-decay basis functions for multi-grid registration in scenes with depth disparities, and CamFlow~\cite{li2025camflow} represents 2D camera motion with hybrid physical and stochastic motion bases. Finding a parameterization that balances flexibility, computational efficiency, and geometric fidelity therefore remains an open challenge in DHE.

\subsubsection{Relative Pose Regression}
RPR methods fall into two categories:
i) rotation-translation regression methods, which directly estimate the 6-DoF pose $(\textbf{R}, \textbf{t}) \in \textbf{SE}(3)$ from an image pair, where $\textbf{R} \in \textbf{SO}(3)$ is the rotation matrix and $\textbf{t} \in \mathbbm{R}^3$ is the translation vector in the camera frame; and
ii) essential/fundamental matrix regression methods, which estimate the essential/fundamental matrix for calibrated/uncalibrated cameras and then decompose it to recover the relative pose up to scale.

\noindent \textbf{Rotation-Translation Regression.}
As a pioneering work, Melekhov \emph{et al.}~\cite{melekhov2017relative} use a pre-trained Siamese network~\cite{zhou2014learning} to encode the two images into holistic embeddings, followed by an MLP that regresses a rotation quaternion and a scale-less translation.
This simple Siamese design has been widely adopted by subsequent methods.
For example, RPNet~\cite{en2018rpnet} explores several regression schemes and chooses to compute the relative pose from two separately regressed absolute poses, using the metric translation as supervision.
DirectionNet~\cite{chen2021wide} targets wide-baseline indoor scenes by decomposing the pose into four 3D unit direction vectors modeled as probability distributions on the sphere. It first estimates the rotation from three of these vectors via orthogonal Procrustes~\cite{schonemann1966generalized} and uses it to derotate the image pair, and then estimates the translation from the fourth vector.
Map-free~\cite{arnold2022map} computes a 4D correlation volume to warp the features and positional encoding of the second image, which are combined with the features of the first image into a scale-aware global embedding. An MLP then regresses the relative pose, and various continuous and discrete output parameterizations are explored for metric-scale RPR.
However, such methods depend on encoders tailored to fixed image sizes and camera intrinsics, which limits their generalizability. SRPose~\cite{yin2025srpose} addresses this by using keypoints and descriptions for metric-scale RPR. Keypoint coordinates are mapped to a unified camera space with the intrinsics, similarity-guided cross-attention establishes matches implicitly, and an MLP regresses the rotation and scaled translation under an epipolar constraint. Reloc3R~\cite{dong2025reloc3r} instead scales up training to millions of posed image pairs with a symmetric pose regression network, and combines relative pose regression with a lightweight motion averaging module for visual localization. Alligat0R~\cite{loiseau2025alligat0r} pre-trains such regressors with covisibility segmentation instead of cross-view completion, predicting whether each pixel is covisible, occluded, or outside the field of view, and EquiPose~\cite{liu2025equipose} enforces pose permutation equivariance so that swapping the two input images yields the inverse relative pose.

Some studies focus on rotation-only regression. Zhou \emph{et al.}~\cite{zhou2019continuity} introduce continuous 5D and 6D representations mapped to $\textbf{SO}(3)$ via stereographic projection and partial Gram-Schmidt orthogonalization, instead of discontinuous representations such as quaternions or Euler angles.
Levinson \emph{et al.}~\cite{levinson2020analysis} project a continuous 9D representation onto $\textbf{SO}(3)$ via SVD orthogonalization.
To handle extreme rotations with limited overlap, DenseCorrVo~\cite{cai2021extreme} constructs a 4D correlation volume that captures cues for both overlapping and non-overlapping pairs, and predicts discretized absolute pitch and relative yaw instead of regressing 3D rotations directly. Bezalel \emph{et al.}~\cite{bezalel2025extreme} extend extreme rotation estimation to Internet image pairs with limited or no overlap and varying intrinsics, and contribute the ExtremeLandmarkPairs dataset.

\noindent \textbf{Fundamental/Essential Matrix Regression.}
For fundamental matrix regression, Poursaeed \emph{et al.}~\cite{poursaeed2018deep} propose a single-stream model that concatenates both images and a Siamese model that processes each image separately before merging features. Rather than regressing the nine matrix entries directly, they explore two parameterizations, one based on camera parameters and the other on the epipolar parameterization, to enforce the rank-$2$ structure and $7$ DoF of the fundamental matrix. For essential matrix regression, Zhou \emph{et al.}~\cite{zhou2020learn} use a neighborhood consensus layer to build a global correlation volume. A CNN regressor then predicts a 9D vector approximating the essential matrix, which is projected onto the valid manifold by averaging its two largest singular values and setting the smallest to zero. Owing to scale ambiguity, low accuracy, and poor generalization, this paradigm has attracted few follow-up works.

In summary, RPR methods bypass explicit correspondence estimation and thus offer an alternative to pipelines that are vulnerable to matching errors. However, most of them provide no confidence measure for their predictions, which limits their reliability in practice.

\section{Experiment}
\label{sec:exp}

\subsection{Datasets}
\label{subsec:dataset}

\noindent \textbf{YFCC100M~\cite{thomee2016yfcc100m}} comprises nearly 100 million Creative Commons Flickr images and videos, accompanied by metadata such as camera parameters, user tags, and partial geolocation. Following the protocol in~\cite{zhang2019learning,zhang2020oanet}, 72 landmark sequences of outdoor scenes are selected (68 for training/validation and 4 for testing), with ground-truth (GT) poses and 3D models reconstructed by COLMAP~\cite{schonberger2016structure}. These GT poses are SfM-based pseudo ground truth, whose accuracy has not been independently verified and can affect benchmark rankings~\cite{jin2021image,brachmann2021limits}.

\noindent \textbf{SUN3D~\cite{xiao2013sun3d}} contains 254 indoor RGB-D sequences with sparse textures, repetitive patterns, and self-occlusions. GT relative poses are refined by generalized bundle adjustment~\cite{hartley2003multiple}. Following~\cite{zhang2020oanet}, 239 sequences are used for training/validation and the rest for testing.

\noindent \textbf{MegaDepth~\cite{li2018megadepth}} contains SfM and MVS reconstructions of 196 landmarks from Internet photos. Built with COLMAP and MVS~\cite{schonberger2016pixelwise}, it provides RGB images, dense depth maps, camera parameters, and sparse 3D models. Its challenging conditions, such as extreme viewpoint changes and repetitive patterns, make it a standard benchmark for outdoor matching and relative pose estimation. Evaluation typically uses splits such as MegaDepth-1500~\cite{sun2021loftr}, which samples 1500 image pairs from scenes 0015 (St.\ Peter's Basilica) and 0022 (Brandenburger Tor).

\noindent \textbf{ScanNet~\cite{dai2017scannet}} comprises 1613 indoor RGB-D sequences with GT poses and depth maps, featuring repetitive structures and weak textures. It is used to benchmark indoor matching with test splits such as the 1500 pairs used in~\cite{sarlin2020superglue,sun2021loftr}.

\noindent \textbf{HPatches~\cite{balntas2017hpatches}} comprises 116 image sequences, each containing one reference image and five query images with GT homographies. Among them, 57 sequences involve illumination changes and 59 involve viewpoint changes, making HPatches a standard benchmark for homography estimation with both handcrafted and learning-based methods.

\noindent \textbf{IMC2021~\cite{jin2021image}} is the benchmark of the Image Matching Challenge 2021, which evaluates wide-baseline two-view matching on outdoor scenes, with ground truth hidden on the official evaluation server. We use its PhotoTourism and PragueParks stereo tracks.

\noindent \textbf{Aachen Day-Night v1.1~\cite{sattler2018benchmarking}} is a large-scale outdoor localization dataset covering the historic city center of Aachen, with 6697 daytime reference images taken by handheld cameras and 1015 query images (824 day and 191 night) captured by mobile phones. Its challenges include extreme illumination changes, viewpoint changes, and complex urban geometry, and the query poses are evaluated on an online benchmark server.

\noindent \textbf{InLoc~\cite{taira2018inloc}} comprises 9972 RGB-D images geometrically registered to floor maps and 329 RGB queries taken with an iPhone 7, whose 6-DoF poses are manually verified. Its indoor scenes exhibit large viewpoint changes, occlusions, illumination changes, moving furniture, and repetitive, low-texture structures, making InLoc a challenging benchmark for indoor localization.

\noindent \textbf{ETH3D~\cite{schops2017multi}} provides high-resolution multi-view images with calibrated camera intrinsics and poses and laser-scan ground truth for evaluating 3D reconstruction. The ground truth of its test split is hidden and evaluated on the official server. We use it to assess whether improvements in two-view matching transfer to multi-view reconstruction.

\subsection{Metrics}
\label{subsec:metrics}
\subsubsection{Relative Pose Estimation}
To evaluate estimated camera poses, a common approach measures the angular errors of rotation and translation~\cite{jin2021image,sarlin2020superglue} and computes the Area Under the Curve (\textbf{AUC}) of the pose error distribution. Specifically, given $N$ test image pairs with ground-truth relative rotations $\{\mathbf{R}_i\}$ and translations $\{\mathbf{t}_i\}$ (known only up to scale), and the corresponding estimates $\{\hat{\mathbf{R}}_i,\hat{\mathbf{t}}_i\}$, the rotation and translation errors of each pair are defined as:

\begin{equation}
	\Delta\theta_i^\mathrm{rot} =
	\arccos\left(
	\tfrac{1}{2}(\mathrm{tr}(\mathbf{R}_i^\top\hat{\mathbf{R}}_i)-1)
	\right),
\end{equation}
\begin{equation}
	\Delta\theta_i^\mathrm{trans} =
	\arccos\left(
	\frac{|\hat{\mathbf{t}}_i^\top \mathbf{t}_i|}{\|\hat{\mathbf{t}}_i\|\|\mathbf{t}_i\|}
	\right).
\end{equation}
where the absolute value accounts for the sign ambiguity of the translation direction. The two errors are combined into a single error per pair by taking the maximum:
\begin{equation}
	\epsilon_i = \max\bigl(\Delta\theta_i^\mathrm{rot},\Delta\theta_i^\mathrm{trans}\bigr).
\end{equation}
For a threshold $\varepsilon$, the recall over all pairs is:
\begin{equation}
	R(\varepsilon) = \frac{1}{N}\sum_{i=1}^N \mathbf{1}\{\epsilon_i < \varepsilon\},
\end{equation}
where $\mathbf{1}\{\cdot\}$ is the indicator function. Plotting $R(\varepsilon)$ against $\varepsilon$ (in degrees) yields the error-recall curve, and the \textbf{AUC} up to a maximum threshold $\varepsilon_{\max}$ is:
\begin{equation}
	\textbf{AUC}@\varepsilon_{\max}
	= \frac{1}{\varepsilon_{\max}} \int_{0}^{\varepsilon_{\max}} R(\varepsilon)\mathrm{d}\varepsilon.
	\label{eq:real_auc}
\end{equation}
In practice, thresholds $\{\varepsilon_j\}_{j=0}^M$ are sampled in $[0,\varepsilon_{\max}]$, and the \textbf{AUC} is approximated with the trapezoidal rule:
\begin{equation}
	\textbf{AUC}@\varepsilon_{\max} \approx \frac{1}{\varepsilon_{\max}} \sum_{j=1}^{M} \frac{R(\varepsilon_{j-1}) + R(\varepsilon_j)}{2} (\varepsilon_j - \varepsilon_{j-1}).
\end{equation}

We report \textbf{AUC}$@5^\circ,10^\circ,20^\circ$~\cite{sarlin2020superglue,zhang2024convmatch}, where $5^\circ$, $10^\circ$, and $20^\circ$ are the maximum thresholds $\varepsilon_{\max}$.
For the IMC2021 benchmark, we report the mean Average Accuracy (\textbf{mAA}) following~\cite{jin2021image}, which averages the accuracy over the thresholds $\mathcal{T}_{10}=\{1^\circ,2^\circ,\ldots,10^\circ\}$:
\begin{equation}
	\textbf{mAA}@10^\circ
	= \frac{1}{|\mathcal{T}_{10}|}
	\sum_{\tau\in\mathcal{T}_{10}}
	\frac{1}{N}\sum_{i=1}^N \mathbf{1}\{\epsilon_i < \tau\}.
\end{equation}

\subsubsection{Homography Estimation}
Following~\cite{balntas2017hpatches}, given the GT homography $\mathbf H$ and the estimated homography $\hat{\mathbf H}$, the estimate is evaluated by the average reprojection error of the four image corners:
\begin{equation}
	\epsilon_i = \frac{1}{4}\sum_{n=1}^4 \bigl\lVert \pi(\mathbf H\,\tilde{\mathbf c}_n) - \pi(\hat{\mathbf H}\,\tilde{\mathbf c}_n)\bigr\rVert_2,
\end{equation}
where $\tilde{\mathbf c}_n=[u_n,v_n,1]^\top$ are the homogeneous coordinates of the four corners $(0,0),(W,0),(W,H),(0,H)$, and $\pi([a,b,c]^\top)=[a/c,b/c]^\top$.
For a threshold $\varepsilon$, the accuracy \textbf{Acc.} is:
\begin{equation}
	\textbf{Acc.}@\varepsilon
	= \frac{1}{N}\sum_{i=1}^N \mathbf{1}\{\epsilon_i < \varepsilon\}\,,
\end{equation}
where $N$ is the number of test pairs.
To summarize performance across thresholds, the \textbf{AUC} for homography estimation is defined as:
\begin{equation}
	\textbf{AUC}@\varepsilon_{\max}
	\approx \frac{1}{\varepsilon_{\max}}
	\sum_{j=1}^{M}
	\frac{\textbf{Acc.}(\varepsilon_{j-1}) + \textbf{Acc.}(\varepsilon_j)}{2}
	(\varepsilon_j - \varepsilon_{j-1})\,.
\end{equation}
We report \textbf{Acc.}$@3,5,10$px and \textbf{AUC}$@3,5,10$px, where each \textbf{AUC} is computed up to the corresponding threshold.

\subsubsection{Matching Accuracy}
For dense matchers, matching accuracy is measured by the Percentage of Correct Keypoints (\textbf{PCK})~\cite{truong2020glu}, \emph{i.e.}, the proportion of accurately matched points among densely sampled correspondences. Given $N$ GT keypoint pairs $\{(\mathbf p_i,\mathbf q_i)\}$ and the predicted match locations $\{\hat{\mathbf q}_i\}$, the per-keypoint error is defined as:
\begin{equation}
	\epsilon_i = \Vert \mathbf q_i - \hat{\mathbf q}_i \Vert_2\,.
\end{equation}
For a threshold $\varepsilon$, \textbf{PCK} is defined as:
\begin{equation}
	\textbf{PCK}@\varepsilon
	= \frac{1}{N}\sum_{i=1}^N \mathbf{1}\{\epsilon_i < \varepsilon\}\,,
\end{equation}
where $\mathbf{1}\{\cdot\}$ is the indicator function, and $N$ is the number of GT keypoint pairs.
We report \textbf{PCK}$@0.5,1,3,5$px.

\subsubsection{Visual Localization}
Visual localization is evaluated by the percentage of queries localized within given distance and orientation thresholds. Given $N$ queries with GT poses $\{\mathbf R_i,\mathbf t_i\}$ and estimates $\{\hat{\mathbf R}_i,\hat{\mathbf t}_i\}$, the precision at a distance threshold $d$ and an orientation threshold $\theta$ is:
\begin{equation}
	\textbf{Prec}(d,\theta)
	= \frac{1}{N}\sum_{i=1}^N \mathbf{1}\bigl\{
	\|\hat{\mathbf t}_i-\mathbf t_i\|_2 < d
	\;\wedge\;
	\arccos\bigl(\tfrac{1}{2}(\mathrm{tr}(\mathbf R_i^\top\hat{\mathbf R}_i)-1)\bigr) < \theta
	\bigr\}\,,
\end{equation}
where $\mathbf{1}\{\cdot\}$ is the indicator function.

\subsubsection{3D Reconstruction}
For 3D reconstruction, we report statistics of the reconstructed COLMAP models that do not require GT geometry, namely the number of triangulated 3D points, the mean track length, and the mean reprojection error.

For the official ETH3D high-resolution benchmark~\cite{schops2017multi}, given a reconstructed point cloud $\mathcal{P}_r$, the laser-scan GT point cloud $\mathcal{P}_g$, and a distance threshold $\tau$, let $d(\mathbf p,\mathcal{P})=\min_{\mathbf q\in\mathcal{P}}\|\mathbf p-\mathbf q\|_2$. \textbf{Acc.}$@\tau$ is the percentage of reconstructed points whose nearest GT point is within $\tau$, and \textbf{Comp.}$@\tau$ is the percentage of GT points whose nearest reconstructed point is within $\tau$:
\begin{equation}
	\textbf{Acc.}@\tau
	= \frac{1}{|\mathcal{P}_r|}\sum_{\mathbf p\in\mathcal{P}_r}\mathbf{1}\{d(\mathbf p,\mathcal{P}_g)<\tau\},
\end{equation}
\begin{equation}
	\textbf{Comp.}@\tau
	= \frac{1}{|\mathcal{P}_g|}\sum_{\mathbf q\in\mathcal{P}_g}\mathbf{1}\{d(\mathbf q,\mathcal{P}_r)<\tau\}.
\end{equation}
The \textbf{F1} score is the harmonic mean of accuracy and completeness:
\begin{equation}
	\textbf{F1}@\tau
	= \frac{2\cdot \textbf{Acc.}@\tau \cdot \textbf{Comp.}@\tau}{\textbf{Acc.}@\tau+\textbf{Comp.}@\tau}.
\end{equation}

\subsection{Experimental Settings}
\label{subsec:experiment_setting}

\subsubsection{Relative Pose Estimation}
We conduct relative pose estimation experiments on four datasets: MegaDepth~\cite{li2018megadepth}, YFCC100M~\cite{thomee2016yfcc100m}, ScanNet~\cite{dai2017scannet}, and SUN3D~\cite{xiao2013sun3d}. For alternative learnable steps, we follow standard evaluation protocols~\cite{sarlin2020superglue,lindenberger2023lightglue}. For MegaDepth (the default setting), images are resized so that the longest side is 1600 pixels, and up to $2048$ keypoints are extracted per image. The RANSAC~\cite{fischler1981random} inlier threshold is set to $0.5$ divided by the focal length. For YFCC100M and SUN3D, images are not resized; for ScanNet, images are resized to $640\times480$.
For middle-end sparse matchers, which take keypoints and descriptions as input, the following settings are used. For MegaDepth (default), the longest side is resized to $1600$ pixels, up to $2048$ keypoints are extracted per image, and the RANSAC threshold is $0.5$ divided by the focal length. For YFCC100M, the RANSAC threshold is $1$ divided by the focal length. For ScanNet, images are resized to $640\times480$ and only $1024$ keypoints are extracted per image. For SUN3D, the longest side is resized to $640$ pixels and $1024$ keypoints are extracted.
For end-to-end semi-dense/dense matchers, we follow an open-source evaluation pipeline\footnote{\url{https://github.com/PruneTruong/DenseMatching}}, whose configurations differ from those above. For YFCC100M, ScanNet, and SUN3D (default), images are resized so that their shortest side is $480$ pixels, and some methods additionally pad images to meet their resolution requirements. The RANSAC threshold is $1$ divided by the focal length. For MegaDepth, most methods resize images to $1152\times1152$, except for DKM~\cite{edstedt2023dkm} ($880\times660$) and RoMa~\cite{edstedt2024roma} ($672\times672$), and the RANSAC threshold is $0.5$ divided by the focal length. For the additionally evaluated dense matchers, PMatch and DenseAffine use a maximum side length of $1200$ pixels outdoors and a $480$-pixel short side indoors, ArgMatch uses symmetric matching with $608\times800$ outdoor and $480\times640$ indoor inputs, and MASt3R uses $512$-pixel inputs. DenseAffine is followed by OpenCV's five-point RANSAC. Pose regressors use their released preprocessing, and SRPose is evaluated on all datasets with its MegaDepth checkpoint.
For IMC2021, we follow the official Image Matching Benchmark protocol on the PhotoTourism and PragueParks stereo test tracks. Detector-descriptor pipelines use up to $2048$ keypoints per image, and SIFT uses RootSIFT descriptors. Learned outlier filters are applied to the same NN matches, with coordinates normalized by the camera intrinsics, and the filtered correspondences are passed to the official RANSAC stage unless another estimator is listed. Semi-dense/dense matchers use their reported preprocessing and input resolutions. Runtime is measured in milliseconds per image pair for the complete pipeline, from the input images to the estimated relative pose. Runtime includes RANSAC, whose number of iterations decreases as the inlier ratio increases.

\subsubsection{Homography Estimation}
Homography estimation experiments are conducted on HPatches~\cite{balntas2017hpatches}, following the protocols of~\cite{lindenberger2023lightglue,edstedt2023dkm}. All 116 sequences are used, including the eight high-resolution sequences excluded in the LoFTR protocol~\cite{sun2021loftr}. For methods that require detector-descriptors (\emph{i.e.}, alternative learnable steps and middle-end sparse matchers), images are resized so that the shortest side is $480$ pixels, up to $2048$ keypoints are extracted per image, and the RANSAC inlier threshold is $0.5$ pixels. For end-to-end semi-dense/dense matchers, image resizing varies across methods. Most resize the longest side to 640 pixels and pad the images to squares. Exceptions include LoFTR~\cite{sun2021loftr}, which resizes the shortest side to 480 pixels, DKM~\cite{edstedt2023dkm}, which resizes images to $880\times660$, and RoMa~\cite{edstedt2024roma}, which uses $672\times672$. PMatch, ArgMatch, and DenseAffine resize the longest side to $1200$ pixels, and MASt3R uses $512$-pixel inputs with reciprocal nearest-neighbor matching. The RANSAC inlier threshold for these methods is $3$ pixels.

\subsubsection{Matching Accuracy Assessment}
Matching accuracy is evaluated on MegaDepth~\cite{li2018megadepth} following the protocol of LoFTR~\cite{sun2021loftr}. All images are resized to $672\times672$, and dense flow is estimated with several end-to-end dense matchers. \textbf{PCK} is computed only within regions with valid GT depth. PMatch and ArgMatch use the same input size.

\subsubsection{Visual Localization}
We use the open-source hierarchical localization framework HLoc~\cite{sarlin2019coarse}, following the protocols of~\cite{lindenberger2023lightglue,sun2021loftr}. For Aachen Day-Night v1.1, we first triangulate a sparse 3D point cloud from the $6697$ daytime reference images with known poses and intrinsics using COLMAP~\cite{schonberger2016structure}. For each of the $824$ daytime and $191$ nighttime queries, we retrieve the top-$50$ reference images with NetVLAD~\cite{arandjelovic2016netvlad}, match the query against each of them, and estimate the camera pose with RANSAC and a Perspective-n-Point (PnP) solver. The RANSAC inlier threshold is set to 12 pixels. Input images are resized so that their longest side is $1024$ pixels, except for DKM~\cite{edstedt2023dkm} and RoMa~\cite{edstedt2024roma}, which follow their original settings with resolutions of $880\times660$ and $672\times672$, respectively. For InLoc, where the sparse 3D point cloud is provided, we retrieve the top-$40$ reference images with NetVLAD, and the subsequent steps are identical to those for Aachen Day-Night v1.1, with a RANSAC inlier threshold of 48 pixels. Input images are resized to $1600$ pixels on the long side for detector-descriptors, outlier filters, and sparse matchers, and to $800$ pixels for semi-dense matchers and DKM, while RoMa keeps its original $672\times672$ resolution. For both benchmarks, we extract up to $4096$ keypoints per image when using detector-descriptors. The pipelines and evaluation settings follow the online visual localization benchmark\footnote{\url{https://www.visuallocalization.net/benchmark}}.

\subsubsection{3D Reconstruction}
Similar to~\cite{lindenberger2021pixsfm}, we evaluate 3D reconstruction with known camera intrinsics and poses on the 12 scenes of the ETH3D high-resolution multi-view test split~\cite{schops2017multi}. For each scene, image pairs are formed from the top-$20$ covisible neighbors of each image according to the provided reference model. The correspondences of each evaluated matcher are imported into a COLMAP~\cite{schonberger2016structure} database and verified against the epipolar geometry given by the known poses. The 3D points of the reference model are then discarded, and new 3D points are triangulated from the verified matches with all camera intrinsics and poses kept fixed. Since the ground truth of the test split is hidden, we report both the COLMAP statistics computed locally and the official metrics returned by the ETH3D benchmark server\footnote{\url{https://www.eth3d.net/high_res_multi_view?set=test}}.
For sparse detector-descriptor pipelines, images are resized to a maximum side length of $1600$ pixels, and SIFT uses RootSIFT descriptors. SuperPoint and ALIKED use their standard pre-trained models without an explicit keypoint cap, while XFeat keeps up to $4096$ keypoints. Detector-descriptor baselines and SIFT-based outlier filters use one-way nearest-neighbor (NN) candidate matching. The DiffGlue row uses the official SuperPoint+DiffGlue setting with up to $2048$ SuperPoint keypoints. For semi-dense/dense matchers, images are resized to a maximum side length of $1024$ pixels, and PDC-Net+, GLU-Net, DKM, RoMa, MASt3R, ArgMatch, DenseAffine, and PMatch sample up to $10000$ correspondences per image pair. No multi-view stereo densification is applied.

\subsection{Quantitative Results}

	\subsubsection{Relative Pose Estimation}
	\label{subsec:pose}

\begin{table}[!t]
	\footnotesize
	\centering
	\caption{
		Quantitative performance of different learning-based image matching methods on relative pose estimation.
		Values are \textbf{AUC} percentages at $5^\circ$, $10^\circ$, and $20^\circ$. U-Match+* uses an inlier prediction threshold of $2.0$ instead of the default $0$.
		The \capfirst{best}, \capsecond{second-best}, and \capthird{third-best} results in each column are shaded.
	}
	\label{table:pose}
	\setlength{\tabcolsep}{0.8mm}
	\setlength{\aboverulesep}{0pt}
	\setlength{\belowrulesep}{0pt}
	\renewcommand{\arraystretch}{1.1}
	\adjustbox{max width=\linewidth}{%
		\begin{tabular}{cccc|ccc|ccc|ccc|ccc}
			\toprule
			\multirow{2}{*}{\makecell[c]{\textbf{Detector-}\\\textbf{Descriptor}}} & \multirow{2}{*}{\textbf{Matcher}} & \multirow{2}{*}{\textbf{Filter}} & \multirow{2}{*}{\textbf{Estimator}}
			& \multicolumn{3}{c|}{\textbf{MegaDepth}~\cite{li2018megadepth}} & \multicolumn{3}{c|}{\textbf{YFCC100M}~\cite{thomee2016yfcc100m}} & \multicolumn{3}{c|}{\textbf{ScanNet}~\cite{dai2017scannet}} & \multicolumn{3}{c}{\textbf{SUN3D}~\cite{xiao2013sun3d}} \\
			\cmidrule(lr){5-7}\cmidrule(lr){8-10}\cmidrule(lr){11-13}\cmidrule(l){14-16}
			& & & & $@5^\circ$ & $@10^\circ$ & $@20^\circ$ & $@5^\circ$ & $@10^\circ$ & $@20^\circ$ & $@5^\circ$ & $@10^\circ$ & $@20^\circ$ & $@5^\circ$ & $@10^\circ$ & $@20^\circ$ \\
			\midrule
			SIFT~\cite{lowe2004distinctive} & NN & -- & RANSAC~\cite{fischler1981random} & 3.44 & 8.12 & 16.10 & 3.68 & 9.41 & 19.49 & 0.73 & 2.38 & 5.61 & 1.11 & 3.65 & 9.68 \\
			SIFT & NN & -- & NG-RANSAC~\cite{brachmann2019neural} & 21.04 & 31.31 & 42.31 & 16.47 & 28.38 & 42.03 & 3.81 & 9.14 & 16.08 & 4.22 & 10.87 & 21.39 \\
			SIFT & NN & -- & ARS-MAGSAC~\cite{wei2023adaptive} & 23.35 & 33.54 & 45.61 & 18.27 & 30.22 & 44.59 & 5.50 & 11.04 & 18.53 & 5.88 & 12.52 & 23.54 \\
			SuperPoint~\cite{detone2018superpoint} & NN & -- & RANSAC & 13.41 & 25.64 & 40.51 & 8.77 & 20.07 & 35.10 & 5.86 & 13.52 & 25.15 & 5.02 & 13.23 & 26.67 \\
			DISK~\cite{tyszkiewicz2020disk} & NN & -- & RANSAC & 19.76 & 33.98 & 48.79 & 22.74 & 41.29 & 60.12 & 2.90 & 8.42 & 17.47 & 3.92 & 11.20 & 23.26 \\
			ALIKED~\cite{zhao2023aliked} & NN & -- & RANSAC & 27.72 & 41.81 & 56.02 & 26.10 & 44.93 & 63.18 & 4.96 & 11.78 & 21.37 & 5.82 & 15.24 & 29.71 \\
			XFeat~\cite{potje2024xfeat} & NN & -- & RANSAC & 11.58 & 23.86 & 40.42 & 14.30 & 29.29 & 47.32 & 3.84 & 11.25 & 24.04 & 4.92 & 13.74 & 28.33 \\
			SuperPoint & MNN & -- & RANSAC & 30.35 & 45.95 & 59.66 & 16.50 & 31.38 & 48.26 & 9.86 & 22.49 & 37.25 & 6.37 & 16.26 & 31.15 \\
			ALIKED & MNN & -- & RANSAC & 44.62 & 59.87 & 72.35 & 32.20 & 53.13 & 70.98 & 9.73 & 22.41 & 36.68 & 7.14 & 17.92 & 33.74 \\
			\hdashline[2pt/1.5pt]
			SIFT & NN & PointCN~\cite{yi2018learning} & RANSAC & 30.22 & 44.67 & 58.10 & 28.11 & 45.35 & 61.24 & 6.02 & 13.37 & 22.96 & 5.77 & 14.33 & 27.06 \\
			SIFT & NN & OANet~\cite{zhang2019learning} & RANSAC & 33.63 & 48.57 & 61.80 & 28.76 & 47.02 & 63.99 & 6.45 & 15.66 & 26.94 & 5.42 & 13.62 & 26.11 \\
			SIFT & NN & CLNet~\cite{zhao2021progressive} & RANSAC & 40.27 & 56.43 & 70.11 & 34.00 & 53.74 & 70.61 & 6.77 & 16.65 & 28.74 & 5.27 & 13.18 & 25.29 \\
			SIFT & NN & ConvMatch+~\cite{zhang2024convmatch} & RANSAC & 38.30 & 54.70 & 68.45 & 34.48 & 53.74 & 70.26 & 8.49 & 18.93 & 31.18 & 5.73 & 14.88 & 28.49 \\
			SIFT & NN & NCMNet+~\cite{liu2024ncmnet} & RANSAC & 41.77 & 57.74 & 71.22 & 34.93 & 55.03 & 71.83 & 9.33 & 20.21 & 33.42 & 6.33 & 15.96 & 30.02 \\
			SIFT & NN & DeMatch~\cite{zhang2024dematch} & RANSAC & 38.07 & 53.78 & 67.53 & 33.91 & 52.84 & 69.20 & 7.68 & 18.14 & 30.25 & 5.80 & 14.71 & 28.02 \\
			SIFT & NN & U-Match+~\cite{li2024u} & RANSAC & 40.38 & 56.81 & 70.13 & 36.85 & 56.42 & 72.54 & 9.47 & 21.16 & 34.32 & 6.42 & 16.11 & 30.29 \\
			SIFT & NN & U-Match+* & RANSAC & 41.22 & 57.54 & 70.73 & 36.74 & 56.70 & 72.82 & 10.85 & 22.80 & 36.26 & 6.46 & 16.42 & 30.81 \\
			SIFT & NN & NCMNet+ & Weighted 8-pt~\cite{yi2018learning} & 27.19 & 43.05 & 59.43 & 27.45 & 47.53 & 65.70 & 2.27 & 6.49 & 15.44 & 2.04 & 6.36 & 15.68 \\
			SIFT & NN & U-Match+ & Weighted 8-pt & 23.87 & 37.58 & 51.33 & 36.62 & 57.97 & 74.11 & 2.97 & 8.93 & 20.26 & 2.95 & 9.41 & 22.33 \\
			XFeat & NN & U-Match+ & RANSAC & 20.89 & 37.92 & 55.03 & 18.53 & 34.72 & 52.82 & 2.97 & 8.85 & 18.39 & 1.98 & 6.41 & 15.46 \\
			ALIKED & NN & U-Match+ & RANSAC & 39.23 & 55.38 & 68.78 & 32.15 & 52.55 & 69.91 & 3.72 & 10.35 & 21.67 & 5.89 & 13.30 & 24.18 \\
			\hdashline[2pt/1.5pt]
			SuperPoint & \multicolumn{2}{c}{SuperGlue~\cite{sarlin2020superglue}} & RANSAC & 48.44 & 65.70 & 78.98 & 39.47 & 59.75 & 75.91 & 15.39 & 32.38 & 49.01 & 7.18 & 17.89 & 33.42 \\
			SuperPoint & \multicolumn{2}{c}{SGMNet~\cite{chen2021learning}} & RANSAC & 39.95 & 58.49 & 73.24 & 34.22 & 54.50 & 71.57 & 15.82 & 31.67 & 49.68 & 6.79 & 17.20 & 32.34 \\
			SuperPoint & \multicolumn{2}{c}{ResMatch~\cite{deng2024resmatch}} & RANSAC & 43.86 & 61.37 & 75.41 & 35.17 & 55.81 & 73.04 & 16.23 & 32.96 & 49.71 & 7.10 & 17.79 & 33.28 \\
			SuperPoint & \multicolumn{2}{c}{IMP~\cite{xue2023imp}} & RANSAC & 44.94 & 62.45 & 76.44 & 38.68 & 59.16 & 75.38 & 15.16 & 31.84 & 48.42 & 6.76 & 16.99 & 32.21 \\
			SuperPoint & \multicolumn{2}{c}{LightGlue~\cite{lindenberger2023lightglue}} & RANSAC & 50.51 & 68.01 & 80.65 & 38.99 & 59.52 & 75.77 & 14.76 & 31.21 & 47.47 & 6.80 & 17.47 & 32.86 \\
			SuperPoint & \multicolumn{2}{c}{SemaGlue~\cite{zhang2025matching}} & RANSAC & 49.41 & 66.86 & 79.97 & 40.10 & 60.35 & 76.24 & 15.10 & 31.25 & 48.36 & 6.75 & 17.12 & 32.18 \\
			SuperPoint & \multicolumn{2}{c}{DiffGlue~\cite{zhang2024diffglue}} & RANSAC & 50.21 & 67.30 & 80.05 & 39.94 & 60.33 & 76.30 & 15.36 & 31.94 & 48.80 & 6.64 & 17.15 & 32.39 \\
			ALIKED & \multicolumn{2}{c}{LightGlue} & RANSAC & 50.83 & 67.93 & 80.55 & 44.22 & 63.90 & 78.86 & 16.03 & 32.39 & 49.28 & 7.65 & 19.02 & 34.97 \\
			ALIKED & \multicolumn{2}{c}{DiffGlue} & RANSAC & 51.31 & 67.99 & 80.46 & 44.55 & 64.39 & 79.23 & 15.41 & 32.38 & 49.60 & 7.45 & 18.73 & 34.64 \\
			ALIKED & \multicolumn{2}{c}{SemaGlue} & RANSAC & 51.55 & 68.66 & 80.82 & 44.65 & 64.51 & 79.31 & 15.70 & 32.08 & 48.91 & 7.73 & 18.91 & 34.50 \\
			\hdashline[2pt/1.5pt]
			\multicolumn{3}{c}{LoFTR~\cite{sun2021loftr}} & RANSAC & 52.89 & 69.23 & 81.30 & 39.80 & 60.03 & 76.07 & 16.82 & 33.37 & 49.95 & 8.49 & 20.85 & 37.99 \\
			\multicolumn{3}{c}{QuadTree~\cite{tang2022quadtree}} & RANSAC & 51.43 & 68.16 & 80.64 & 37.57 & 58.21 & 74.71 & 20.02 & 38.61 & 55.86 & 8.56 & 21.03 & 38.38 \\
			\multicolumn{3}{c}{MatchFormer~\cite{wang2022matchformer}} & RANSAC & 54.19 & 70.53 & 82.52 & 39.35 & 59.95 & 76.21 & 17.44 & 34.83 & 51.14 & 8.34 & 20.71 & 37.96 \\
			\multicolumn{3}{c}{TopicFM+~\cite{giang2024topicfm+}} & RANSAC & 53.15 & 68.89 & 82.16 & 39.57 & 60.05 & 76.37 & 17.87 & 36.52 & 53.99 & 8.85 & 21.39 & 38.71 \\
			\multicolumn{3}{c}{ASpanFormer~\cite{chen2022aspanformer}} & RANSAC & 55.36 & 71.71 & 83.33 & 37.42 & 58.08 & 74.67 & 20.82 & 39.51 & 57.23 & 8.74 & 21.49 & 38.89 \\
			\multicolumn{3}{c}{ELoFTR~\cite{wang2024efficient}} & RANSAC & 56.38 & 72.18 & 83.48 & 41.56 & 61.89 & 77.30 & 18.59 & 36.93 & 54.18 & 8.63 & 21.20 & 38.33 \\
			\multicolumn{3}{c}{JamMa~\cite{lu2025jamma}} & RANSAC & 56.02 & 71.25 & 82.15 & 33.32 & 53.07 & 69.92 & 11.46 & 25.32 & 40.46 & 7.89 & 19.57 & 36.29 \\
			\multicolumn{3}{c}{PDC-Net+~\cite{truong2023pdc}} & RANSAC & 51.53 & 67.27 & 78.58 & 36.47 & 56.91 & 73.67 & 19.98 & 39.15 & 56.86 & 8.43 & 21.02 & 38.32 \\
			\multicolumn{3}{c}{DKM~\cite{edstedt2023dkm}} & RANSAC & \tabthird{60.89} & 75.23 & 85.27 & 44.27 & 63.96 & 78.56 & 24.15 & 44.34 & 61.67 & 9.07 & 22.11 & 39.41 \\
			\multicolumn{3}{c}{RoMa~\cite{edstedt2024roma}} & RANSAC & \tabfirst{62.76} & \tabfirst{77.00} & \tabfirst{86.69} & 44.25 & 64.07 & 78.96 & 25.97 & 46.52 & 64.48 & \tabthird{9.53} & 22.84 & 40.56 \\
			\multicolumn{3}{c}{PMatch~\cite{zhu2023pmatch}} & RANSAC & \tabsecond{61.23} & \tabthird{75.61} & \tabsecond{85.70} & \tabsecond{48.11} & \tabthird{66.31} & \tabthird{79.88} & 28.37 & 49.55 & 67.35 & \tabsecond{9.84} & \tabsecond{23.36} & \tabsecond{41.16} \\
			\multicolumn{3}{c}{ArgMatch~\cite{deng2025argmatch}} & RANSAC & \tabsecond{61.23} & \tabsecond{75.63} & \tabthird{85.59} & \tabfirst{50.32} & \tabfirst{67.96} & \tabfirst{80.91} & 28.35 & 48.81 & 65.90 & 9.30 & 22.38 & 39.73 \\
			\multicolumn{3}{c}{DenseAffine~\cite{sun2025learning}} & RANSAC & 57.63 & 73.04 & 83.79 & 47.25 & 65.56 & 79.44 & \tabthird{29.79} & \tabthird{50.69} & \tabthird{68.12} & \tabsecond{9.84} & \tabthird{23.34} & \tabthird{40.99} \\
			\multicolumn{3}{c}{MASt3R~\cite{leroy2024grounding}} & RANSAC & 45.35 & 64.66 & 79.37 & \tabthird{47.62} & \tabsecond{67.04} & \tabsecond{80.87} & \tabsecond{31.67} & \tabsecond{54.79} & \tabsecond{72.71} & \tabfirst{10.41} & \tabfirst{24.90} & \tabfirst{43.30} \\
			\hdashline[2pt/1.5pt]
			\multicolumn{4}{c|}{SRPose~\cite{yin2025srpose}} & 20.13 & 42.06 & 64.32 & 22.81 & 48.52 & 70.61 & 3.61 & 13.62 & 33.22 & 4.86 & 15.30 & 33.15 \\
			\multicolumn{4}{c|}{Reloc3R~\cite{dong2025reloc3r}} & 48.92 & 67.35 & 80.93 & 46.04 & 64.46 & 78.26 & \tabfirst{33.60} & \tabfirst{57.51} & \tabfirst{75.04} & 7.75 & 21.13 & 40.01 \\
			\bottomrule
		\end{tabular}%
	}
\end{table}

	We evaluate relative pose estimation on two outdoor datasets (MegaDepth-1500~\cite{li2018megadepth} and YFCC100M~\cite{thomee2016yfcc100m}) and two indoor datasets (ScanNet~\cite{dai2017scannet} and SUN3D~\cite{xiao2013sun3d}).
	We evaluate representative alternative learnable steps (detector-descriptors, outlier filters, and geometric estimators) and merged learnable modules (middle-end sparse matchers and semi-dense/dense matchers), together with two pose regressors, SRPose~\cite{yin2025srpose} and Reloc3R~\cite{dong2025reloc3r}. \textbf{AUC} at $5^\circ$, $10^\circ$, and $20^\circ$ is reported in Table~\ref{table:pose}.
	Outlier filters are also evaluated with the weighted eight-point solver~\cite{yi2018learning}, which they use during training to predict transformation models and compute their geometric loss. Middle-end sparse matchers and semi-dense/dense matchers use their outdoor models on all datasets, because most of them are not trained on indoor scenes; their indoor results therefore measure cross-scene generalization. MASt3R uses its released general checkpoint.
	Replacing a single step already yields substantial gains. ALIKED+MNN reaches $44.62\%@5^\circ$ on MegaDepth and $7.14\%@5^\circ$ on SUN3D, and SIFT+U-Match+ achieves $36.85\%@5^\circ$ on YFCC100M, while U-Match+* (with the inlier prediction threshold changed from the default $0$ to $2.0$) reaches $10.85\%@5^\circ$ on ScanNet. However, U-Match+, which is trained on SIFT matches, improves the MegaDepth results of XFeat and ALIKED NN matches but lowers their ScanNet \textbf{AUC}$@5^\circ$ and $@10^\circ$, \emph{e.g.}, from $4.96\%$ to $3.72\%@5^\circ$ for ALIKED, so learned filters do not transfer reliably across detector-descriptors and scenes.
	Merged modules perform even better. ALIKED+SemaGlue attains $51.55\%@5^\circ$ on MegaDepth, and dense matchers lead on this dataset, with RoMa achieving $62.76\%@5^\circ$, followed by PMatch and ArgMatch ($61.23\%@5^\circ$). ArgMatch also achieves the best YFCC100M result ($50.32\%@5^\circ$). On the cross-scene ScanNet dataset, RoMa reaches $25.97\%@5^\circ$, while MASt3R obtains the best result among semi-dense/dense matchers ($31.67\%@5^\circ$) and the pose regressor Reloc3R the best result overall ($33.60\%@5^\circ$). Both are trained on large-scale data that include indoor scenes. In contrast, SRPose, evaluated with its MegaDepth checkpoint, is considerably less accurate than learned sparse, semi-dense, and dense matchers (\emph{e.g.}, $20.13\%@5^\circ$ on MegaDepth).

	\subsubsection{IMC2021 Official Benchmark}
	\label{subsec:imc2021}

\begin{table}[!t]
	\footnotesize
	\centering
	\caption{
		Official IMC2021 test split relative pose results.
		We report \textbf{mAA}$@10^\circ$ on PhotoTourism and PragueParks, model size, and average pipeline time in milliseconds per image pair.
		The mean column is the macro average of the two track-level \textbf{mAA} scores, and all \textbf{mAA} values are percentages.
		Model size denotes checkpoint size in MB; ``--'' denotes non-learned or non-applicable components.
		The \capfirst{best}, \capsecond{second-best}, and \capthird{third-best} results in each column are shaded.
	}
	\label{table:imc2021}
	\setlength{\tabcolsep}{1.2mm}
	\setlength{\aboverulesep}{0pt}
	\setlength{\belowrulesep}{0pt}
	\renewcommand{\arraystretch}{1.1}
	\adjustbox{max width=\linewidth}{%
		\begin{tabular}{cccc|ccc|c|c}
			\toprule
			\multirow{2}{*}{\makecell[c]{\textbf{Detector-}\\\textbf{Descriptor}}} & \multirow{2}{*}{\textbf{Matcher}} & \multirow{2}{*}{\textbf{Filter}} & \multirow{2}{*}{\textbf{Estimator}}
			& \multicolumn{3}{c|}{\textbf{mAA}$@10^\circ$ (\%) $\uparrow$} & \multirow{2}{*}{\makecell[c]{\textbf{Time} $\downarrow$\\(ms/pair)}} & \multirow{2}{*}{\makecell[c]{\textbf{Size} $\downarrow$\\(MB)}} \\
			\cmidrule(lr){5-7}
			& & & & PhotoTourism & PragueParks & Mean & & \\
			\midrule
			SIFT~\cite{lowe2004distinctive} & NN & -- & RANSAC & 5.86 & 2.45 & 4.16 & 133.5 & -- \\
			SIFT & NN & -- & NG-RANSAC~\cite{brachmann2019neural} & 22.83 & 24.33 & 23.58 & \tabfirst{36.4} & \tabfirst{1.6} \\
			SuperPoint~\cite{detone2018superpoint} & NN & -- & RANSAC & 11.36 & 16.92 & 14.14 & 98.2 & \tabthird{5.0} \\
			ALIKED~\cite{zhao2023aliked} & NN & -- & RANSAC & 27.99 & 23.37 & 25.68 & 116.6 & \tabsecond{2.6} \\
			XFeat~\cite{potje2024xfeat} & NN & -- & RANSAC & 14.36 & 10.28 & 12.32 & 129.6 & 6.0 \\
			\hdashline[2pt/1.5pt]
			SIFT & NN & PointCN~\cite{yi2018learning} & RANSAC & 22.92 & 15.34 & 19.13 & 137.9 & 23.0 \\
			SIFT & NN & OANet~\cite{zhang2019learning} & RANSAC & 29.15 & 20.96 & 25.05 & \tabsecond{39.9} & 28.5 \\
			SIFT & NN & ConvMatch+~\cite{zhang2024convmatch} & RANSAC & 31.65 & 26.69 & 29.17 & 63.7 & 161.3 \\
			SIFT & NN & NCMNet+~\cite{liu2024ncmnet} & RANSAC & 31.58 & 29.59 & 30.58 & 72.3 & 59.1 \\
			SIFT & NN & U-Match+~\cite{li2024u} & RANSAC & 30.73 & 25.30 & 28.01 & 65.6 & 89.9 \\
			SIFT & NN & NCMNet+ & Weighted 8-pt & 15.59 & 5.32 & 10.46 & 76.3 & 59.1 \\
			SIFT & NN & U-Match+ & Weighted 8-pt & 17.29 & 8.99 & 13.14 & 62.1 & 89.9 \\
			XFeat & NN & U-Match+ & RANSAC & 20.42 & 21.62 & 21.02 & 53.1 & 95.8 \\
			ALIKED & NN & U-Match+ & RANSAC & 36.18 & 43.76 & 39.97 & 53.8 & 92.5 \\
			\hdashline[2pt/1.5pt]
			SuperPoint & \multicolumn{2}{c}{SuperGlue~\cite{sarlin2020superglue}} & RANSAC & 39.71 & 59.46 & 49.59 & 107.5 & 51.0 \\
			SuperPoint & \multicolumn{2}{c}{LightGlue~\cite{lindenberger2023lightglue}} & RANSAC & 38.67 & 59.09 & 48.88 & 62.7 & 50.3 \\
			SuperPoint & \multicolumn{2}{c}{DiffGlue~\cite{zhang2024diffglue}} & RANSAC & 40.44 & 61.53 & 50.99 & 96.6 & 255.9 \\
			ALIKED & \multicolumn{2}{c}{LightGlue} & RANSAC & 45.99 & 65.83 & 55.91 & \tabthird{45.8} & 48.0 \\
			ALIKED & \multicolumn{2}{c}{DiffGlue} & RANSAC & 47.43 & 66.93 & 57.18 & 80.8 & 251.5 \\
			\hdashline[2pt/1.5pt]
			\multicolumn{3}{c}{LoFTR~\cite{sun2021loftr}} & RANSAC & 34.95 & 63.38 & 49.16 & 294.2 & 44.2 \\
			\multicolumn{3}{c}{MatchFormer~\cite{wang2022matchformer}} & RANSAC & 50.76 & 56.75 & 53.76 & 297.0 & 77.4 \\
			\multicolumn{3}{c}{ELoFTR~\cite{wang2024efficient}} & RANSAC & 58.51 & 66.47 & 62.49 & 276.9 & 183.9 \\
			\multicolumn{3}{c}{PDC-Net+~\cite{truong2023pdc}} & RANSAC & 40.80 & 47.65 & 44.23 & 212.0 & 211.7 \\
			\multicolumn{3}{c}{DKM~\cite{edstedt2023dkm}} & RANSAC & 65.05 & \tabsecond{77.99} & \tabthird{71.52} & 723.6 & 276.4 \\
			\multicolumn{3}{c}{RoMa~\cite{edstedt2024roma}} & RANSAC & \tabsecond{67.26} & \tabfirst{79.82} & \tabfirst{73.54} & 985.7 & 1586.2 \\
			\multicolumn{3}{c}{MASt3R~\cite{leroy2024grounding}} & RANSAC & 62.61 & \tabthird{75.90} & 69.26 & 276.9 & 2627.3 \\
			\multicolumn{3}{c}{ArgMatch~\cite{deng2025argmatch}} & RANSAC & \tabfirst{67.43} & 75.68 & \tabsecond{71.56} & 257.5 & 147.5 \\
			\multicolumn{3}{c}{DenseAffine~\cite{sun2025learning}} & RANSAC & 63.68 & 64.11 & 63.90 & 758.3 & 391.1 \\
			\multicolumn{3}{c}{PMatch~\cite{zhu2023pmatch}} & RANSAC & 64.21 & 75.14 & 69.67 & 735.9 & 162.6 \\
			\hdashline[2pt/1.5pt]
			\multicolumn{4}{c|}{SRPose~\cite{yin2025srpose}} & 39.49 & 14.67 & 27.08 & 49.0 & 99.8 \\
			\multicolumn{4}{c|}{Reloc3R~\cite{dong2025reloc3r}} & \tabthird{66.83} & 54.11 & 60.47 & 126.6 & 3254.2 \\
			\bottomrule
		\end{tabular}%
	}
\end{table}

	To complement the MegaDepth and YFCC100M protocols, we also evaluate relative pose estimation on the official IMC2021 PhotoTourism and PragueParks stereo tracks~\cite{jin2021image}. Table~\ref{table:imc2021} reports \textbf{mAA}$@10^\circ$ on each track, their mean, model size, and average runtime per image pair.
	Plain NN baselines lag behind; for example, ALIKED+NN reaches only $25.68\%$ mean \textbf{mAA}$@10^\circ$. Learned outlier filters improve the SIFT+NN pipeline when followed by RANSAC, with NCMNet+ and ConvMatch+ reaching $30.58\%$ and $29.17\%$, but they are much less effective with the weighted eight-point solver. Learned sparse matchers are more competitive, with ALIKED+DiffGlue and ALIKED+LightGlue reaching $57.18\%$ and $55.91\%$, respectively.
	Dense matchers lead the ranking. RoMa achieves the best mean score ($73.54\%$), followed by ArgMatch ($71.56\%$) and DKM ($71.52\%$). ArgMatch is the best on PhotoTourism ($67.43\%$) and RoMa on PragueParks ($79.82\%$), while ArgMatch runs about four times faster than RoMa. PMatch and MASt3R reach $69.67\%$ and $69.26\%$, whereas DenseAffine is less accurate than DKM and PMatch at a similar runtime ($758.3$ ms/pair). Among pose regressors, SRPose is fast but inaccurate, and Reloc3R performs well on PhotoTourism ($66.83\%$) but drops on PragueParks ($54.11\%$).

	\subsubsection{Homography Estimation}
	\label{subsec:homography}

\begin{table}[!t]
	\footnotesize
	\centering
	\caption{
		Quantitative performance of different learning-based image matching methods on homography estimation.
		\textbf{Acc.} and \textbf{AUC} values are percentages at $3$, $5$, and $10$ px.
		The default estimator is RANSAC~\cite{fischler1981random}.
		The \capfirst{best}, \capsecond{second-best}, and \capthird{third-best} results in each column are shaded.
	}
	\label{table:homo}
	\setlength{\tabcolsep}{0.6mm}
	\setlength{\aboverulesep}{0pt}
	\setlength{\belowrulesep}{0pt}
	\renewcommand{\arraystretch}{1.1}
	\adjustbox{max width=\linewidth}{%
		\begin{tabular}{cc|ccc|ccc}
			\toprule
			\multirow{3}{*}{\makecell[c]{\textbf{Detector-}\\\textbf{Descriptor}}} & \multirow{3}{*}{\makecell[c]{\textbf{Matcher+}\\\textbf{Filter}}} & \multicolumn{6}{c}{\textbf{HPatches}~\cite{balntas2017hpatches}} \\
			\cmidrule(l){3-8}
			& & \multicolumn{3}{c|}{\textbf{Acc.} (\%)} & \multicolumn{3}{c}{\textbf{AUC} (\%)} \\
			\cmidrule(lr){3-5}\cmidrule(l){6-8}
			& & $@3$px & $@5$px & $@10$px & $@3$px & $@5$px & $@10$px \\
			\midrule
			SIFT~\cite{lowe2004distinctive} & NN & 49.14 & 58.79 & 69.83 & 30.93 & 40.50 & 52.91 \\
			SuperPoint~\cite{detone2018superpoint} & NN & 42.93 & 57.59 & 75.34 & 30.39 & 38.45 & 53.53 \\
			DISK~\cite{tyszkiewicz2020disk} & NN & 53.28 & 67.41 & 83.28 & 38.89 & 47.63 & 62.27 \\
			ALIKED~\cite{zhao2023aliked} & NN & 56.90 & 72.41 & 84.66 & 33.78 & 46.69 & 63.28 \\
			XFeat~\cite{potje2024xfeat} & NN & 41.37 & 48.45 & 61.72 & 33.04 & 37.86 & 46.75 \\
			SuperPoint & MNN & 54.14 & 68.28 & 82.93 & 38.72 & 48.18 & 62.67 \\
			ALIKED & MNN & 63.79 & 77.41 & 89.14 & 39.42 & 52.24 & 68.31 \\
			\hdashline[2pt/1.5pt]
			SIFT & NN+PointCN~\cite{yi2018learning} & 62.07 & 74.65 & 85.69 & 39.12 & 50.97 & 66.18 \\
			SIFT & NN+OANet~\cite{zhang2019learning} & 61.55 & 75.00 & 85.69 & 38.62 & 50.72 & 66.13 \\
			SIFT & NN+CLNet~\cite{zhao2021progressive} & 63.97 & 75.52 & 87.58 & 40.44 & 52.49 & 67.66 \\
			SIFT & NN+ConvMatch+~\cite{zhang2024convmatch} & 61.90 & 75.17 & 86.90 & 39.43 & 51.32 & 66.55 \\
			SIFT & NN+NCMNet+~\cite{liu2024ncmnet} & 65.86 & 77.76 & 87.59 & 41.49 & 53.75 & 68.70 \\
			SIFT & NN+DeMatch~\cite{zhang2024dematch} & 61.55 & 74.83 & 86.55 & 39.01 & 50.95 & 66.47 \\
			SIFT & NN+U-Match+~\cite{li2024u} & 62.76 & 76.38 & 86.90 & 39.57 & 52.06 & 67.35 \\
			XFeat & NN+U-Match+ & 42.59 & 48.10 & 57.41 & 36.67 & 40.33 & 46.53 \\
			ALIKED & NN+U-Match+ & 43.45 & 59.48 & 78.10 & 22.54 & 34.56 & 52.30 \\
			\hdashline[2pt/1.5pt]
			SuperPoint & SuperGlue~\cite{sarlin2020superglue} & 64.83 & 78.28 & 90.34 & 44.83 & 55.84 & 70.87 \\
			SuperPoint & SGMNet~\cite{chen2021learning} & 59.83 & 74.66 & 87.24 & 42.28 & 52.62 & 67.09 \\
			SuperPoint & ResMatch~\cite{deng2024resmatch} & 62.76 & 76.90 & 90.00 & 43.43 & 54.35 & 69.55 \\
			SuperPoint & IMP~\cite{xue2023imp} & 63.97 & 78.45 & 89.83 & 43.08 & 54.50 & 69.98 \\
			SuperPoint & LightGlue~\cite{lindenberger2023lightglue} & 65.34 & 78.45 & 88.45 & 45.27 & 56.27 & 70.22 \\
			SuperPoint & DiffGlue~\cite{zhang2024diffglue} & 63.28 & 78.45 & 89.14 & 44.14 & 55.24 & 69.97 \\
			ALIKED & LightGlue & 65.86 & 78.79 & 90.52 & 39.79 & 53.52 & 69.99 \\
			ALIKED & DiffGlue & 65.69 & 79.83 & 90.34 & 40.05 & 53.57 & 70.43 \\
			\hdashline[2pt/1.5pt]
			\multicolumn{2}{c|}{LoFTR~\cite{sun2021loftr}} & 72.58 & 83.62 & 90.86 & 50.03 & 61.63 & 74.79 \\
			\multicolumn{2}{c|}{QuadTree~\cite{tang2022quadtree}} & 76.90 & 85.69 & 92.41 & 52.73 & 64.59 & 77.15 \\
			\multicolumn{2}{c|}{MatchFormer~\cite{wang2022matchformer}} & 73.97 & 85.17 & 91.55 & 51.40 & 63.00 & 75.96 \\
			\multicolumn{2}{c|}{TopicFM+~\cite{giang2024topicfm+}} & 75.34 & 88.62 & 93.45 & 50.02 & 63.35 & 77.76 \\
			\multicolumn{2}{c|}{ASpanFormer~\cite{chen2022aspanformer}} & 77.93 & 86.38 & 91.55 & 53.27 & 65.18 & 77.65 \\
			\multicolumn{2}{c|}{ELoFTR~\cite{wang2024efficient}} & 77.24 & 85.34 & 92.07 & 54.77 & 65.64 & 77.70 \\
			\multicolumn{2}{c|}{JamMa~\cite{lu2025jamma}} & 72.76 & 81.03 & 87.93 & 49.99 & 61.01 & 73.11 \\
			\multicolumn{2}{c|}{DKM~\cite{edstedt2023dkm}} & \tabfirst{83.62} & \tabthird{90.52} & \tabthird{94.83} & \tabsecond{59.54} & \tabsecond{70.71} & \tabthird{81.75} \\
			\multicolumn{2}{c|}{RoMa~\cite{edstedt2024roma}} & 82.76 & \tabfirst{91.38} & \tabsecond{95.52} & \tabfirst{59.97} & \tabfirst{71.25} & \tabfirst{82.39} \\
			\multicolumn{2}{c|}{PMatch~\cite{zhu2023pmatch}} & \tabthird{83.10} & 90.17 & 94.66 & 58.27 & 69.86 & 81.49 \\
			\multicolumn{2}{c|}{ArgMatch~\cite{deng2025argmatch}} & 81.38 & 89.48 & 94.48 & 59.09 & 70.04 & 81.32 \\
			\multicolumn{2}{c|}{DenseAffine~\cite{sun2025learning}} & \tabsecond{83.28} & \tabsecond{91.21} & \tabfirst{95.69} & \tabthird{59.27} & \tabthird{70.70} & \tabsecond{82.33} \\
			\multicolumn{2}{c|}{MASt3R~\cite{leroy2024grounding}} & 80.52 & 88.79 & 94.31 & 57.07 & 68.44 & 80.33 \\
			\bottomrule
		\end{tabular}%
	}
\end{table}

	We evaluate homography estimation on HPatches~\cite{balntas2017hpatches}, reporting \textbf{Acc.} at $3$, $5$, and $10$ pixels and the corresponding \textbf{AUC} in Table~\ref{table:homo}. The evaluated methods are largely the same as in the relative pose experiments. Since the evaluated learnable RANSAC variants are trained for epipolar geometry, only RANSAC~\cite{fischler1981random} is used as the estimator.
	Even single-step replacements yield marked gains; for instance, SIFT+NCMNet+ achieves $65.86\%$ \textbf{Acc.}$@3$px and $41.49\%$ \textbf{AUC}$@3$px. Merged modules deliver the strongest results, with DKM reaching $83.62\%$ \textbf{Acc.}$@3$px and RoMa $59.97\%$ \textbf{AUC}$@3$px. PMatch, ArgMatch, and DenseAffine perform comparably, while MASt3R is slightly weaker on this planar benchmark.

	\subsubsection{Matching Accuracy Assessment}
	\label{subsec:flow}

\begin{table}[!t]
	\footnotesize
	\centering
	\caption{
		Matching accuracy of different dense matchers.
		Values are \textbf{PCK} percentages at the listed pixel thresholds.
		The \capfirst{best}, \capsecond{second-best}, and \capthird{third-best} results in each column are shaded.
	}
	\label{table:flow}
	\setlength{\tabcolsep}{2.2mm}
	\setlength{\aboverulesep}{0pt}
	\setlength{\belowrulesep}{0pt}
	\renewcommand{\arraystretch}{1.1}
	{%
		\begin{tabular}{c|cccc}
			\toprule
			\multirow{2}{*}{\textbf{Method}} & \multicolumn{4}{c}{\textbf{PCK} (\%)} \\
			\cmidrule(l){2-5}
			& $@0.5$px & $@1$px & $@3$px & $@5$px \\
			\midrule
			PDC-Net+~\cite{truong2023pdc} & \tabthird{33.62} & 60.38 & 83.90 & 87.49 \\
			DKM~\cite{edstedt2023dkm} & \tabsecond{56.21} & \tabsecond{79.83} & 94.40 & 96.01 \\
			RoMa~\cite{edstedt2024roma} & \tabfirst{58.68} & \tabfirst{82.32} & \tabsecond{96.28} & \tabsecond{97.74} \\
			PMatch~\cite{zhu2023pmatch} & 23.07 & \tabthird{69.31} & \tabthird{95.48} & \tabthird{97.04} \\
			ArgMatch~\cite{deng2025argmatch} & 23.68 & 69.10 & \tabfirst{97.17} & \tabfirst{98.61} \\
			\bottomrule
		\end{tabular}%
	}
\end{table}

	Both relative pose estimation and homography estimation show the strong performance of dense matchers, which predict dense warping maps.
	We therefore assess the matching accuracy (\textbf{PCK}) of these warping maps at $0.5$, $1$, $3$, and $5$ pixels on MegaDepth~\cite{li2018megadepth} in Table~\ref{table:flow}. RoMa achieves the best results at $0.5$ and $1$ pixels, while ArgMatch is slightly better at $3$ and $5$ pixels.

	\subsubsection{Visual Localization}
	\label{subsec:localization}

\begin{table}[!t]
	\footnotesize
	\centering
	\caption{
		Quantitative performance of different learning-based image matching methods on visual localization.
		Values are percentages at the listed distance-orientation thresholds.
		The \capfirst{best}, \capsecond{second-best}, and \capthird{third-best} results in each column are shaded.
	}
	\label{tab:visloc_perf}
	\label{table:hloc}
	\setlength{\tabcolsep}{0.8mm}
	\setlength{\aboverulesep}{0pt}
	\setlength{\belowrulesep}{0pt}
	\renewcommand{\arraystretch}{1.1}
	\adjustbox{max width=\linewidth}{%
		\begin{tabular}{cc|ccc|ccc|ccc|ccc}
			\toprule
			\multirow{3}{*}{\makecell[c]{\textbf{Detector-}\\\textbf{Descriptor}}} & \multirow{3}{*}{\makecell[c]{\textbf{Matcher+}\\\textbf{Filter}}} & \multicolumn{6}{c|}{\textbf{Aachen Day-Night v1.1}~\cite{sattler2018benchmarking}} & \multicolumn{6}{c}{\textbf{InLoc}~\cite{taira2018inloc}} \\
			\cmidrule(lr){3-8}\cmidrule(l){9-14}
			& & \multicolumn{3}{c|}{Day} & \multicolumn{3}{c|}{Night} & \multicolumn{3}{c|}{DUC1} & \multicolumn{3}{c}{DUC2} \\
			\cmidrule(lr){3-5}\cmidrule(lr){6-8}\cmidrule(lr){9-11}\cmidrule(l){12-14}
			& & {\scriptsize\makecell[c]{0.25m\\$2^\circ$}} & {\scriptsize\makecell[c]{0.5m\\$5^\circ$}} & {\scriptsize\makecell[c]{5.0m\\$10^\circ$}} & {\scriptsize\makecell[c]{0.25m\\$2^\circ$}} & {\scriptsize\makecell[c]{0.5m\\$5^\circ$}} & {\scriptsize\makecell[c]{5.0m\\$10^\circ$}} & {\scriptsize\makecell[c]{0.25m\\$10^\circ$}} & {\scriptsize\makecell[c]{0.5m\\$10^\circ$}} & {\scriptsize\makecell[c]{1.0m\\$10^\circ$}} & {\scriptsize\makecell[c]{0.25m\\$10^\circ$}} & {\scriptsize\makecell[c]{0.5m\\$10^\circ$}} & {\scriptsize\makecell[c]{1.0m\\$10^\circ$}} \\
			\midrule
			SIFT~\cite{lowe2004distinctive} & MNN & 79.1 & 85.1 & 89.6 & 24.6 & 30.9 & 41.9 & 19.7 & 31.3 & 38.4 & 11.5 & 21.4 & 22.9 \\
			SuperPoint~\cite{detone2018superpoint} & MNN & 85.6 & 91.3 & 95.5 & 61.8 & 75.9 & 89.0 & 31.3 & 49.0 & 61.6 & 29.0 & 48.9 & 58.0 \\
			DISK~\cite{tyszkiewicz2020disk} & MNN & 88.1 & 94.9 & 98.3 & \tabfirst{78.0} & 89.5 & 97.9 & 35.9 & 54.0 & 66.7 & 27.5 & 41.2 & 57.3 \\
			ALIKED~\cite{zhao2023aliked} & MNN & 87.3 & 94.1 & 97.3 & 73.8 & 88.5 & 95.8 & 36.4 & 52.5 & 64.1 & 26.7 & 44.3 & 48.9 \\
			XFeat~\cite{potje2024xfeat} & MNN & 84.0 & 90.9 & 96.2 & 66.0 & 82.2 & 93.7 & 33.8 & 51.5 & 65.7 & 38.9 & 53.4 & 62.6 \\
			\hdashline[2pt/1.5pt]
			SIFT & MNN+PointCN~\cite{yi2018learning} & 85.4 & 91.1 & 96.1 & 40.8 & 55.5 & 71.2 & 32.8 & 47.0 & 58.1 & 19.1 & 31.3 & 38.9 \\
			SIFT & MNN+OANet~\cite{zhang2019learning} & 85.9 & 91.9 & 95.9 & 36.6 & 50.3 & 67.0 & 30.3 & 47.0 & 58.6 & 19.8 & 32.8 & 40.5 \\
			SIFT & MNN+ConvMatch+~\cite{zhang2024convmatch} & 85.2 & 92.1 & 96.7 & 42.9 & 58.1 & 73.3 & 37.4 & 53.0 & 62.6 & 22.9 & 38.9 & 48.9 \\
			SIFT & MNN+NCMNet+~\cite{liu2024ncmnet} & 85.7 & 92.8 & 97.7 & 55.0 & 69.1 & 88.0 & 33.8 & 50.5 & 61.1 & 20.6 & 34.4 & 45.0 \\
			SIFT & MNN+U-Match+~\cite{li2024u} & 86.2 & 93.1 & 97.2 & 49.2 & 64.4 & 80.1 & 36.9 & 54.0 & 62.6 & 22.1 & 35.1 & 47.3 \\
			\hdashline[2pt/1.5pt]
			SuperPoint & SuperGlue~\cite{sarlin2020superglue} & \tabsecond{89.7} & \tabfirst{96.5} & \tabsecond{99.3} & 73.8 & \tabfirst{91.1} & \tabfirst{99.5} & 50.0 & 69.7 & 79.8 & 47.3 & 77.9 & 80.2 \\
			SuperPoint & SGMNet~\cite{chen2021learning} & 88.7 & 95.8 & 99.0 & 72.8 & 89.5 & \tabsecond{99.0} & 42.9 & 61.1 & 72.2 & 43.5 & 66.4 & 70.2 \\
			SuperPoint & ResMatch~\cite{deng2024resmatch} & 88.5 & 95.4 & 98.9 & 72.3 & \tabsecond{90.6} & \tabsecond{99.0} & 46.0 & 66.2 & 78.8 & 42.7 & 65.6 & 72.5 \\
			SuperPoint & LightGlue~\cite{lindenberger2023lightglue} & 89.2 & \tabfirst{96.5} & \tabsecond{99.3} & 72.3 & 89.5 & \tabsecond{99.0} & 48.0 & 68.7 & 79.8 & 44.3 & 71.0 & 75.6 \\
			SuperPoint & DiffGlue~\cite{zhang2024diffglue} & \tabthird{89.6} & \tabthird{96.1} & \tabthird{99.2} & 74.3 & \tabfirst{91.1} & \tabfirst{99.5} & 49.0 & 68.7 & 80.8 & 51.9 & 73.3 & 78.6 \\
			ALIKED & LightGlue & \tabfirst{89.9} & 95.9 & \tabfirst{99.5} & \tabthird{76.4} & \tabsecond{90.6} & \tabfirst{99.5} & 49.5 & 66.2 & 79.3 & 45.0 & 71.0 & 74.0 \\
			\hdashline[2pt/1.5pt]
			\multicolumn{2}{c|}{LoFTR~\cite{sun2021loftr}} & 88.7 & \tabthird{96.1} & 98.9 & \tabsecond{77.0} & \tabsecond{90.6} & \tabfirst{99.5} & 49.0 & 71.7 & 84.3 & 51.1 & 73.3 & 81.7 \\
			\multicolumn{2}{c|}{QuadTree~\cite{tang2022quadtree}} & 87.7 & 95.8 & 98.7 & \tabfirst{78.0} & \tabfirst{91.1} & \tabfirst{99.5} & 48.5 & \tabsecond{74.7} & 83.8 & \tabthird{55.7} & 76.3 & \tabthird{83.2} \\
			\multicolumn{2}{c|}{MatchFormer~\cite{wang2022matchformer}} & 89.4 & 96.0 & 98.8 & 75.9 & \tabsecond{90.6} & \tabfirst{99.5} & 50.0 & 73.7 & \tabthird{85.4} & \tabsecond{58.0} & \tabsecond{80.9} & \tabfirst{87.0} \\
			\multicolumn{2}{c|}{TopicFM+~\cite{giang2024topicfm+}} & 88.7 & \tabthird{96.1} & 99.0 & \tabsecond{77.0} & 89.5 & \tabsecond{99.0} & \tabthird{51.5} & \tabthird{74.2} & \tabsecond{87.4} & \tabfirst{59.5} & \tabthird{78.6} & \tabsecond{85.5} \\
			\multicolumn{2}{c|}{ASpanFormer~\cite{chen2022aspanformer}} & 89.1 & \tabsecond{96.4} & 98.9 & \tabthird{76.4} & \tabsecond{90.6} & \tabfirst{99.5} & 50.0 & \tabthird{74.2} & \tabthird{85.4} & 55.0 & 73.3 & \tabthird{83.2} \\
			\multicolumn{2}{c|}{ELoFTR~\cite{wang2024efficient}} & 88.1 & 95.1 & 98.4 & 73.8 & \tabsecond{90.6} & \tabthird{98.4} & \tabsecond{52.0} & 72.2 & 84.8 & \tabfirst{59.5} & \tabfirst{82.4} & \tabfirst{87.0} \\
			\multicolumn{2}{c|}{JamMa~\cite{lu2025jamma}} & 85.9 & 94.7 & 98.1 & 72.8 & \tabthird{90.1} & 97.9 & 47.5 & 67.2 & 78.3 & 35.9 & 53.4 & 69.5 \\
			\multicolumn{2}{c|}{DKM~\cite{edstedt2023dkm}} & 88.1 & 95.3 & 98.5 & 72.3 & \tabfirst{91.1} & 97.9 & 50.5 & 73.7 & 84.8 & 53.4 & 72.5 & 74.0 \\
			\multicolumn{2}{c|}{RoMa~\cite{edstedt2024roma}} & 88.1 & 95.6 & 98.4 & 71.7 & \tabthird{90.1} & 97.9 & \tabfirst{55.6} & \tabfirst{77.3} & \tabfirst{88.4} & \tabfirst{59.5} & \tabsecond{80.9} & \tabthird{83.2} \\
			\bottomrule
		\end{tabular}%
	}
\end{table}

	We evaluate visual localization on Aachen Day-Night~\cite{sattler2018benchmarking} and InLoc~\cite{taira2018inloc}, reporting the percentage of correctly localized queries within given distance and angular thresholds in Table~\ref{table:hloc}. MNN is the default matcher and RANSAC~\cite{fischler1981random} the estimator.
	On Aachen Day-Night, semi-dense/dense matchers are not always superior. ALIKED+LightGlue localizes $89.9\%$ of the daytime queries and $76.4\%$ of the nighttime queries within $(0.25\text{m},2^\circ)$, on par with semi-dense/dense methods. On the indoor InLoc dataset, where large viewpoint changes, lighting changes, and weak textures prevail, most semi-dense/dense matchers perform better, and RoMa localizes $55.6\%$ and $59.5\%$ of the queries within $(0.25\text{m},10^\circ)$ on DUC1 and DUC2, respectively.

	\subsubsection{3D Reconstruction}
	\label{subsec:sfm}

\begin{table}[!t]
	\footnotesize
	\centering
	\caption{
		3D reconstruction results on the ETH3D high-resolution test split with known camera intrinsics and poses.
		\# Points (K), track length, and reprojection error are local COLMAP statistics averaged over the 12 test scenes.
		Official \textbf{Acc.}, \textbf{Comp.}, and \textbf{F1} are surface metrics reported by the ETH3D server on the hidden test split; these scores are percentages.
		Detector-descriptor and outlier-filter rows use one-way NN matches.
		Semi-dense/dense rows use the settings described in Section~\ref{subsec:experiment_setting}.
		The \capfirst{best}, \capsecond{second-best}, and \capthird{third-best} results in each column are shaded.
	}
	\label{table:eth3d_sfm}
	\setlength{\tabcolsep}{0.8mm}
	\setlength{\aboverulesep}{0pt}
	\setlength{\belowrulesep}{0pt}
	\renewcommand{\arraystretch}{1.1}
	\adjustbox{max width=\linewidth}{%
		\begin{tabular}{cc|ccc|ccc|ccc|ccc}
			\toprule
			\multirow{2}{*}{\makecell[c]{\textbf{Detector-}\\\textbf{Descriptor}}} & \multirow{2}{*}{\makecell[c]{\textbf{Matcher+}\\\textbf{Filter}}} & \multicolumn{3}{c|}{\textbf{COLMAP Statistics}} & \multicolumn{3}{c|}{\textbf{Acc.} (\%) $\uparrow$} & \multicolumn{3}{c|}{\textbf{Comp.} (\%) $\uparrow$} & \multicolumn{3}{c}{\textbf{F1} (\%) $\uparrow$} \\
			\cmidrule(lr){3-5}\cmidrule(lr){6-8}\cmidrule(lr){9-11}\cmidrule(l){12-14}
			& & \makecell[c]{\# Points\\(K) $\uparrow$} & \makecell[c]{Track\\Len. $\uparrow$} & \makecell[c]{Reproj.\\Err. $\downarrow$} & $@1$cm & $@2$cm & $@5$cm & $@1$cm & $@2$cm & $@5$cm & $@1$cm & $@2$cm & $@5$cm \\
			\midrule
			SIFT~\cite{lowe2004distinctive} & NN & 5.0 & 3.876 & 1.270 & 56.62 & 71.05 & 84.10 & 0.30 & 1.43 & 7.34 & 0.60 & 2.77 & 12.98 \\
			SuperPoint~\cite{detone2018superpoint} & NN & 6.7 & 4.180 & 1.518 & 62.65 & 76.57 & 88.21 & 0.43 & 2.08 & 11.53 & 0.85 & 4.00 & 19.44 \\
			ALIKED~\cite{zhao2023aliked} & NN & 14.3 & 4.157 & 1.238 & 61.86 & 74.80 & 85.96 & 1.00 & 4.04 & 16.22 & 1.96 & 7.51 & 26.07 \\
			XFeat~\cite{potje2024xfeat} & NN & 9.4 & 3.108 & 1.487 & 44.60 & 59.92 & 75.97 & 0.44 & 1.98 & 9.17 & 0.87 & 3.78 & 15.96 \\
			\hdashline[2pt/1.5pt]
			SIFT & NN+PointCN~\cite{yi2018learning} & 3.8 & 4.080 & 1.244 & 57.17 & 72.35 & 86.59 & 0.22 & 1.06 & 5.63 & 0.44 & 2.07 & 10.26 \\
			SIFT & NN+ConvMatch+~\cite{zhang2024convmatch} & 4.2 & 4.053 & 1.270 & 57.48 & 72.68 & 86.64 & 0.25 & 1.18 & 6.23 & 0.49 & 2.31 & 11.26 \\
			SIFT & NN+NCMNet+~\cite{liu2024ncmnet} & 4.5 & 4.023 & 1.266 & 57.48 & 72.70 & 86.55 & 0.27 & 1.29 & 6.77 & 0.53 & 2.52 & 12.12 \\
			SIFT & NN+U-Match+~\cite{li2024u} & 4.2 & 4.058 & 1.267 & 57.27 & 72.46 & 86.43 & 0.24 & 1.16 & 6.11 & 0.48 & 2.26 & 11.06 \\
			\hdashline[2pt/1.5pt]
			SuperPoint & SuperGlue~\cite{sarlin2020superglue} & 6.8 & \tabfirst{4.549} & 1.590 & 66.58 & 80.13 & 91.30 & 0.47 & 2.21 & 12.00 & 0.93 & 4.24 & 20.22 \\
			SuperPoint & LightGlue~\cite{lindenberger2023lightglue} & 6.8 & \tabsecond{4.509} & 1.582 & 66.16 & 79.92 & 90.98 & 0.48 & 2.23 & 12.02 & 0.94 & 4.27 & 20.24 \\
			SuperPoint & DiffGlue~\cite{zhang2024diffglue} & 6.8 & 4.356 & 1.577 & 64.90 & 78.90 & 90.34 & 0.50 & 2.24 & 11.69 & 0.98 & 4.29 & 20.02 \\
			ALIKED & LightGlue & 16.3 & \tabthird{4.375} & 1.273 & 64.85 & 78.19 & 89.78 & 1.21 & 4.67 & 17.53 & 2.35 & 8.61 & 28.00 \\
			\hdashline[2pt/1.5pt]
			\multicolumn{2}{c|}{LoFTR~\cite{sun2021loftr}} & 85.3 & 3.970 & 1.331 & 61.29 & 76.52 & 90.05 & 5.36 & 18.06 & \tabsecond{45.13} & 9.45 & 27.52 & \tabsecond{57.22} \\
			\multicolumn{2}{c|}{MatchFormer~\cite{wang2022matchformer}} & 72.4 & 3.751 & 1.428 & 53.09 & 68.98 & 85.84 & 3.89 & 14.02 & 39.45 & 7.03 & 22.13 & 51.11 \\
			\multicolumn{2}{c|}{ELoFTR~\cite{wang2024efficient}} & 35.6 & 3.097 & 1.322 & 57.63 & 73.12 & 89.04 & 2.17 & 8.49 & 28.06 & 4.04 & 14.23 & 40.15 \\
			\multicolumn{2}{c|}{PDC-Net+~\cite{truong2023pdc}} & \tabsecond{230.5} & 3.708 & 1.541 & 38.77 & 57.49 & 79.66 & 5.70 & 17.19 & 37.10 & 9.23 & 24.75 & 48.34 \\
			\multicolumn{2}{c|}{GLU-Net~\cite{truong2020glu}} & 126.1 & 3.328 & 1.482 & 24.03 & 36.71 & 58.71 & 2.60 & 8.91 & 23.82 & 4.33 & 12.90 & 31.36 \\
			\multicolumn{2}{c|}{DKM~\cite{edstedt2023dkm}} & 131.8 & 3.001 & 0.974 & \tabsecond{69.82} & \tabsecond{82.99} & \tabsecond{92.97} & 5.93 & 16.33 & 38.02 & 10.33 & 25.67 & 50.92 \\
			\multicolumn{2}{c|}{RoMa~\cite{edstedt2024roma}} & 160.5 & 3.006 & \tabthird{0.960} & \tabfirst{70.84} & \tabfirst{83.52} & \tabfirst{93.45} & \tabthird{7.02} & \tabthird{18.71} & 40.68 & \tabthird{11.96} & \tabthird{28.67} & 53.80 \\
			\multicolumn{2}{c|}{MASt3R~\cite{leroy2024grounding}} & 35.0 & 3.242 & 1.593 & 45.75 & 63.18 & 82.76 & 1.80 & 8.20 & 33.79 & 3.33 & 13.55 & 45.78 \\
			\multicolumn{2}{c|}{ArgMatch~\cite{deng2025argmatch}} & \tabfirst{340.0} & 3.180 & 1.082 & 64.74 & 79.08 & 91.28 & \tabfirst{13.59} & \tabfirst{33.21} & \tabfirst{60.14} & \tabfirst{20.82} & \tabfirst{44.04} & \tabfirst{70.39} \\
			\multicolumn{2}{c|}{DenseAffine~\cite{sun2025learning}} & \tabthird{224.4} & 3.048 & \tabfirst{0.950} & 67.25 & 81.00 & \tabthird{91.92} & \tabsecond{8.44} & \tabsecond{21.45} & \tabthird{43.98} & \tabsecond{14.16} & \tabsecond{31.97} & \tabthird{56.77} \\
			\multicolumn{2}{c|}{PMatch~\cite{zhu2023pmatch}} & 32.0 & 2.969 & \tabsecond{0.953} & \tabthird{67.96} & \tabthird{81.28} & 91.76 & 2.38 & 8.29 & 26.26 & 4.42 & 14.15 & 38.18 \\
			\bottomrule
		\end{tabular}%
	}
\end{table}

	We further evaluate 3D reconstruction with known camera intrinsics and poses on the ETH3D high-resolution test split~\cite{schops2017multi}. With all cameras fixed, differences between methods come only from their correspondences, so this setting examines whether pairwise correspondences are accurate and consistent enough to form multi-view tracks for triangulation. Table~\ref{table:eth3d_sfm} reports local COLMAP statistics together with the official ETH3D metrics. The number of points and completeness mainly reflect reconstruction density, whereas reprojection error and accuracy reflect geometric precision.
	Among sparse pipelines, ALIKED is the strongest NN front-end, reconstructing 14.3K points and reaching $7.51\%$ official \textbf{F1}$@2$cm, and ALIKED+LightGlue further increases the point count to 16.3K and obtains the best sparse \textbf{F1}$@2$cm ($8.61\%$). SuperPoint+SuperGlue and SuperPoint+LightGlue produce the longest tracks (mean lengths $4.549$ and $4.509$), but SuperPoint-based matchers reconstruct fewer points and have larger reprojection errors than ALIKED-based pipelines. The SIFT-based NN+filter variants keep reprojection errors comparable to SIFT+NN but reduce the number of triangulated points and the official completeness, indicating that the filters also remove valid correspondences.
	Semi-dense/dense matchers substantially improve completeness and \textbf{F1}, although their tracks are shorter (mean length $2.97$--$3.97$) than those of learned sparse matchers ($4.36$--$4.55$). ArgMatch reconstructs the densest point clouds (340.0K points on average) and achieves the best official \textbf{F1}$@2$cm and \textbf{F1}$@5$cm ($44.04\%$ and $70.39\%$), mainly owing to its higher completeness ($33.21\%$ and $60.14\%$). RoMa yields the highest official accuracy ($83.52\%$ at $2$ cm and $93.45\%$ at $5$ cm), and DenseAffine has the lowest reprojection error ($0.950$ pixels) and the second-best \textbf{F1}$@2$cm ($31.97\%$). DKM and LoFTR also outperform sparse pipelines by a large margin, whereas PDC-Net+ and GLU-Net produce many points but with lower accuracy, and PMatch is accurate but less complete. MASt3R has the largest reprojection error among semi-dense/dense matchers ($1.593$ pixels) and reaches only $13.55\%$ \textbf{F1}$@2$cm.

	Overall, most semi-dense/dense matchers excel in challenging scenarios and outperform sparse pipelines even on indoor datasets they were not trained on, whereas sparse matchers are limited by the quality of their keypoints and descriptors. This advantage comes at a cost. On IMC2021, RoMa and DKM are more than an order of magnitude slower than ALIKED+LightGlue (Table~\ref{table:imc2021}), and on ETH3D the tracks of semi-dense/dense matchers are shorter than those of learned sparse matchers (Table~\ref{table:eth3d_sfm}), because their matched pixels are not consistent across views~\cite{shen2022semi,sekikawa2026det}. Learned components also remain tied to their training conditions. Outlier filters trained on SIFT matches do not transfer reliably to other detector-descriptors, and the ranking of matchers changes between outdoor and indoor datasets (Table~\ref{table:pose}).

	\FloatBarrier
	\section{Conclusion and Future Trends}
	\label{subsec:future}
In this survey, we reviewed deep learning-based image matching methods, a key component of many visual applications. We first examined how individual stages of the classical pipeline, namely the detector-descriptor, outlier filter, and geometric estimator, can be replaced by neural networks. We then discussed frameworks that merge multiple stages into end-to-end modules, including sparse, semi-dense, and dense matchers and pose regressors.
We analyzed their design principles, advantages, and limitations, and benchmarked representative methods on relative pose estimation, homography estimation, matching accuracy assessment, visual localization, and 3D reconstruction.
Building on this review, we highlight the following open problems and directions for future research:

\begin{itemize}
\item \textbf{Generalization across Domains:} Learned matchers are only as general as the correspondences they are trained on. Most ground truth comes from SfM and MVS reconstructions of Internet photos~\cite{li2018megadepth} or from RGB-D scans of indoor scenes~\cite{dai2017scannet}. The former exist only where classical reconstruction already succeeds, and the latter are limited by depth-sensor noise and indoor settings, so supervision is scarce in exactly the textureless, reflective, dynamic, and cross-modal cases where matching fails most often. Data engines break this circularity by deriving supervision from other sources, such as self-training on Internet videos~\cite{shen2024gim}, lifting single images to 3D~\cite{liang2025l2m,yang2026anymatch}, rendering novel views~\cite{chen2026matchgs}, and collecting aerial-ground imagery~\cite{vuong2025aerialmegadepth,cheng2026airzoo}. Scaling data, model capacity, and compute also benefits local feature matching, as it does other vision tasks~\cite{nordstrom2026loma}. Each new data source, however, carries its own bias, such as depth-estimation errors, rendering artifacts, or the failure modes of a teacher matcher. The central open problems are therefore how to certify the geometric accuracy of synthesized labels and how to combine data sources so that their biases do not accumulate. Evaluation faces a similar circularity. Pseudo ground truth favors methods that resemble the pipeline that produced it~\cite{brachmann2021limits}, and popular test sets are approaching saturation~\cite{nordstrom2026loma}. Benchmarks with manually annotated or sensor-based ground truth, such as WxBS~\cite{mishkin2015wxbs}, HardMatch~\cite{nordstrom2026loma}, and LaMAR~\cite{sarlin2022lamar}, and large or difficulty-stratified benchmarks, such as IMC~\cite{jin2021image}, the large-scale homography benchmark~\cite{barath2023homography}, RUBIK~\cite{loiseau2025rubik}, and AerialExtreMatch~\cite{wu2026aerialextrematch}, measure progress more reliably than saturated test sets.
\item \textbf{Composability and Calibrated Uncertainty:} The classical pipeline owes much of its longevity to modularity. Any detector can be paired with any robust estimator, and each module can be improved independently. Learned modules lose this property because each one is trained on the outputs of a particular upstream module and implicitly adapts to their error statistics. An outlier filter or a learned sampler trained on SIFT matches, for instance, faces a different noise distribution when it receives matches from another detector-descriptor. Training on synthetic noise~\cite{wang2026mcd} or random keypoints~\cite{xu2026matchany}, and refining matches independently of the detector~\cite{schmid2026xrefine}, move toward modules that remain valid for arbitrary inputs. A more general remedy is a common interface between modules, namely correspondences with calibrated uncertainties whose meaning does not depend on the module that produced them. Uncertainty estimates already exist at several stages, from keypoint scores and covariances~\cite{santellani2024gmm,shenoi2026raco} to probabilistic~\cite{truong2021learning} and evidential~\cite{li2026sure,cai2025evidential} match certainties, but they are rarely calibrated across domains, and they mostly model local localization noise, whereas semi-dense matchers also fail through wrong coarse assignments~\cite{hoang2026semidense}. Calibrated uncertainties would allow robust estimation to weight sampling~\cite{brachmann2019neural}, hypothesis scoring, and bundle adjustment consistently, and would make the uncertainty of the final pose or 3D model computable, as safety-critical applications require.
\item \textbf{Efficiency at the System Level:} In multi-view applications, the cost of matching depends less on the time per image pair than on how computation scales with the number of images. A collection of $N$ images yields $O(N^2)$ candidate pairs. Sparse pipelines extract features once per image and keep the pairwise stage light, whereas semi-dense and dense matchers run most of their networks on every pair, so their accuracy becomes expensive at scale. Faster architectures~\cite{wang2024efficient,deng2025argmatch} reduce the cost per pair, and quantization or distillation~\cite{kanakis2023zippypoint,potje2024xfeat} the cost per image, but larger savings come from changing where computation is spent, by caching rich per-image representations so that the pairwise stage stays lightweight and by computing only where matches can exist. Pruning unmatchable keypoints and non-covisible tokens~\cite{lindenberger2023lightglue,toida2026covisibility}, restricting matching to overlapping regions or matched semantic areas~\cite{chen2022oetr,zhang2026area,zhang2024mesa,zhang2026mesa,liu2025sgad}, routing coarse matching through a few candidate paths~\cite{le2026ultramatch}, and summarizing dense matches before robust estimation~\cite{astermark2025dms} show that much of the current computation goes to correspondences that barely affect the final estimate. The same reasoning applies one level up, to selecting which image pairs to match at all~\cite{arandjelovic2016netvlad}. Combining cached per-image representations with a pairwise stage whose cost adapts to the overlap and difficulty of each pair would bring the accuracy of dense matching to the scale of sparse pipelines.
\item \textbf{From Pairwise to Multi-View Matching:} Most matchers are trained and evaluated on image pairs, whereas their main consumers, such as SfM, SLAM, and novel view synthesis~\cite{kerbl20233d}, operate on many views and require long, consistent tracks. Detector-based pipelines obtain tracks almost for free, because a repeatable detector fires on the same 3D point across images. Semi-dense and dense matchers give up this repeatability to gain robustness. The pixels they match depend on the image pair, so matches from different pairs rarely chain into long tracks~\cite{shen2022semi}. Current remedies approach tracks from both sides. Detect by Track~\cite{sekikawa2026det} and MV-RoMa~\cite{lee2026mvroma} bring tracks to semi-dense and dense matching by anchoring matches to existing tracks or by matching one image against several views, while XRefine~\cite{schmid2026xrefine} and TraqPoint~\cite{liu2026traqpoint} make sparse keypoints more track-consistent by refining tracks or by rewarding keypoints that persist along sequences. A more fundamental shift is to treat the track, not the pair, as the unit of matching, with training objectives, architectures, and benchmarks defined on tracks, at a cost that grows subquadratically with the number of views.
\item \textbf{Large Geometric Models:} Feed-forward 3D models trained on massive data, such as DUSt3R~\cite{wang2024dust3r}, VGGT~\cite{wang2025vggt}, $\pi^3$~\cite{wang2026pi3}, and Depth Anything 3~\cite{lin2026depth}, predict pointmaps and cameras directly from two or more images. They invert the order of the classical pipeline, deriving matches from estimated geometry instead of estimating geometry from matches. This gives them a global, 3D-aware prior that helps where local appearance is ambiguous, \emph{e.g.}, in low-texture or repetitive scenes and under extreme viewpoint changes. Their predictions, however, are made at low resolution and lack the pixel-level precision that bundle adjustment requires. Their features perform poorly for zero-shot matching but provide strong representations when a full matcher is trained on top~\cite{nordstrom2026romaomega}, and their reconstructions are commonly refined by bundle adjustment on predicted tracks~\cite{wang2025vggt}. A division of labor is therefore emerging, in which 3D models provide coarse, globally consistent priors and dedicated matchers deliver precise correspondences. Such priors can enter the pipeline as backbones of dense matchers, either taken from 3D models~\cite{nordstrom2026romaomega} or pre-trained with multi-view objectives~\cite{nordstrom2026mum}, as pose initializations, or as monocular depth priors inside minimal solvers~\cite{yu2025madpose,ding2025reposed}. Two questions follow. The first is where the correspondence knowledge of these models resides and how to read it out. The cross-attention maps of cross-view completion models~\cite{weinzaepfel2023croco} already act as zero-shot correspondence estimators~\cite{an2025zeroco}, and the interface to the rest of the pipeline could be correspondences, pointmaps, or tracks. The second is how to retain the verifiability of the classical pipeline, in which each estimated pose comes with an explicit set of inliers, when part of the geometry is regressed instead of fitted, especially under weak overlap, where too few correspondences exist to constrain a pose.
\item \textbf{Multi-Modal Matching:} Infrared, multi-spectral, SAR, event, and depth sensors increasingly complement RGB cameras, and fusing their data requires spatially aligned images~\cite{gallego2020event,zhang2021image}. Cross-modal alignment is also central to remote sensing~\cite{li2025deep,wu2025mapglue} and medical imaging~\cite{jiang2021review}, and image matching now extends even to genomic analysis~\cite{qiu2024spatiotemporal}. Current cross-modal matchers close the modality gap mainly from the data side, either by pre-training on large-scale synthetic multi-modal pairs~\cite{he2025matchanything,ren2025minima,yang2026anymatch} or by translating new modalities into the domain of existing matchers~\cite{frolov2026pym}. Synthetic modalities do not reproduce sensor-specific imaging physics, such as speckle in SAR or emissivity effects in thermal images, and real benchmarks with accurate ground truth cover only a few modality pairs. Since appearance changes fundamentally across modalities, a universal matcher has to rely on what they share, namely scene structure and geometry, which favors structure- and geometry-aware representations over appearance similarity. Learning such representations from weakly aligned or unaligned real data~\cite{shrivastava2025selfsupervised} is therefore a key direction. The same view extends across dimensions, to matching images with point clouds~\cite{wang2024dgc}. UniCorrn~\cite{goswami2026unicorrn} shows that 2D-2D, 2D-3D, and 3D-3D matching can share one architecture, and extending such a model to all sensing modalities without losing accuracy is the next step.
\end{itemize}

Several of these directions point to the same trade-off. Each step toward merged, end-to-end matching has improved robustness, but it has also given up properties that made the classical pipeline practical, namely features computed once per image and reused across pairs, modules that can be exchanged independently, keypoints that stay consistent across many views, and explicit inliers that show whether an estimated pose can be trusted.

In summary, deep learning has substantially improved the robustness and accuracy of image matching under challenging conditions. We expect the next generation of methods to recover these properties within learned matchers instead of trading them for accuracy on two-view benchmarks, so that the gains hold at scale, across domains and sensors, and in applications that depend on accurate geometry, such as robotics, augmented reality, and autonomous driving.
\section*{Acknowledgements}
This work was supported by National Natural Science Foundation of China (Grant No. 62276192).
\bibliographystyle{scis}
\bibliography{egbib_scis}

\begin{thebibliography}{100}
\providecommand{\url}[1]{\texttt{#1}}
\providecommand{\urlprefix}{URL }
\providecommand{\bibinfo}[2]{#2}

\bibitem{datta2008image}
Datta R, Joshi D, Li J,  et~al.
\newblock Image retrieval: ideas, influences, and trends of the new age.
\newblock ACM Comput Surv, 2008, 40: 5

\bibitem{sattler2018benchmarking}
Sattler T, Maddern W, Toft C,  et~al.
\newblock Benchmarking {6DOF} outdoor visual localization in changing
  conditions.
\newblock In: Proc IEEE Conf Comput Vis Pattern Recognit, 2018.
\newblock 8601--8610

\bibitem{hartley2003multiple}
Hartley R,  Zisserman A.
\newblock Multiple view geometry in computer vision.
\newblock Cambridge: Cambridge University Press, 2003

\bibitem{pan2024global}
Pan L, Barath D, Pollefeys M,  et~al.
\newblock Global structure-from-motion revisited.
\newblock In: Proc Eur Conf Comput Vis, 2024.
\newblock 58--77

\bibitem{engel2014lsd}
Engel J, Sch{\"o}ps T,  Cremers D.
\newblock {LSD-SLAM}: large-scale direct monocular {SLAM}.
\newblock In: Proc Eur Conf Comput Vis, 2014.
\newblock 834--849

\bibitem{kerbl20233d}
Kerbl B, Kopanas G, Leimk{\"u}hler T,  et~al.
\newblock {3D} {Gaussian} splatting for real-time radiance field rendering.
\newblock ACM Trans Graph, 2023, 42: 139

\bibitem{marr1976cooperative}
Marr D,  Poggio T.
\newblock Cooperative computation of stereo disparity.
\newblock Science, 1976, 194: 283--287

\bibitem{brice1970scene}
Brice C~R,  Fennema C~L.
\newblock Scene analysis using regions.
\newblock Artif Intell, 1970, 1: 205--226

\bibitem{gruen1985adaptive}
Gruen A.
\newblock Adaptive least squares correlation: a powerful image matching
  technique.
\newblock S Afr J Photogramm Remote Sens Cartogr, 1985, 14: 175--187

\bibitem{moravec1981rover}
Moravec H~P.
\newblock Rover visual obstacle avoidance.
\newblock In: Proc Int Jt Conf Artif Intell, 1981.
\newblock 785--790

\bibitem{schmid1995matching}
Schmid C,  Mohr R.
\newblock Matching by local invariants.
\newblock Research Report RR-2644, INRIA, 1995

\bibitem{aanaes2012interesting}
Aan{\ae}s H, Dahl A~L,  Steenstrup~Pedersen K.
\newblock Interesting interest points: a comparative study of interest point
  performance on a unique data set.
\newblock Int J Comput Vis, 2012, 97: 18--35

\bibitem{heinly2012comparative}
Heinly J, Dunn E,  Frahm J~M.
\newblock Comparative evaluation of binary features.
\newblock In: Proc Eur Conf Comput Vis, 2012.
\newblock 759--773

\bibitem{awrangjeb2012performance}
Awrangjeb M, Lu G,  Fraser C~S.
\newblock Performance comparisons of contour-based corner detectors.
\newblock IEEE Trans Image Process, 2012, 21: 4167--4179

\bibitem{balntas2017hpatches}
Balntas V, Lenc K, Vedaldi A,  et~al.
\newblock {HPatches}: a benchmark and evaluation of handcrafted and learned
  local descriptors.
\newblock In: Proc IEEE Conf Comput Vis Pattern Recognit, 2017.
\newblock 5173--5182

\bibitem{schonberger2017comparative}
Sch{\"o}nberger J~L, Hardmeier H, Sattler T,  et~al.
\newblock Comparative evaluation of hand-crafted and learned local features.
\newblock In: Proc IEEE Conf Comput Vis Pattern Recognit, 2017.
\newblock 1482--1491

\bibitem{zitova2003image}
Zitov{\'a} B,  Flusser J.
\newblock Image registration methods: a survey.
\newblock Image Vis Comput, 2003, 21: 977--1000

\bibitem{ma2021image}
Ma J, Jiang X, Fan A,  et~al.
\newblock Image matching from handcrafted to deep features: a survey.
\newblock Int J Comput Vis, 2021, 129: 23--79

\bibitem{xu2024local}
Xu S, Chen S, Xu R,  et~al.
\newblock Local feature matching using deep learning: a survey.
\newblock Inf Fusion, 2024, 107: 102344

\bibitem{liao2024local}
Liao Y, Di Y, Zhu K,  et~al.
\newblock Local feature matching from detector-based to detector-free: a
  survey.
\newblock Appl Intell, 2024, 54: 3954--3989

\bibitem{mishkin2021thesis}
Mishkin D.
\newblock Learning and crafting for the wide multiple baseline stereo.
\newblock Dissertation for Ph.D. Degree. Prague: Czech Technical University in
  Prague, 2021

\bibitem{edstedt2025thesis}
Edstedt J.
\newblock Towards the next generation of {3D} reconstruction.
\newblock Dissertation for Ph.D. Degree. Link{\"o}ping: Link{\"o}ping
  University, 2025

\bibitem{yi2018learning}
Yi K~M, Trulls E, Ono Y,  et~al.
\newblock Learning to find good correspondences.
\newblock In: Proc IEEE Conf Comput Vis Pattern Recognit, 2018.
\newblock 2666--2674

\bibitem{detone2018superpoint}
DeTone D, Malisiewicz T,  Rabinovich A.
\newblock {SuperPoint}: self-supervised interest point detection and
  description.
\newblock In: Proc IEEE Conf Comput Vis Pattern Recognit Workshops, 2018.
\newblock 224--236

\bibitem{truong2023pdc}
Truong P, Danelljan M, Timofte R,  et~al.
\newblock {PDC-Net+}: enhanced probabilistic dense correspondence network.
\newblock IEEE Trans Pattern Anal Mach Intell, 2023, 45: 10247--10266

\bibitem{sarlin2019coarse}
Sarlin P~E, Cadena C, Siegwart R,  et~al.
\newblock From coarse to fine: robust hierarchical localization at large scale.
\newblock In: Proc IEEE/CVF Conf Comput Vis Pattern Recognit, 2019.
\newblock 12716--12725

\bibitem{lindenberger2021pixsfm}
Lindenberger P, Sarlin P~E, Larsson V,  et~al.
\newblock Pixel-perfect structure-from-motion with featuremetric refinement.
\newblock In: Proc IEEE/CVF Int Conf Comput Vis, 2021.
\newblock 5987--5997

\bibitem{schonberger2016structure}
Sch{\"o}nberger J~L,  Frahm J~M.
\newblock Structure-from-motion revisited.
\newblock In: Proc IEEE Conf Comput Vis Pattern Recognit, 2016.
\newblock 4104--4113

\bibitem{lowe2004distinctive}
Lowe D~G.
\newblock Distinctive image features from scale-invariant keypoints.
\newblock Int J Comput Vis, 2004, 60: 91--110

\bibitem{harris1988combined}
Harris C,  Stephens M.
\newblock A combined corner and edge detector.
\newblock In: Proc Alvey Vis Conf, 1988.
\newblock 147--151

\bibitem{rosten2006machine}
Rosten E,  Drummond T.
\newblock Machine learning for high-speed corner detection.
\newblock In: Proc Eur Conf Comput Vis, 2006.
\newblock 430--443

\bibitem{matas2004robust}
Matas J, Chum O, Urban M,  et~al.
\newblock Robust wide-baseline stereo from maximally stable extremal regions.
\newblock Image Vis Comput, 2004, 22: 761--767

\bibitem{mikolajczyk2005comparison}
Mikolajczyk K, Tuytelaars T, Schmid C,  et~al.
\newblock A comparison of affine region detectors.
\newblock Int J Comput Vis, 2005, 65: 43--72

\bibitem{lowe1999object}
Lowe D~G.
\newblock Object recognition from local scale-invariant features.
\newblock In: Proc IEEE Int Conf Comput Vis, 1999.
\newblock 1150--1157

\bibitem{rublee2011orb}
Rublee E, Rabaud V, Konolige K,  et~al.
\newblock {ORB}: an efficient alternative to {SIFT} or {SURF}.
\newblock In: Proc IEEE Int Conf Comput Vis, 2011.
\newblock 2564--2571

\bibitem{calonder2010brief}
Calonder M, Lepetit V, Strecha C,  et~al.
\newblock {BRIEF}: binary robust independent elementary features.
\newblock In: Proc Eur Conf Comput Vis, 2010.
\newblock 778--792

\bibitem{lin2023illumination}
Lin X, Zhou Y, Zhang X,  et~al.
\newblock Illumination-insensitive binary descriptor for visual measurement
  based on local inter-patch invariance.
\newblock IEEE Trans Instrum Meas, 2023, 72: 1--13

\bibitem{ma2014robust}
Ma J, Zhao J, Tian J,  et~al.
\newblock Robust point matching via vector field consensus.
\newblock IEEE Trans Image Process, 2014, 23: 1706--1721

\bibitem{bian2017gms}
Bian J, Lin W~Y, Matsushita Y,  et~al.
\newblock {GMS}: grid-based motion statistics for fast, ultra-robust feature
  correspondence.
\newblock In: Proc IEEE Conf Comput Vis Pattern Recognit, 2017.
\newblock 4181--4190

\bibitem{torr1997development}
Torr P~H~S,  Murray D~W.
\newblock The development and comparison of robust methods for estimating the
  fundamental matrix.
\newblock Int J Comput Vis, 1997, 24: 271--300

\bibitem{fischler1981random}
Fischler M~A,  Bolles R~C.
\newblock Random sample consensus: a paradigm for model fitting with
  applications to image analysis and automated cartography.
\newblock Commun ACM, 1981, 24: 381--395

\bibitem{raguram2012usac}
Raguram R, Chum O, Pollefeys M,  et~al.
\newblock {USAC}: a universal framework for random sample consensus.
\newblock IEEE Trans Pattern Anal Mach Intell, 2013, 35: 2022--2038

\bibitem{torr2000mlesac}
Torr P~H~S,  Zisserman A.
\newblock {MLESAC}: a new robust estimator with application to estimating image
  geometry.
\newblock Comput Vis Image Underst, 2000, 78: 138--156

\bibitem{chum2003locally}
Chum O, Matas J,  Kittler J.
\newblock Locally optimized {RANSAC}.
\newblock In: Proc DAGM Ger Conf Pattern Recognit, 2003.
\newblock 236--243

\bibitem{barath2018graph}
Barath D,  Matas J.
\newblock Graph-cut {RANSAC}.
\newblock In: Proc IEEE Conf Comput Vis Pattern Recognit, 2018.
\newblock 6733--6741

\bibitem{barath2020magsac++}
Barath D, Noskova J, Ivashechkin M,  et~al.
\newblock {MAGSAC++}, a fast, reliable and accurate robust estimator.
\newblock In: Proc IEEE/CVF Conf Comput Vis Pattern Recognit, 2020.
\newblock 1304--1312

\bibitem{trajkovic1998fast}
Trajkovi{\'c} M,  Hedley M.
\newblock Fast corner detection.
\newblock Image Vis Comput, 1998, 16: 75--87

\bibitem{kienzle2006learning}
Kienzle W, Wichmann F~A, Sch{\"o}lkopf B,  et~al.
\newblock Learning an interest operator from human eye movements.
\newblock In: Proc IEEE Conf Comput Vis Pattern Recognit Workshops, 2006.
\newblock 24

\bibitem{rosten2008faster}
Rosten E, Porter R,  Drummond T.
\newblock Faster and better: a machine learning approach to corner detection.
\newblock IEEE Trans Pattern Anal Mach Intell, 2010, 32: 105--119

\bibitem{hartmann2014predicting}
Hartmann W, Havlena M,  Schindler K.
\newblock Predicting matchability.
\newblock In: Proc IEEE Conf Comput Vis Pattern Recognit, 2014.
\newblock 9--16

\bibitem{richardson2013learning}
Richardson A,  Olson E.
\newblock Learning convolutional filters for interest point detection.
\newblock In: Proc IEEE Int Conf Robot Autom, 2013.
\newblock 631--637

\bibitem{verdie2015tilde}
Verdie Y, Yi K~M, Fua P,  et~al.
\newblock {TILDE}: a temporally invariant learned detector.
\newblock In: Proc IEEE Conf Comput Vis Pattern Recognit, 2015.
\newblock 5279--5288

\bibitem{detone2017toward}
DeTone D, Malisiewicz T,  Rabinovich A.
\newblock Toward geometric deep {SLAM}.
\newblock ArXiv:1707.07410, 2017

\bibitem{lenc2016learning}
Lenc K,  Vedaldi A.
\newblock Learning covariant feature detectors.
\newblock In: Proc Eur Conf Comput Vis Workshops, 2016.
\newblock 100--117

\bibitem{zhang2017learning}
Zhang X, Yu F~X, Karaman S,  et~al.
\newblock Learning discriminative and transformation covariant local feature
  detectors.
\newblock In: Proc IEEE Conf Comput Vis Pattern Recognit, 2017.
\newblock 6818--6826

\bibitem{barroso2019key}
Barroso-Laguna A, Riba E, Ponsa D,  et~al.
\newblock {Key.Net}: keypoint detection by handcrafted and learned {CNN}
  filters.
\newblock In: Proc IEEE/CVF Int Conf Comput Vis, 2019.
\newblock 5836--5844

\bibitem{barroso2022key}
Barroso-Laguna A,  Mikolajczyk K.
\newblock {Key.Net}: keypoint detection by handcrafted and learned {CNN}
  filters revisited.
\newblock IEEE Trans Pattern Anal Mach Intell, 2023, 45: 698--711

\bibitem{savinov2017quad}
Savinov N, Seki A, Ladick{\'y} L,  et~al.
\newblock Quad-networks: unsupervised learning to rank for interest point
  detection.
\newblock In: Proc IEEE Conf Comput Vis Pattern Recognit, 2017.
\newblock 1822--1830

\bibitem{benbihi2019elf}
Benbihi A, Geist M,  Pradalier C.
\newblock {ELF}: embedded localisation of features in pre-trained {CNN}.
\newblock In: Proc IEEE/CVF Int Conf Comput Vis, 2019.
\newblock 7940--7949

\bibitem{tian2020d2d}
Tian Y, Balntas V, Ng T,  et~al.
\newblock {D2D}: keypoint extraction with describe to detect approach.
\newblock In: Proc Asian Conf Comput Vis, 2020.
\newblock 223--240

\bibitem{pakulev2023ness}
Pakulev K, Vakhitov A,  Ferrer G.
\newblock {NeSS-ST}: detecting good and stable keypoints with a neural
  stability score and the {Shi-Tomasi} detector.
\newblock In: Proc IEEE/CVF Int Conf Comput Vis, 2023.
\newblock 9578--9588

\bibitem{shi1994good}
Shi J,  Tomasi C.
\newblock Good features to track.
\newblock In: Proc IEEE Conf Comput Vis Pattern Recognit, 1994.
\newblock 593--600

\bibitem{lee2022self}
Lee J, Kim B,  Cho M.
\newblock Self-supervised equivariant learning for oriented keypoint detection.
\newblock In: Proc IEEE/CVF Conf Comput Vis Pattern Recognit, 2022.
\newblock 4847--4857

\bibitem{barbaraniscale}
Barbarani G, Vaccarino F, Trivigno G,  et~al.
\newblock Scale-free image keypoints using differentiable persistent homology.
\newblock In: Proc Int Conf Mach Learn, 2024.
\newblock 2990--3002

\bibitem{edstedt2025dad}
Edstedt J, B{\"o}kman G, Wadenb{\"a}ck M,  et~al.
\newblock {DaD}: distilled reinforcement learning for diverse keypoint
  detection.
\newblock ArXiv:2503.07347, 2025

\bibitem{shenoi2026raco}
Shenoi A, Lindenberger P, Sarlin P~E,  et~al.
\newblock {RaCo}: ranking and covariance for practical learned keypoints.
\newblock In: Proc Int Conf 3D Vis, 2026.
\newblock 267--276

\bibitem{ke2004pca}
Ke Y,  Sukthankar R.
\newblock {PCA-SIFT}: a more distinctive representation for local image
  descriptors.
\newblock In: Proc IEEE Conf Comput Vis Pattern Recognit, 2004.
\newblock 506--513

\bibitem{cai2010learning}
Cai H, Mikolajczyk K,  Matas J.
\newblock Learning linear discriminant projections for dimensionality reduction
  of image descriptors.
\newblock IEEE Trans Pattern Anal Mach Intell, 2011, 33: 338--352

\bibitem{brown2010discriminative}
Brown M, Hua G,  Winder S.
\newblock Discriminative learning of local image descriptors.
\newblock IEEE Trans Pattern Anal Mach Intell, 2011, 33: 43--57

\bibitem{hua2007discriminant}
Hua G, Brown M,  Winder S.
\newblock Discriminant embedding for local image descriptors.
\newblock In: Proc IEEE Int Conf Comput Vis, 2007.
\newblock 1--8

\bibitem{winder2007learning}
Winder S~A~J,  Brown M.
\newblock Learning local image descriptors.
\newblock In: Proc IEEE Conf Comput Vis Pattern Recognit, 2007.
\newblock 1--8

\bibitem{strecha2011ldahash}
Strecha C, Bronstein A, Bronstein M,  et~al.
\newblock {LDAHash}: improved matching with smaller descriptors.
\newblock IEEE Trans Pattern Anal Mach Intell, 2012, 34: 66--78

\bibitem{trzcinski2013boosting}
Trzci{\'n}ski T, Christoudias M, Fua P,  et~al.
\newblock Boosting binary keypoint descriptors.
\newblock In: Proc IEEE Conf Comput Vis Pattern Recognit, 2013.
\newblock 2874--2881

\bibitem{simonyan2014learning}
Simonyan K, Vedaldi A,  Zisserman A.
\newblock Learning local feature descriptors using convex optimisation.
\newblock IEEE Trans Pattern Anal Mach Intell, 2014, 36: 1573--1585

\bibitem{bromley1993signature}
Bromley J, Guyon I, LeCun Y,  et~al.
\newblock Signature verification using a ``{Siamese}'' time delay neural
  network.
\newblock In: Adv Neural Inf Process Syst, 1993.
\newblock 737--744

\bibitem{zagoruyko2015learning}
Zagoruyko S,  Komodakis N.
\newblock Learning to compare image patches via convolutional neural networks.
\newblock In: Proc IEEE Conf Comput Vis Pattern Recognit, 2015.
\newblock 4353--4361

\bibitem{han2015matchnet}
Han X, Leung T, Jia Y,  et~al.
\newblock {MatchNet}: unifying feature and metric learning for patch-based
  matching.
\newblock In: Proc IEEE Conf Comput Vis Pattern Recognit, 2015.
\newblock 3279--3286

\bibitem{simo2015discriminative}
Simo-Serra E, Trulls E, Ferraz L,  et~al.
\newblock Discriminative learning of deep convolutional feature point
  descriptors.
\newblock In: Proc IEEE Int Conf Comput Vis, 2015.
\newblock 118--126

\bibitem{balntas2016learning}
Balntas V, Riba E, Ponsa D,  et~al.
\newblock Learning local feature descriptors with triplets and shallow
  convolutional neural networks.
\newblock In: Proc Br Mach Vis Conf, 2016.
\newblock 119.1--119.11

\bibitem{balntas2016pnnet}
Balntas V, Johns E, Tang L,  et~al.
\newblock {PN-Net}: conjoined triple deep network for learning local image
  descriptors.
\newblock ArXiv:1601.05030, 2016

\bibitem{kumar2016learning}
Kumar~BG V, Carneiro G,  Reid I.
\newblock Learning local image descriptors with deep {Siamese} and triplet
  convolutional networks by minimising global loss functions.
\newblock In: Proc IEEE Conf Comput Vis Pattern Recognit, 2016.
\newblock 5385--5394

\bibitem{tian2017l2}
Tian Y, Fan B,  Wu F.
\newblock {L2-Net}: deep learning of discriminative patch descriptor in
  {Euclidean} space.
\newblock In: Proc IEEE Conf Comput Vis Pattern Recognit, 2017.
\newblock 661--669

\bibitem{mishchuk2017working}
Mishchuk A, Mishkin D, Radenovi{\'c} F,  et~al.
\newblock Working hard to know your neighbor's margins: local descriptor
  learning loss.
\newblock In: Adv Neural Inf Process Syst, 2017.
\newblock 4826--4837

\bibitem{zhang2017learning1}
Zhang X, Yu F~X, Kumar S,  et~al.
\newblock Learning spread-out local feature descriptors.
\newblock In: Proc IEEE Int Conf Comput Vis, 2017.
\newblock 4595--4603

\bibitem{tian2019sosnet}
Tian Y, Yu X, Fan B,  et~al.
\newblock {SOSNet}: second order similarity regularization for local descriptor
  learning.
\newblock In: Proc IEEE/CVF Conf Comput Vis Pattern Recognit, 2019.
\newblock 11016--11025

\bibitem{tian2020hynet}
Tian Y, Barroso-Laguna A, Ng T,  et~al.
\newblock {HyNet}: learning local descriptor with hybrid similarity measure and
  triplet loss.
\newblock In: Adv Neural Inf Process Syst, 2020.
\newblock 7401--7412

\bibitem{deng2024sdgmnet}
Deng Y,  Ma J.
\newblock {SDGMNet}: statistic-based dynamic gradient modulation for local
  descriptor learning.
\newblock In: Proc AAAI Conf Artif Intell, 2024.
\newblock 1510--1518

\bibitem{keller2018learning}
Keller M, Chen Z, Maffra F,  et~al.
\newblock Learning deep descriptors with scale-aware triplet networks.
\newblock In: Proc IEEE Conf Comput Vis Pattern Recognit, 2018.
\newblock 2762--2770

\bibitem{he2018local}
He K, Lu Y,  Sclaroff S.
\newblock Local descriptors optimized for average precision.
\newblock In: Proc IEEE Conf Comput Vis Pattern Recognit, 2018.
\newblock 596--605

\bibitem{luo2018geodesc}
Luo Z, Shen T, Zhou L,  et~al.
\newblock {GeoDesc}: learning local descriptors by integrating geometry
  constraints.
\newblock In: Proc Eur Conf Comput Vis, 2018.
\newblock 170--185

\bibitem{wang2020learning}
Wang Q, Zhou X, Hariharan B,  et~al.
\newblock Learning feature descriptors using camera pose supervision.
\newblock In: Proc Eur Conf Comput Vis, 2020.
\newblock 757--774

\bibitem{pultar2019leveraging}
Pultar M, Mishkin D,  Matas J.
\newblock Leveraging outdoor webcams for local descriptor learning.
\newblock In: Proc Comput Vis Winter Workshop, 2019.
\newblock 51--60

\bibitem{toft2020single}
Toft C, Turmukhambetov D, Sattler T,  et~al.
\newblock Single-image depth prediction makes feature matching easier.
\newblock In: Proc Eur Conf Comput Vis, 2020.
\newblock 473--492

\bibitem{luo2019contextdesc}
Luo Z, Shen T, Zhou L,  et~al.
\newblock {ContextDesc}: local descriptor augmentation with cross-modality
  context.
\newblock In: Proc IEEE/CVF Conf Comput Vis Pattern Recognit, 2019.
\newblock 2527--2536

\bibitem{wang2022mtldesc}
Wang C, Xu R, Zhang Y,  et~al.
\newblock {MTLDesc}: looking wider to describe better.
\newblock In: Proc AAAI Conf Artif Intell, 2022.
\newblock 2388--2396

\bibitem{wang2023awd}
Wang C, Xu R, Lu K,  et~al.
\newblock Attention weighted local descriptors.
\newblock IEEE Trans Pattern Anal Mach Intell, 2023, 45: 10632--10649

\bibitem{wang2023featurebooster}
Wang X, Liu Z, Hu Y,  et~al.
\newblock {FeatureBooster}: boosting feature descriptors with a lightweight
  neural network.
\newblock In: Proc IEEE/CVF Conf Comput Vis Pattern Recognit, 2023.
\newblock 7630--7639

\bibitem{shrestha2023residual}
Shrestha R, Chhatkuli A, Kanakis M,  et~al.
\newblock Residual learning for image point descriptors.
\newblock ArXiv:2312.15471, 2023

\bibitem{schneider2026crossfeat}
Schneider P,  Haouchine N.
\newblock {CrossFeat}: bridging imaging modalities in feature descriptor space.
\newblock In: Proc Eur Conf Comput Vis, 2026.
\newblock 53--71

\bibitem{mishkin2018repeatability}
Mishkin D, Radenovi{\'c} F,  Matas J.
\newblock Repeatability is not enough: learning affine regions via
  discriminability.
\newblock In: Proc Eur Conf Comput Vis, 2018.
\newblock 287--304

\bibitem{ebel2019beyond}
Ebel P, Mishchuk A, Yi K~M,  et~al.
\newblock Beyond {Cartesian} representations for local descriptors.
\newblock In: Proc IEEE/CVF Int Conf Comput Vis, 2019.
\newblock 253--262

\bibitem{liu2019gift}
Liu Y, Shen Z, Lin Z,  et~al.
\newblock {GIFT}: learning transformation-invariant dense visual descriptors
  via group {CNNs}.
\newblock In: Adv Neural Inf Process Syst, 2019.
\newblock 6992--7003

\bibitem{cohen2016group}
Cohen T,  Welling M.
\newblock Group equivariant convolutional networks.
\newblock In: Proc Int Conf Mach Learn, 2016.
\newblock 2990--2999

\bibitem{lee2023learning}
Lee J, Kim B, Kim S,  et~al.
\newblock Learning rotation-equivariant features for visual correspondence.
\newblock In: Proc IEEE/CVF Conf Comput Vis Pattern Recognit, 2023.
\newblock 21887--21897

\bibitem{weiler2019general}
Weiler M,  Cesa G.
\newblock General {E(2)}-equivariant steerable {CNNs}.
\newblock In: Adv Neural Inf Process Syst, 2019.
\newblock 14334--14345

\bibitem{parihar2021rord}
Parihar U~S, Gujarathi A, Mehta K,  et~al.
\newblock {RoRD}: rotation-robust descriptors and orthographic views for local
  feature matching.
\newblock In: Proc IEEE/RSJ Int Conf Intell Robots Syst, 2021.
\newblock 1593--1600

\bibitem{pautrat2020online}
Pautrat R, Larsson V, Oswald M~R,  et~al.
\newblock Online invariance selection for local feature descriptors.
\newblock In: Proc Eur Conf Comput Vis, 2020.
\newblock 707--724

\bibitem{durso2026sandesc}
D'Urso M, Santellani E, Sormann C,  et~al.
\newblock A streamlined attention-based network for descriptor extraction.
\newblock In: Proc Int Conf 3D Vis, 2026.
\newblock 247--256

\bibitem{xue2025matcha}
Xue F, Elflein S, Leal-Taix{\'e} L,  et~al.
\newblock {MATCHA}: towards matching anything.
\newblock In: Proc IEEE/CVF Conf Comput Vis Pattern Recognit, 2025.
\newblock 27081--27091

\bibitem{mikolajczyk2005performance}
Mikolajczyk K,  Schmid C.
\newblock A performance evaluation of local descriptors.
\newblock IEEE Trans Pattern Anal Mach Intell, 2005, 27: 1615--1630

\bibitem{yi2016lift}
Yi K~M, Trulls E, Lepetit V,  et~al.
\newblock {LIFT}: learned invariant feature transform.
\newblock In: Proc Eur Conf Comput Vis, 2016.
\newblock 467--483

\bibitem{ono2018lf}
Ono Y, Trulls E, Fua P,  et~al.
\newblock {LF-Net}: learning local features from images.
\newblock In: Adv Neural Inf Process Syst, 2018.
\newblock 6234--6244

\bibitem{jaderberg2015spatial}
Jaderberg M, Simonyan K, Zisserman A,  et~al.
\newblock Spatial transformer networks.
\newblock In: Adv Neural Inf Process Syst, 2015.
\newblock 2017--2025

\bibitem{shen2019rf}
Shen X, Wang C, Li X,  et~al.
\newblock {RF-Net}: an end-to-end image matching network based on receptive
  field.
\newblock In: Proc IEEE/CVF Conf Comput Vis Pattern Recognit, 2019.
\newblock 8132--8140

\bibitem{santellani2023strek}
Santellani E, Sormann C, Rossi M,  et~al.
\newblock {S-TREK}: sequential translation and rotation equivariant keypoints
  for local feature extraction.
\newblock In: Proc IEEE/CVF Int Conf Comput Vis, 2023.
\newblock 9728--9737

\bibitem{dusmanu2019d2}
Dusmanu M, Rocco I, Pajdla T,  et~al.
\newblock {D2-Net}: a trainable {CNN} for joint description and detection of
  local features.
\newblock In: Proc IEEE/CVF Conf Comput Vis Pattern Recognit, 2019.
\newblock 8092--8101

\bibitem{luo2020aslfeat}
Luo Z, Zhou L, Bai X,  et~al.
\newblock {ASLFeat}: learning local features of accurate shape and
  localization.
\newblock In: Proc IEEE/CVF Conf Comput Vis Pattern Recognit, 2020.
\newblock 6589--6598

\bibitem{dai2017deformable}
Dai J, Qi H, Xiong Y,  et~al.
\newblock Deformable convolutional networks.
\newblock In: Proc IEEE Int Conf Comput Vis, 2017.
\newblock 764--773

\bibitem{deng2022redfeat}
Deng Y,  Ma J.
\newblock {ReDFeat}: recoupling detection and description for multimodal
  feature learning.
\newblock IEEE Trans Image Process, 2023, 32: 591--602

\bibitem{deng2026sgpfeat}
Deng Y, Wang B, Zhang K,  et~al.
\newblock {SGPFeat}: semantic and geometric priors for multi-modal image
  matching.
\newblock In: Proc AAAI Conf Artif Intell, 2026.
\newblock 3569--3577

\bibitem{christiansen2019unsuperpoint}
Christiansen P~H, Kragh M~F, Brodskiy Y,  et~al.
\newblock {UnsuperPoint}: end-to-end unsupervised interest point detector and
  descriptor.
\newblock ArXiv:1907.04011, 2019

\bibitem{song2020sekd}
Song Y, Cai L, Li J,  et~al.
\newblock {SEKD}: self-evolving keypoint detection and description.
\newblock ArXiv:2006.05077, 2020

\bibitem{tang2019neural}
Tang J, Kim H, Guizilini V,  et~al.
\newblock Neural outlier rejection for self-supervised keypoint learning.
\newblock In: Proc Int Conf Learn Represent, 2020

\bibitem{gleize2023silk}
Gleize P, Wang W,  Feiszli M.
\newblock {SiLK}: simple learned keypoints.
\newblock In: Proc IEEE/CVF Int Conf Comput Vis, 2023.
\newblock 22499--22508

\bibitem{kanakis2023zippypoint}
Kanakis M, Maurer S, Spallanzani M,  et~al.
\newblock {ZippyPoint}: fast interest point detection, description, and
  matching through mixed precision discretization.
\newblock In: Proc IEEE/CVF Conf Comput Vis Pattern Recognit Workshops, 2023.
\newblock 6114--6123

\bibitem{bhowmik2020reinforced}
Bhowmik A, Gumhold S, Rother C,  et~al.
\newblock Reinforced feature points: optimizing feature detection and
  description for a high-level task.
\newblock In: Proc IEEE/CVF Conf Comput Vis Pattern Recognit, 2020.
\newblock 4948--4957

\bibitem{revaud2019r2d2}
Revaud J, de~Souza C, Humenberger M,  et~al.
\newblock {R2D2}: reliable and repeatable detector and descriptor.
\newblock In: Adv Neural Inf Process Syst, 2019.
\newblock 12405--12415

\bibitem{hu2025hldd}
Hu M, Sun B, Zhang F,  et~al.
\newblock {HLDD}: hierarchically learned detector and descriptor for robust
  image matching.
\newblock IEEE Trans Image Process, 2025, 34: 3123--3138

\bibitem{zhang2026larked}
Zhang H, Shan J, Chang K,  et~al.
\newblock {LARKED}: a lightweight and reliable keypoint detection method for
  feature matching.
\newblock Comput Vis Image Underst, 2026, 268: 104780

\bibitem{tyszkiewicz2020disk}
Tyszkiewicz M, Fua P,  Trulls E.
\newblock {DISK}: learning local features with policy gradient.
\newblock In: Adv Neural Inf Process Syst, 2020.
\newblock 14254--14265

\bibitem{sutton2018reinforcement}
Sutton R~S,  Barto A~G.
\newblock Reinforcement learning: an introduction.
\newblock Cambridge, MA: MIT Press, 2018

\bibitem{kunzel2025ripe}
K{\"u}nzel J, Hilsmann A,  Eisert P.
\newblock {RIPE}: reinforcement learning on unlabeled image pairs for robust
  keypoint extraction.
\newblock In: Proc IEEE/CVF Int Conf Comput Vis, 2025.
\newblock 4868--4877

\bibitem{zhao2022alike}
Zhao X, Wu X, Miao J,  et~al.
\newblock {ALIKE}: accurate and lightweight keypoint detection and descriptor
  extraction.
\newblock IEEE Trans Multimedia, 2023, 25: 3101--3112

\bibitem{zhao2023aliked}
Zhao X, Wu X, Chen W,  et~al.
\newblock {ALIKED}: a lighter keypoint and descriptor extraction network via
  deformable transformation.
\newblock IEEE Trans Instrum Meas, 2023, 72: 1--16

\bibitem{costa2026upal}
Costa F, Kreft R, Goedeke E,  et~al.
\newblock Unified and efficient point-line local features.
\newblock In: Proc Eur Conf Comput Vis, 2026.
\newblock 175--194

\bibitem{xue2023sfd2}
Xue F, Budvytis I,  Cipolla R.
\newblock {SFD2}: semantic-guided feature detection and description.
\newblock In: Proc IEEE/CVF Conf Comput Vis Pattern Recognit, 2023.
\newblock 5206--5216

\bibitem{wu2025samfeat}
Wu J, Xu R, Wood-Doughty Z,  et~al.
\newblock Segment anything model is a good teacher for local feature learning.
\newblock IEEE Trans Image Process, 2025, 34: 2097--2111

\bibitem{li2025sde2d}
Li J, Zhang R, Li G,  et~al.
\newblock {SDE2D}: semantic-guided discriminability enhancement feature
  detector and descriptor.
\newblock IEEE Trans Multimedia, 2025, 27: 275--286

\bibitem{wang2025semantic}
Wang L, Zhang Y, Ge F,  et~al.
\newblock Learning local features by jointly semantic-guided and task rewards.
\newblock IEEE Trans Circuits Syst Video Technol, 2025, 35: 2045--2056

\bibitem{li2021rap}
Li D, Miao J, Shi X,  et~al.
\newblock {RaP-Net}: a region-wise and point-wise weighting network to extract
  robust features for indoor localization.
\newblock In: Proc IEEE/RSJ Int Conf Intell Robots Syst, 2021.
\newblock 1331--1338

\bibitem{jung2023salient}
Jung Y, Syazwany N~S,  Lee S~C.
\newblock Local feature extraction from salient regions by feature map
  transformation.
\newblock In: Proc Br Mach Vis Conf, 2022

\bibitem{wang2022rethinking}
Wang C, Zhang G, Cheng Z,  et~al.
\newblock Rethinking low-level features for interest point detection and
  description.
\newblock In: Proc Asian Conf Comput Vis, 2022.
\newblock 108--123

\bibitem{wang2023transformation}
Wang Z, Wu C, Yang Y,  et~al.
\newblock Learning transformation-predictive representations for detection and
  description of local features.
\newblock In: Proc IEEE/CVF Conf Comput Vis Pattern Recognit, 2023.
\newblock 11464--11473

\bibitem{he2023d2former}
He J, Gao Y, Zhang T,  et~al.
\newblock {D2Former}: jointly learning hierarchical detectors and contextual
  descriptors via agent-based transformers.
\newblock In: Proc IEEE/CVF Conf Comput Vis Pattern Recognit, 2023.
\newblock 2904--2914

\bibitem{li2022decoupling}
Li K, Wang L, Liu L,  et~al.
\newblock Decoupling makes weakly supervised local feature better.
\newblock In: Proc IEEE/CVF Conf Comput Vis Pattern Recognit, 2022.
\newblock 15838--15848

\bibitem{sun2023shared}
Sun J, Ji L,  Zhu J.
\newblock Shared coupling-bridge scheme for weakly supervised local feature
  learning.
\newblock IEEE Trans Multimedia, 2024, 26: 1200--1212

\bibitem{edstedt2024dedode}
Edstedt J, B{\"o}kman G, Wadenb{\"a}ck M,  et~al.
\newblock {DeDoDe}: detect, don't describe---describe, don't detect for local
  feature matching.
\newblock In: Proc Int Conf 3D Vis, 2024.
\newblock 148--157

\bibitem{edstedt2024dedode2}
Edstedt J, B{\"o}kman G,  Zhao Z.
\newblock {DeDoDe} v2: analyzing and improving the {DeDoDe} keypoint detector.
\newblock In: Proc IEEE/CVF Conf Comput Vis Pattern Recognit Workshops, 2024.
\newblock 4245--4253

\bibitem{bokman2024steerers}
B{\"o}kman G, Edstedt J, Felsberg M,  et~al.
\newblock Steerers: a framework for rotation equivariant keypoint descriptors.
\newblock In: Proc IEEE/CVF Conf Comput Vis Pattern Recognit, 2024.
\newblock 4885--4895

\bibitem{bokman2024affine}
B{\"o}kman G, Edstedt J, Felsberg M,  et~al.
\newblock Affine steerers for structured keypoint description.
\newblock In: Proc Eur Conf Comput Vis, 2024.
\newblock 449--468

\bibitem{chen2025rdd}
Chen G, Fu T, Chen H,  et~al.
\newblock {RDD}: robust feature detector and descriptor using deformable
  transformer.
\newblock In: Proc IEEE/CVF Conf Comput Vis Pattern Recognit, 2025.
\newblock 6394--6403

\bibitem{liu2026traqpoint}
Liu Y, Li H, Yang L,  et~al.
\newblock From pairs to sequences: track-aware policy gradients for keypoint
  detection.
\newblock In: Proc IEEE/CVF Conf Comput Vis Pattern Recognit, 2026.
\newblock 21453--21463

\bibitem{potje2024xfeat}
Potje G, Cadar F, Araujo A,  et~al.
\newblock {XFeat}: accelerated features for lightweight image matching.
\newblock In: Proc IEEE/CVF Conf Comput Vis Pattern Recognit, 2024.
\newblock 2682--2691

\bibitem{liu2025liftfeat}
Liu Y, Lai W, Zhao Z,  et~al.
\newblock {LiftFeat}: {3D} geometry-aware local feature matching.
\newblock In: Proc IEEE Int Conf Robot Autom, 2025.
\newblock 11714--11720

\bibitem{shi2026semlight}
Shi G, Liu X, Li Y,  et~al.
\newblock {SemLight}: distilled semantic--geometric fusion for efficient local
  feature matching.
\newblock In: Proc Eur Conf Comput Vis, 2026.
\newblock 442--459

\bibitem{zhang2018learningtexture}
Zhang L,  Rusinkiewicz S.
\newblock Learning to detect features in texture images.
\newblock In: Proc IEEE Conf Comput Vis Pattern Recognit, 2018.
\newblock 6325--6333

\bibitem{liu2022retinal}
Liu J, Li X, Wei Q,  et~al.
\newblock Semi-supervised keypoint detector and descriptor for retinal image
  matching.
\newblock In: Proc Eur Conf Comput Vis, 2022.
\newblock 593--609

\bibitem{he2023darkfeat}
He Y, Hu Y, Zhao W,  et~al.
\newblock {DarkFeat}: noise-robust feature detector and descriptor for
  extremely low-light {RAW} images.
\newblock In: Proc AAAI Conf Artif Intell, 2023.
\newblock 826--834

\bibitem{wang2025derfeat}
Wang F, Lyu H, Xing G,  et~al.
\newblock Degradation-equivariant representations for robust feature detection
  and description in low-light environments.
\newblock IEEE Trans Multimedia, 2025, 27: 7483--7496

\bibitem{yang2023knowledge}
Yang J, Gong M, Nair G,  et~al.
\newblock Knowledge distillation for feature extraction in underwater {VSLAM}.
\newblock In: Proc IEEE Int Conf Robot Autom, 2023.
\newblock 5163--5169

\bibitem{cadar2023deformable}
Cadar F, Melo W, Kanagasabapathi V,  et~al.
\newblock Improving the matching of deformable objects by learning to detect
  keypoints.
\newblock Pattern Recognit Lett, 2023, 175: 83--89

\bibitem{potje2023dalf}
Potje G, Cadar F, Araujo A,  et~al.
\newblock Enhancing deformable local features by jointly learning to detect and
  describe keypoints.
\newblock In: Proc IEEE/CVF Conf Comput Vis Pattern Recognit, 2023.
\newblock 1306--1315

\bibitem{zhao2024balf}
Zhao Z.
\newblock {BALF}: simple and efficient blur aware local feature detector.
\newblock In: Proc IEEE/CVF Winter Conf Appl Comput Vis, 2024.
\newblock 3362--3372

\bibitem{gao2025motionfeat}
Gao Y, Zhang D, Yu X,  et~al.
\newblock {Motion-Feat}: motion blur-aware local feature description for image
  matching.
\newblock In: Proc IEEE/RSJ Int Conf Intell Robots Syst, 2025.
\newblock 2448--2454

\bibitem{guo2023descriptor}
Guo X, Hu J, Bao H,  et~al.
\newblock Descriptor distillation for efficient multi-robot {SLAM}.
\newblock In: Proc IEEE Int Conf Robot Autom, 2023.
\newblock 6210--6216

\bibitem{liu2023taskaligned}
Liu C, Liu H, Zhang L,  et~al.
\newblock Learning task-aligned local features for visual localization.
\newblock IEEE Robot Autom Lett, 2023, 8: 3366--3373

\bibitem{santellani2024gmm}
Santellani E, Zach M, Sormann C,  et~al.
\newblock {GMM-IKRS}: {Gaussian} mixture models for interpretable keypoint
  refinement and scoring.
\newblock In: Proc Eur Conf Comput Vis, 2024.
\newblock 77--93

\bibitem{kim2024learning}
Kim S, Pollefeys M,  Barath D.
\newblock Learning to make keypoints sub-pixel accurate.
\newblock In: Proc Eur Conf Comput Vis, 2024.
\newblock 413--431

\bibitem{schmid2026xrefine}
Schmid J~F,  Hagemann A.
\newblock {XRefine}: attention-guided keypoint match refinement.
\newblock In: Proc Int Conf 3D Vis, 2026.
\newblock 166--174

\bibitem{qi2017pointnet}
Qi C~R, Su H, Mo K,  et~al.
\newblock {PointNet}: deep learning on point sets for {3D} classification and
  segmentation.
\newblock In: Proc IEEE Conf Comput Vis Pattern Recognit, 2017.
\newblock 652--660

\bibitem{ma2019lmr}
Ma J, Jiang X, Jiang J,  et~al.
\newblock {LMR}: learning a two-class classifier for mismatch removal.
\newblock IEEE Trans Image Process, 2019, 28: 4045--4059

\bibitem{zhang2019learning}
Zhang J, Sun D, Luo Z,  et~al.
\newblock Learning two-view correspondences and geometry using order-aware
  network.
\newblock In: Proc IEEE/CVF Int Conf Comput Vis, 2019.
\newblock 5845--5854

\bibitem{zhang2020oanet}
Zhang J, Sun D, Luo Z,  et~al.
\newblock {OANet}: learning two-view correspondences and geometry using
  order-aware network.
\newblock IEEE Trans Pattern Anal Mach Intell, 2022, 44: 3110--3122

\bibitem{sun2020acne}
Sun W, Jiang W, Trulls E,  et~al.
\newblock {ACNe}: attentive context normalization for robust
  permutation-equivariant learning.
\newblock In: Proc IEEE/CVF Conf Comput Vis Pattern Recognit, 2020.
\newblock 11286--11295

\bibitem{xiao2024t}
Xiao G, Liu X, Zhong Z,  et~al.
\newblock {T-Net++}: effective permutation-equivariance network for two-view
  correspondence pruning.
\newblock IEEE Trans Pattern Anal Mach Intell, 2024, 46: 10629--10644

\bibitem{liu2021learnable}
Liu Y, Liu L, Lin C,  et~al.
\newblock Learnable motion coherence for correspondence pruning.
\newblock In: Proc IEEE/CVF Conf Comput Vis Pattern Recognit, 2021.
\newblock 3237--3246

\bibitem{zhao2019nm}
Zhao C, Cao Z, Li C,  et~al.
\newblock {NM-Net}: mining reliable neighbors for robust feature
  correspondences.
\newblock In: Proc IEEE/CVF Conf Comput Vis Pattern Recognit, 2019.
\newblock 215--224

\bibitem{zhao2021progressive}
Zhao C, Ge Y, Zhu F,  et~al.
\newblock Progressive correspondence pruning by consensus learning.
\newblock In: Proc IEEE/CVF Int Conf Comput Vis, 2021.
\newblock 6464--6473

\bibitem{kipf2017semi}
Kipf T~N,  Welling M.
\newblock Semi-supervised classification with graph convolutional networks.
\newblock In: Proc Int Conf Learn Represent, 2017

\bibitem{dai2022ms2dg}
Dai L, Liu Y, Ma J,  et~al.
\newblock {MS2DG-Net}: progressive correspondence learning via multiple sparse
  semantics dynamic graph.
\newblock In: Proc IEEE/CVF Conf Comput Vis Pattern Recognit, 2022.
\newblock 8973--8982

\bibitem{liu2023progressive}
Liu X,  Yang J.
\newblock Progressive neighbor consistency mining for correspondence pruning.
\newblock In: Proc IEEE/CVF Conf Comput Vis Pattern Recognit, 2023.
\newblock 9527--9537

\bibitem{luanyuan2024mgnet}
Dai L, Du X, Zhang H,  et~al.
\newblock {MGNet}: learning correspondences via multiple graphs.
\newblock In: Proc AAAI Conf Artif Intell, 2024.
\newblock 3945--3953

\bibitem{vaswani2017attention}
Vaswani A, Shazeer N, Parmar N,  et~al.
\newblock Attention is all you need.
\newblock In: Adv Neural Inf Process Syst, 2017.
\newblock 5998--6008

\bibitem{jiang2022learning}
Jiang X, Wang Y, Fan A,  et~al.
\newblock Learning for mismatch removal via graph attention networks.
\newblock ISPRS J Photogramm Remote Sens, 2022, 190: 181--195

\bibitem{ye2023learning}
Ye X, Zhao W, Lu H,  et~al.
\newblock Learning second-order attentive context for efficient correspondence
  pruning.
\newblock In: Proc AAAI Conf Artif Intell, 2023.
\newblock 3250--3258

\bibitem{li2023u}
Li Z, Zhang S,  Ma J.
\newblock {U-Match}: two-view correspondence learning with hierarchy-aware
  local context aggregation.
\newblock In: Proc Int Jt Conf Artif Intell, 2023.
\newblock 1169--1176

\bibitem{gao2019graph}
Gao H,  Ji S.
\newblock Graph {U-Nets}.
\newblock In: Proc Int Conf Mach Learn, 2019.
\newblock 2083--2092

\bibitem{li2024u}
Li Z, Zhang S,  Ma J.
\newblock {U-Match}: exploring hierarchy-aware local context for two-view
  correspondence learning.
\newblock IEEE Trans Pattern Anal Mach Intell, 2024, 46: 10960--10977

\bibitem{miao2024bclnet}
Miao X, Xiao G, Wang S,  et~al.
\newblock {BCLNet}: bilateral consensus learning for two-view correspondence
  pruning.
\newblock In: Proc AAAI Conf Artif Intell, 2024.
\newblock 4225--4232

\bibitem{liao2024vsformer}
Liao T, Zhang X, Zhao L,  et~al.
\newblock {VSFormer}: visual-spatial fusion transformer for correspondence
  pruning.
\newblock In: Proc AAAI Conf Artif Intell, 2024.
\newblock 3369--3377

\bibitem{dai2026lecot}
Dai L, Du X,  Tang J.
\newblock {LeCoT}: revisiting network architecture for two-view correspondence
  pruning.
\newblock Sci China Inf Sci, 2026, 69: 112104

\bibitem{dai2026clct}
Dai L, Yan R, Tang J,  et~al.
\newblock {CLCT}: complementary local consensus transformer for two-view
  correspondence pruning.
\newblock IEEE Trans Multimedia, 2026, doi: 10.1109/TMM.2026.3687024

\bibitem{fang2026corrmamba}
Fang X, Zhang S, Zhang H,  et~al.
\newblock Selecting and pruning: a differentiable causal sequentialized
  state-space model for two-view correspondence learning.
\newblock IEEE Trans Image Process, 2026, 35: 816--829

\bibitem{liao2026scalable}
Liao T, Zhang X, Wang T,  et~al.
\newblock Scalable and generalizable correspondence pruning via
  geometry-consistent pre-training.
\newblock IEEE Trans Pattern Anal Mach Intell, 2026, 48: 10048--10065

\bibitem{zhang2023convmatch}
Zhang S,  Ma J.
\newblock {ConvMatch}: rethinking network design for two-view correspondence
  learning.
\newblock In: Proc AAAI Conf Artif Intell, 2023.
\newblock 3472--3479

\bibitem{zhang2024convmatch}
Zhang S,  Ma J.
\newblock {ConvMatch}: rethinking network design for two-view correspondence
  learning.
\newblock IEEE Trans Pattern Anal Mach Intell, 2024, 46: 2920--2935

\bibitem{lu2025demo}
Lu Y, Le J, Li Z,  et~al.
\newblock {DeMo}: deep motion field consensus with learnable kernels for
  two-view correspondence learning.
\newblock In: Proc AAAI Conf Artif Intell, 2025.
\newblock 5829--5837

\bibitem{zhang2024dematch}
Zhang S, Li Z, Gao Y,  et~al.
\newblock {DeMatch}: deep decomposition of motion field for two-view
  correspondence learning.
\newblock In: Proc IEEE/CVF Conf Comput Vis Pattern Recognit, 2024.
\newblock 20278--20287

\bibitem{zhang2025dematchpp}
Zhang S, Li Z,  Ma J.
\newblock {DeMatch++}: two-view correspondence learning via deep motion field
  decomposition and respective local-context aggregation.
\newblock IEEE Trans Pattern Anal Mach Intell, 2025, 47: 11234--11251

\bibitem{li2025corrnext}
Li Z, Su C, Fan F,  et~al.
\newblock {CorrNeXt}: making the {ConvNet}-style correspondence pruner stronger
  for two-view geometry.
\newblock In: Proc ACM Int Conf Multimed, 2025.
\newblock 4203--4212

\bibitem{lin2026scnet}
Lin S, Liao H, Qi Q,  et~al.
\newblock {SC-Net}: robust correspondence learning via spatial and
  cross-channel context.
\newblock In: Proc AAAI Conf Artif Intell, 2026.
\newblock 6979--6987

\bibitem{le2026geomoe}
Le J,  Ma J.
\newblock {GeoMoE}: divide-and-conquer motion field modeling with
  mixture-of-experts for two-view geometry.
\newblock In: Proc AAAI Conf Artif Intell, 2026.
\newblock 5845--5853

\bibitem{brachmann2019neural}
Brachmann E,  Rother C.
\newblock Neural-guided {RANSAC}: learning where to sample model hypotheses.
\newblock In: Proc IEEE/CVF Int Conf Comput Vis, 2019.
\newblock 4322--4331

\bibitem{ranftl2018deep}
Ranftl R,  Koltun V.
\newblock Deep fundamental matrix estimation.
\newblock In: Proc Eur Conf Comput Vis, 2018.
\newblock 292--309

\bibitem{brachmann2017dsac}
Brachmann E, Krull A, Nowozin S,  et~al.
\newblock {DSAC} -- differentiable {RANSAC} for camera localization.
\newblock In: Proc IEEE Conf Comput Vis Pattern Recognit, 2017.
\newblock 6684--6692

\bibitem{miao2024survey}
Miao J, Jiang K, Wen T,  et~al.
\newblock A survey on monocular re-localization: from the perspective of scene
  map representation.
\newblock IEEE Trans Intell Veh, 2025, 10: 2519--2550

\bibitem{wei2023adaptive}
Wei T, Matas J,  Barath D.
\newblock Adaptive reordering sampler with neurally guided {MAGSAC}.
\newblock In: Proc IEEE/CVF Int Conf Comput Vis, 2023.
\newblock 18163--18173

\bibitem{piedade2023bansac}
Piedade V,  Miraldo P.
\newblock {BANSAC}: a dynamic {Bayesian} network for adaptive sample consensus.
\newblock In: Proc IEEE/CVF Int Conf Comput Vis, 2023.
\newblock 3738--3747

\bibitem{wei2023generalized}
Wei T, Patel Y, Shekhovtsov A,  et~al.
\newblock Generalized differentiable {RANSAC}.
\newblock In: Proc IEEE/CVF Int Conf Comput Vis, 2023.
\newblock 17649--17660

\bibitem{jang2017categorical}
Jang E, Gu S,  Poole B.
\newblock Categorical reparameterization with {Gumbel-Softmax}.
\newblock In: Proc Int Conf Learn Represent, 2017

\bibitem{wang2026mcd}
Wang J, Zhao C, Ke W,  et~al.
\newblock {Monte Carlo} diffusion for generalizable learning-based {RANSAC}.
\newblock In: Proc AAAI Conf Artif Intell, 2026.
\newblock 9894--9902

\bibitem{barath2022learning}
Barath D, Cavalli L,  Pollefeys M.
\newblock Learning to find good models in {RANSAC}.
\newblock In: Proc IEEE/CVF Conf Comput Vis Pattern Recognit, 2022.
\newblock 15744--15753

\bibitem{cavalli2022nefsac}
Cavalli L, Pollefeys M,  Barath D.
\newblock {NeFSAC}: neurally filtered minimal samples.
\newblock In: Proc Eur Conf Comput Vis, 2022.
\newblock 351--366

\bibitem{barroso2023two}
Barroso-Laguna A, Brachmann E, Prisacariu V~A,  et~al.
\newblock Two-view geometry scoring without correspondences.
\newblock In: Proc IEEE/CVF Conf Comput Vis Pattern Recognit, 2023.
\newblock 8979--8989

\bibitem{probst2019unsupervised}
Probst T, Paudel D~P, Chhatkuli A,  et~al.
\newblock Unsupervised learning of consensus maximization for {3D} vision
  problems.
\newblock In: Proc IEEE/CVF Conf Comput Vis Pattern Recognit, 2019.
\newblock 929--938

\bibitem{hu2012fast}
Hu Y, Zhang D, Ye J,  et~al.
\newblock Fast and accurate matrix completion via truncated nuclear norm
  regularization.
\newblock IEEE Trans Pattern Anal Mach Intell, 2013, 35: 2117--2130

\bibitem{truong2021unsupervised}
Truong G, Le H, Suter D,  et~al.
\newblock Unsupervised learning for robust fitting: a reinforcement learning
  approach.
\newblock In: Proc IEEE/CVF Conf Comput Vis Pattern Recognit, 2021.
\newblock 10348--10357

\bibitem{mnih2013playing}
Mnih V, Kavukcuoglu K, Silver D,  et~al.
\newblock Playing {Atari} with deep reinforcement learning.
\newblock In: Adv Neural Inf Process Syst Workshops, 2013

\bibitem{wang2019dynamic}
Wang Y, Sun Y, Liu Z,  et~al.
\newblock Dynamic graph {CNN} for learning on point clouds.
\newblock ACM Trans Graph, 2019, 38: 146

\bibitem{truong2022unsupervised}
Truong G, Le H, Zhang E,  et~al.
\newblock Unsupervised learning for maximum consensus robust fitting: a
  reinforcement learning approach.
\newblock IEEE Trans Pattern Anal Mach Intell, 2023, 45: 3890--3903

\bibitem{nie2023rlsac}
Nie C, Wang G, Liu Z,  et~al.
\newblock {RLSAC}: reinforcement learning enhanced sample consensus for
  end-to-end robust estimation.
\newblock In: Proc IEEE/CVF Int Conf Comput Vis, 2023.
\newblock 9891--9900

\bibitem{sarlin2020superglue}
Sarlin P~E, DeTone D, Malisiewicz T,  et~al.
\newblock {SuperGlue}: learning feature matching with graph neural networks.
\newblock In: Proc IEEE/CVF Conf Comput Vis Pattern Recognit, 2020.
\newblock 4938--4947

\bibitem{chen2021learning}
Chen H, Luo Z, Zhang J,  et~al.
\newblock Learning to match features with seeded graph matching network.
\newblock In: Proc IEEE/CVF Int Conf Comput Vis, 2021.
\newblock 6301--6310

\bibitem{shi2022clustergnn}
Shi Y, Cai J~X, Shavit Y,  et~al.
\newblock {ClusterGNN}: cluster-based coarse-to-fine graph neural network for
  efficient feature matching.
\newblock In: Proc IEEE/CVF Conf Comput Vis Pattern Recognit, 2022.
\newblock 12517--12526

\bibitem{lindenberger2023lightglue}
Lindenberger P, Sarlin P~E,  Pollefeys M.
\newblock {LightGlue}: local feature matching at light speed.
\newblock In: Proc IEEE/CVF Int Conf Comput Vis, 2023.
\newblock 17627--17638

\bibitem{jiang2024omniglue}
Jiang H, Karpur A, Cao B,  et~al.
\newblock {OmniGlue}: generalizable feature matching with foundation model
  guidance.
\newblock In: Proc IEEE/CVF Conf Comput Vis Pattern Recognit, 2024.
\newblock 19865--19875

\bibitem{wu2020comprehensive}
Wu Z, Pan S, Chen F,  et~al.
\newblock A comprehensive survey on graph neural networks.
\newblock IEEE Trans Neural Netw Learn Syst, 2021, 32: 4--24

\bibitem{cuturi2013sinkhorn}
Cuturi M.
\newblock {Sinkhorn} distances: lightspeed computation of optimal transport.
\newblock In: Adv Neural Inf Process Syst, 2013.
\newblock 2292--2300

\bibitem{lu2023paraformer}
Lu X, Yan Y, Kang B,  et~al.
\newblock {ParaFormer}: parallel attention transformer for efficient feature
  matching.
\newblock In: Proc AAAI Conf Artif Intell, 2023.
\newblock 1853--1860

\bibitem{xue2023imp}
Xue F, Budvytis I,  Cipolla R.
\newblock {IMP}: iterative matching and pose estimation with adaptive pooling.
\newblock In: Proc IEEE/CVF Conf Comput Vis Pattern Recognit, 2023.
\newblock 21317--21326

\bibitem{su2024roformer}
Su J, Ahmed M, Lu Y,  et~al.
\newblock {RoFormer}: enhanced transformer with rotary position embedding.
\newblock Neurocomputing, 2024, 568: 127063

\bibitem{li2024learning}
Li Z,  Ma J.
\newblock Learning feature matching via matchable keypoint-assisted graph
  neural network.
\newblock IEEE Trans Image Process, 2025, 34: 154--169

\bibitem{ryoo2025mambaglue}
Ryoo K, Lim H,  Myung H.
\newblock {MambaGlue}: fast and robust local feature matching with {Mamba}.
\newblock In: Proc IEEE Int Conf Robot Autom, 2025.
\newblock 5758--5765

\bibitem{gu2024mamba}
Gu A,  Dao T.
\newblock {Mamba}: linear-time sequence modeling with selective state spaces.
\newblock In: Proc Conf Lang Model, 2024

\bibitem{lu2023scene}
Lu X, Yan Y, Wei T,  et~al.
\newblock Scene-aware feature matching.
\newblock In: Proc IEEE/CVF Int Conf Comput Vis, 2023.
\newblock 3704--3713

\bibitem{du2026sceneglue}
Du S, Lu X, Yan Y,  et~al.
\newblock {SceneGlue}: scene-aware transformer for feature matching without
  scene-level annotation.
\newblock IEEE Trans Circuits Syst Video Technol, 2026, 36: 11844--11857

\bibitem{deng2024resmatch}
Deng Y, Zhang K, Zhang S,  et~al.
\newblock {ResMatch}: residual attention learning for feature matching.
\newblock In: Proc AAAI Conf Artif Intell, 2024.
\newblock 1501--1509

\bibitem{liu2026cfm}
Liu X, Han Y, Qin R,  et~al.
\newblock Collaborative feature matching with progressive correspondence
  learning.
\newblock In: Proc AAAI Conf Artif Intell, 2026.
\newblock 7314--7322

\bibitem{zhang2026sgat}
Zhang Y, Sun K, Tang C,  et~al.
\newblock {SGAT}: learning feature matching with singularity-enhanced graph
  attention network.
\newblock In: Proc AAAI Conf Artif Intell, 2026.
\newblock 12952--12960

\bibitem{oquab2024dinov2}
Oquab M, Darcet T, Moutakanni T,  et~al.
\newblock {DINOv2}: learning robust visual features without supervision.
\newblock Trans Mach Learn Res, 2024

\bibitem{zhang2025matching}
Zhang S, Zhu Z, Li Z,  et~al.
\newblock Matching while perceiving: enhance image feature matching with
  applicable semantic amalgamation.
\newblock In: Proc AAAI Conf Artif Intell, 2025.
\newblock 10094--10102

\bibitem{guo2022segnext}
Guo M~H, Lu C~Z, Hou Q,  et~al.
\newblock {SegNeXt}: rethinking convolutional attention design for semantic
  segmentation.
\newblock In: Adv Neural Inf Process Syst, 2022.
\newblock 1140--1156

\bibitem{wu2025mapglue}
Wu P, Yao Y, Zhang W,  et~al.
\newblock {MapGlue}: multimodal remote sensing image matching.
\newblock ISPRS J Photogramm Remote Sens, 2026, 240: 166--183

\bibitem{zhang2024diffglue}
Zhang S,  Ma J.
\newblock {DiffGlue}: diffusion-aided image feature matching.
\newblock In: Proc ACM Int Conf Multimed, 2024.
\newblock 8451--8460

\bibitem{croitoru2023diffusion}
Croitoru F~A, Hondru V, Ionescu R~T,  et~al.
\newblock Diffusion models in vision: a survey.
\newblock IEEE Trans Pattern Anal Mach Intell, 2023, 45: 10850--10869

\bibitem{xu2026matchany}
Xu H, Pan R, Zhou J,  et~al.
\newblock Match any keypoints.
\newblock IEEE Trans Image Process, 2026, 35: 4051--4063

\bibitem{nordstrom2026loma}
Nordstr{\"o}m D, Edstedt J, B{\"o}kman G,  et~al.
\newblock {LoMa}: local feature matching revisited.
\newblock In: Proc Eur Conf Comput Vis, 2026.
\newblock 637--655

\bibitem{he2025matchanything}
He X, Yu H, Peng S,  et~al.
\newblock Multiple modalities image matching with large-scale pre-training.
\newblock IEEE Trans Pattern Anal Mach Intell, 2026, doi:
  10.1109/TPAMI.2026.3724652

\bibitem{rocco2018neighbourhood}
Rocco I, Cimpoi M, Arandjelovi{\'c} R,  et~al.
\newblock Neighbourhood consensus networks.
\newblock In: Adv Neural Inf Process Syst, 2018.
\newblock 1651--1662

\bibitem{rocco2020efficient}
Rocco I, Arandjelovi{\'c} R,  Sivic J.
\newblock Efficient neighbourhood consensus networks via submanifold sparse
  convolutions.
\newblock In: Proc Eur Conf Comput Vis, 2020.
\newblock 605--621

\bibitem{zhou2021patch2pix}
Zhou Q, Sattler T,  Leal-Taix{\'e} L.
\newblock {Patch2Pix}: epipolar-guided pixel-level correspondences.
\newblock In: Proc IEEE/CVF Conf Comput Vis Pattern Recognit, 2021.
\newblock 4669--4678

\bibitem{li2020dual}
Li X, Han K, Li S,  et~al.
\newblock Dual-resolution correspondence networks.
\newblock In: Adv Neural Inf Process Syst, 2020.
\newblock 17346--17357

\bibitem{li2023dualrc}
Li X, Han K, Li S,  et~al.
\newblock {DualRC}: a dual-resolution learning framework with neighbourhood
  consensus for visual correspondences.
\newblock IEEE Trans Pattern Anal Mach Intell, 2024, 46: 236--249

\bibitem{he2023efficient}
He J, Zhang T, Zhang Z,  et~al.
\newblock Efficient dynamic correspondence network.
\newblock IEEE Trans Image Process, 2024, 33: 228--240

\bibitem{sun2021loftr}
Sun J, Shen Z, Wang Y,  et~al.
\newblock {LoFTR}: detector-free local feature matching with transformers.
\newblock In: Proc IEEE/CVF Conf Comput Vis Pattern Recognit, 2021.
\newblock 8922--8931

\bibitem{lin2017feature}
Lin T~Y, Doll{\'a}r P, Girshick R,  et~al.
\newblock Feature pyramid networks for object detection.
\newblock In: Proc IEEE Conf Comput Vis Pattern Recognit, 2017.
\newblock 2117--2125

\bibitem{katharopoulos2020transformers}
Katharopoulos A, Vyas A, Pappas N,  et~al.
\newblock Transformers are {RNNs}: fast autoregressive transformers with linear
  attention.
\newblock In: Proc Int Conf Mach Learn, 2020.
\newblock 5156--5165

\bibitem{parmar2018image}
Parmar N, Vaswani A, Uszkoreit J,  et~al.
\newblock Image transformer.
\newblock In: Proc Int Conf Mach Learn, 2018.
\newblock 4055--4064

\bibitem{wang2022matchformer}
Wang Q, Zhang J, Yang K,  et~al.
\newblock {MatchFormer}: interleaving attention in transformers for feature
  matching.
\newblock In: Proc Asian Conf Comput Vis, 2022.
\newblock 256--273

\bibitem{chen2022aspanformer}
Chen H, Luo Z, Zhou L,  et~al.
\newblock {ASpanFormer}: detector-free image matching with adaptive span
  transformer.
\newblock In: Proc Eur Conf Comput Vis, 2022.
\newblock 20--36

\bibitem{chen2024affine}
Chen H, Luo Z, Tian Y,  et~al.
\newblock Affine-based deformable attention and selective fusion for semi-dense
  matching.
\newblock In: Proc IEEE/CVF Conf Comput Vis Pattern Recognit Workshops, 2024.
\newblock 4254--4263

\bibitem{mao20223dg}
Mao R, Bai C, An Y,  et~al.
\newblock {3DG-STFM}: {3D} geometric guided student-teacher feature matching.
\newblock In: Proc Eur Conf Comput Vis, 2022.
\newblock 125--142

\bibitem{wang2023guiding}
Wang S, Kannala J, Pollefeys M,  et~al.
\newblock Guiding local feature matching with surface curvature.
\newblock In: Proc IEEE/CVF Int Conf Comput Vis, 2023.
\newblock 17981--17991

\bibitem{ranftl2021vision}
Ranftl R, Bochkovskiy A,  Koltun V.
\newblock Vision transformers for dense prediction.
\newblock In: Proc IEEE/CVF Int Conf Comput Vis, 2021.
\newblock 12179--12188

\bibitem{chang2023sem}
Chang J, Yu J,  Zhang T.
\newblock Structured epipolar matcher for local feature matching.
\newblock In: Proc IEEE/CVF Conf Comput Vis Pattern Recognit Workshops, 2023.
\newblock 6177--6186

\bibitem{leroy2024grounding}
Leroy V, Cabon Y,  Revaud J.
\newblock Grounding image matching in {3D} with {MASt3R}.
\newblock In: Proc Eur Conf Comput Vis, 2024.
\newblock 71--91

\bibitem{wang2024dust3r}
Wang S, Leroy V, Cabon Y,  et~al.
\newblock {DUSt3R}: geometric {3D} vision made easy.
\newblock In: Proc IEEE/CVF Conf Comput Vis Pattern Recognit, 2024.
\newblock 20697--20709

\bibitem{fang2026sigma}
Fang X, Li Z,  Ma J.
\newblock {SigMa}: semantic similarity-guided semi-dense feature matching.
\newblock IEEE Trans Image Process, 2026, 35: 872--887

\bibitem{liu2025mindgap}
Liu Y, Fu J, Wu Y,  et~al.
\newblock Mind the gap: aligning vision foundation models to image feature
  matching.
\newblock In: Proc IEEE/CVF Int Conf Comput Vis, 2025.
\newblock 20313--20323

\bibitem{zheng2026textfm}
Zheng Z, Feng J, Savaliya N,  et~al.
\newblock {TextFM}: robust semi-dense feature matching with language guidance.
\newblock In: Proc IEEE/CVF Conf Comput Vis Pattern Recognit, 2026.
\newblock 16635--16644

\bibitem{giang2024topicfm+}
Giang K~T, Song S,  Jo S.
\newblock {TopicFM+}: boosting accuracy and efficiency of topic-assisted
  feature matching.
\newblock IEEE Trans Image Process, 2024, 33: 6016--6028

\bibitem{cao2023improving}
Cao C,  Fu Y.
\newblock Improving transformer-based image matching by cascaded capturing
  spatially informative keypoints.
\newblock In: Proc IEEE/CVF Int Conf Comput Vis, 2023.
\newblock 12129--12139

\bibitem{chen2025casp}
Chen P, Yu L, Wan Y,  et~al.
\newblock {CasP}: improving semi-dense feature matching pipeline leveraging
  cascaded correspondence priors for guidance.
\newblock In: Proc IEEE/CVF Int Conf Comput Vis, 2025.
\newblock 28063--28072

\bibitem{huang2023adaptive}
Huang D, Chen Y, Liu Y,  et~al.
\newblock Adaptive assignment for geometry aware local feature matching.
\newblock In: Proc IEEE/CVF Conf Comput Vis Pattern Recognit, 2023.
\newblock 5425--5434

\bibitem{li2025comatch}
Li Z, Lu Y, Tang L,  et~al.
\newblock {CoMatch}: dynamic covisibility-aware transformer for bilateral
  subpixel-level semi-dense image matching.
\newblock In: Proc IEEE/CVF Int Conf Comput Vis, 2025.
\newblock 18521--18530

\bibitem{ni2023pats}
Ni J, Li Y, Huang Z,  et~al.
\newblock {PATS}: patch area transportation with subdivision for local feature
  matching.
\newblock In: Proc IEEE/CVF Conf Comput Vis Pattern Recognit, 2023.
\newblock 17776--17786

\bibitem{yu2023adaptive}
Yu J, Chang J, He J,  et~al.
\newblock Adaptive spot-guided transformer for consistent local feature
  matching.
\newblock In: Proc IEEE/CVF Conf Comput Vis Pattern Recognit, 2023.
\newblock 21898--21908

\bibitem{jin2026entropy}
Jin K, Chen J,  Ye Q.
\newblock Improving local feature matching by entropy-inspired scale
  adaptability and flow-endowed local consistency.
\newblock IEEE Trans Circuits Syst Video Technol, 2026, 36: 10957--10968

\bibitem{cai2024prism}
Cai X, Wang Y, Luo L,  et~al.
\newblock {PRISM}: progressive dependency maximization for scale-invariant
  image matching.
\newblock In: Proc ACM Int Conf Multimed, 2024.
\newblock 5250--5259

\bibitem{wang2025homomatcher}
Wang X, Yu L, Zhang Y,  et~al.
\newblock {HomoMatcher}: achieving dense feature matching with semi-dense
  efficiency by homography estimation.
\newblock In: Proc AAAI Conf Artif Intell, 2025.
\newblock 7952--7960

\bibitem{tang2022quadtree}
Tang S, Zhang J, Zhu S,  et~al.
\newblock {QuadTree} attention for vision transformers.
\newblock In: Proc Int Conf Learn Represent, 2022

\bibitem{giang2023topicfm}
Giang K~T, Song S,  Jo S.
\newblock {TopicFM}: robust and interpretable topic-assisted feature matching.
\newblock In: Proc AAAI Conf Artif Intell, 2023.
\newblock 2447--2455

\bibitem{chen2025ecomatcher}
Chen P, Yu L, Wan Y,  et~al.
\newblock {EcoMatcher}: efficient clustering oriented matcher for detector-free
  image matching.
\newblock In: Proc Eur Conf Comput Vis, 2024.
\newblock 344--360

\bibitem{wang2024efficient}
Wang Y, He X, Peng S,  et~al.
\newblock Efficient {LoFTR}: semi-dense local feature matching with sparse-like
  speed.
\newblock In: Proc IEEE/CVF Conf Comput Vis Pattern Recognit, 2024.
\newblock 21666--21675

\bibitem{ding2021repvgg}
Ding X, Zhang X, Ma N,  et~al.
\newblock {RepVGG}: making {VGG}-style {ConvNets} great again.
\newblock In: Proc IEEE/CVF Conf Comput Vis Pattern Recognit, 2021.
\newblock 13733--13742

\bibitem{ni2024etoefficient}
Ni J, Zhang G, Li G,  et~al.
\newblock {ETO}: efficient transformer-based local feature matching by
  organizing multiple homography hypotheses.
\newblock In: Adv Neural Inf Process Syst, 2024.
\newblock 60260--60274

\bibitem{ni2025etoplus}
Ni J, Shen Y, Li Y,  et~al.
\newblock {ETO+}: revisit the refinement stage in efficient feature matching.
\newblock In: Proc IEEE/RSJ Int Conf Intell Robots Syst, 2025.
\newblock 18806--18813

\bibitem{lu2025jamma}
Lu X,  Du S.
\newblock {JamMa}: ultra-lightweight local feature matching with joint {Mamba}.
\newblock In: Proc IEEE/CVF Conf Comput Vis Pattern Recognit, 2025.
\newblock 14934--14943

\bibitem{li2026pyramatch}
Li Z, Jin R, Sekikawa Y,  et~al.
\newblock {PyraMatch}: multi-head pyramid scan for {Mamba}-based image
  matching.
\newblock In: Proc IEEE Int Conf Acoust Speech Signal Process, 2026.
\newblock 11417--11421

\bibitem{lu2026pmhi}
Lu X, Du S, Xiao G,  et~al.
\newblock Progressive matchability guided hybrid interaction for
  latency-balanced local feature matching.
\newblock IEEE Trans Multimedia, 2026, doi: 10.1109/TMM.2026.3723600

\bibitem{choo2026slim}
Choo S~W,  Li B.
\newblock Scalable feature matching via state space modeling and sparse
  correlation.
\newblock In: Proc IEEE/CVF Conf Comput Vis Pattern Recognit, 2026.
\newblock 6685--6694

\bibitem{toida2026covisibility}
Toida K,  Hotta K.
\newblock Covisibility-aware token pruning for scalable semi-dense image
  matching.
\newblock In: Proc IEEE/CVF Conf Comput Vis Pattern Recognit Workshops, 2026.
\newblock 149--158

\bibitem{le2026ultramatch}
Le J, Lu Y, Li Z,  et~al.
\newblock {UltraMatch}: transport path routing for ultra-fast and
  memory-efficient image matching.
\newblock ArXiv:2609.36980, 2026

\bibitem{li2025edm}
Li X, Rao T,  Pan C.
\newblock {EDM}: efficient deep feature matching.
\newblock In: Proc IEEE/CVF Int Conf Comput Vis, 2025.
\newblock 26198--26208

\bibitem{li2026sure}
Li S, Gu Z, Zhang J,  et~al.
\newblock {SURE}: semi-dense uncertainty-refined feature matching.
\newblock In: Proc IEEE Int Conf Robot Autom, 2026

\bibitem{sekikawa2026det}
Sekikawa Y, Shirai H, Eto R,  et~al.
\newblock {Detect by Track}: making detector-free matcher trackable.
\newblock In: Proc Eur Conf Comput Vis, 2026.
\newblock 231--250

\bibitem{chen2026matchgs}
Chen J, Xu C,  Cao Y.
\newblock Unlocking zero-shot potential of semi-dense image matching via
  {Gaussian} splatting.
\newblock In: Proc IEEE/RSJ Int Conf Intell Robots Syst, 2026

\bibitem{tuzcuoglu2024xoftr}
Tuzcuo{\u{g}}lu {\"O}, K{\"o}ksal A, Sofu B,  et~al.
\newblock {XoFTR}: cross-modal feature matching transformer.
\newblock In: Proc IEEE/CVF Conf Comput Vis Pattern Recognit Workshops, 2024.
\newblock 4275--4286

\bibitem{dosovitskiy2015flownet}
Dosovitskiy A, Fischer P, Ilg E,  et~al.
\newblock {FlowNet}: learning optical flow with convolutional networks.
\newblock In: Proc IEEE Int Conf Comput Vis, 2015.
\newblock 2758--2766

\bibitem{rocco2017convolutional}
Rocco I, Arandjelovi{\'c} R,  Sivic J.
\newblock Convolutional neural network architecture for geometric matching.
\newblock In: Proc IEEE Conf Comput Vis Pattern Recognit, 2017.
\newblock 6148--6157

\bibitem{melekhov2019dgc}
Melekhov I, Tiulpin A, Sattler T,  et~al.
\newblock {DGC-Net}: dense geometric correspondence network.
\newblock In: Proc IEEE Winter Conf Appl Comput Vis, 2019.
\newblock 1034--1042

\bibitem{truong2020glu}
Truong P, Danelljan M,  Timofte R.
\newblock {GLU-Net}: global-local universal network for dense flow and
  correspondences.
\newblock In: Proc IEEE/CVF Conf Comput Vis Pattern Recognit, 2020.
\newblock 6258--6268

\bibitem{truong2020gocor}
Truong P, Danelljan M, Van~Gool L,  et~al.
\newblock {GOCor}: bringing globally optimized correspondence volumes into your
  neural network.
\newblock In: Adv Neural Inf Process Syst, 2020.
\newblock 14278--14290

\bibitem{shen2020ransac}
Shen X, Darmon F, Efros A~A,  et~al.
\newblock {RANSAC-Flow}: generic two-stage image alignment.
\newblock In: Proc Eur Conf Comput Vis, 2020.
\newblock 618--637

\bibitem{truong2021warp}
Truong P, Danelljan M, Yu F,  et~al.
\newblock Warp consistency for unsupervised learning of dense correspondences.
\newblock In: Proc IEEE/CVF Int Conf Comput Vis, 2021.
\newblock 10346--10356

\bibitem{truong2021learning}
Truong P, Danelljan M, Van~Gool L,  et~al.
\newblock Learning accurate dense correspondences and when to trust them.
\newblock In: Proc IEEE/CVF Conf Comput Vis Pattern Recognit, 2021.
\newblock 5714--5724

\bibitem{edstedt2023dkm}
Edstedt J, Athanasiadis I, Wadenb{\"a}ck M,  et~al.
\newblock {DKM}: dense kernelized feature matching for geometry estimation.
\newblock In: Proc IEEE/CVF Conf Comput Vis Pattern Recognit, 2023.
\newblock 17765--17775

\bibitem{zhu2023pmatch}
Zhu S,  Liu X.
\newblock {PMatch}: paired masked image modeling for dense geometric matching.
\newblock In: Proc IEEE/CVF Conf Comput Vis Pattern Recognit, 2023.
\newblock 21909--21918

\bibitem{edstedt2024roma}
Edstedt J, Sun Q, B{\"o}kman G,  et~al.
\newblock {RoMa}: robust dense feature matching.
\newblock In: Proc IEEE/CVF Conf Comput Vis Pattern Recognit, 2024.
\newblock 19790--19800

\bibitem{edstedt2026romav2}
Edstedt J, Nordstr{\"o}m D, Zhang Y,  et~al.
\newblock {RoMa v2}: harder better faster denser feature matching.
\newblock In: Proc Eur Conf Comput Vis, 2026.
\newblock 664--682

\bibitem{nordstrom2026romaomega}
Nordstr{\"o}m D, Zhang X, Loiseau T,  et~al.
\newblock {RoMa-$\Omega$}: what feed-forward {3D} models know about image
  matching.
\newblock ArXiv:2609.09507, 2026

\bibitem{shao2026spider}
Shao Z, Yadav A, Chellappa R,  et~al.
\newblock {SPIDER}: spatial image correspondence estimator for robust
  calibration.
\newblock In: Proc IEEE/CVF Conf Comput Vis Pattern Recognit Findings, 2026.
\newblock 253--263

\bibitem{lee2026mvroma}
Lee J, Kang S,  Yoo S.
\newblock {MV-RoMa}: from pairwise matching into multi-view track
  reconstruction.
\newblock In: Proc IEEE/CVF Conf Comput Vis Pattern Recognit, 2026.
\newblock 7446--7456

\bibitem{cai2025evidential}
Cai L, Foo C~S, Xu X,  et~al.
\newblock Evidential learning-based certainty estimation for robust dense
  feature matching.
\newblock In: Proc Int Conf Learn Represent, 2025

\bibitem{ge2026redimatch}
Ge Y, Zheng G, Guo W,  et~al.
\newblock {REDI-Match}: rotation-equivariant distillation for efficient and
  robust dense matching.
\newblock ArXiv:2606.24330, 2026

\bibitem{sun2025learning}
Sun P, Guan B, Yu Z,  et~al.
\newblock Learning affine correspondences by integrating geometric constraints.
\newblock In: Proc IEEE/CVF Conf Comput Vis Pattern Recognit, 2025.
\newblock 27038--27048

\bibitem{teed2020raft}
Teed Z,  Deng J.
\newblock {RAFT}: recurrent all-pairs field transforms for optical flow.
\newblock In: Proc Eur Conf Comput Vis, 2020.
\newblock 402--419

\bibitem{xu2022gmflow}
Xu H, Zhang J, Cai J,  et~al.
\newblock {GMFlow}: learning optical flow via global matching.
\newblock In: Proc IEEE/CVF Conf Comput Vis Pattern Recognit, 2022.
\newblock 8121--8130

\bibitem{xu2023unimatch}
Xu H, Zhang J, Cai J,  et~al.
\newblock Unifying flow, stereo and depth estimation.
\newblock IEEE Trans Pattern Anal Mach Intell, 2023, 45: 13941--13958

\bibitem{zhao2025ufm}
Zhang Y, Keetha N, Lyu C,  et~al.
\newblock {UFM}: a simple path towards unified dense correspondence with flow.
\newblock In: Adv Neural Inf Process Syst, 2025.
\newblock 121092--121119

\bibitem{zhang2025panmatch}
Zhang Y, Wang L, Li K,  et~al.
\newblock {PanMatch}: unleashing the potential of large vision models for
  unified matching models.
\newblock ArXiv:2507.08400, 2025

\bibitem{liu2026crft}
Liu X, Ding M, Sun Z,  et~al.
\newblock {CRFT}: consistent-recurrent feature flow transformer for cross-modal
  image registration.
\newblock In: Proc IEEE/CVF Conf Comput Vis Pattern Recognit, 2026.
\newblock 34784--34794

\bibitem{deng2025argmatch}
Deng Y, Zhang K, Tang L,  et~al.
\newblock {ArgMatch}: adaptive refinement gathering for efficient dense
  matching.
\newblock In: Proc IEEE/CVF Int Conf Comput Vis, 2025.
\newblock 27369--27379

\bibitem{astermark2025dms}
Astermark J, Heyden A,  Larsson V.
\newblock Dense match summarization for faster two-view estimation.
\newblock In: Proc IEEE/CVF Conf Comput Vis Pattern Recognit, 2025.
\newblock 1093--1102

\bibitem{jiang2021cotr}
Jiang W, Trulls E, Hosang J,  et~al.
\newblock {COTR}: correspondence transformer for matching across images.
\newblock In: Proc IEEE/CVF Int Conf Comput Vis, 2021.
\newblock 6207--6217

\bibitem{tan2022eco}
Tan D, Liu J~J, Chen X,  et~al.
\newblock {ECO-TR}: efficient correspondences finding via coarse-to-fine
  refinement.
\newblock In: Proc Eur Conf Comput Vis, 2022.
\newblock 317--334

\bibitem{goswami2026unicorrn}
Goswami P, Ding T, Liu F,  et~al.
\newblock {UniCorrn}: unified correspondence transformer across {2D} and {3D}.
\newblock In: Proc IEEE/CVF Conf Comput Vis Pattern Recognit, 2026.
\newblock 9943--9954

\bibitem{cao2022iterative}
Cao S~Y, Hu J, Sheng Z,  et~al.
\newblock Iterative deep homography estimation.
\newblock In: Proc IEEE/CVF Conf Comput Vis Pattern Recognit, 2022.
\newblock 1879--1888

\bibitem{nguyen2018unsupervised}
Nguyen T, Chen S~W, Shivakumar S~S,  et~al.
\newblock Unsupervised deep homography: a fast and robust homography estimation
  model.
\newblock IEEE Robot Autom Lett, 2018, 3: 2346--2353

\bibitem{liu2022content}
Liu S, Ye N, Wang C,  et~al.
\newblock Content-aware unsupervised deep homography estimation and its
  extensions.
\newblock IEEE Trans Pattern Anal Mach Intell, 2023, 45: 2849--2863

\bibitem{zhang2025adapting}
Zhang K, Deng Y, Ma J,  et~al.
\newblock Adapting dense matching for homography estimation with grid-based
  acceleration.
\newblock In: Proc IEEE/CVF Conf Comput Vis Pattern Recognit, 2025.
\newblock 6294--6303

\bibitem{li2024dmhomo}
Li H, Jiang H, Luo A,  et~al.
\newblock {DMHomo}: learning homography with diffusion models.
\newblock ACM Trans Graph, 2024, 43: 1--16

\bibitem{jiang2026supervised}
Jiang H, Li H, Han S,  et~al.
\newblock Supervised small-baseline and large-baseline homography learning with
  diffusion-based data generation.
\newblock IEEE Trans Pattern Anal Mach Intell, 2026, 48: 8120--8133

\bibitem{liu2025homogan}
Liu S, Hong M, Lu Y,  et~al.
\newblock Unsupervised global and local homography estimation with
  coplanarity-aware {GAN}.
\newblock IEEE Trans Pattern Anal Mach Intell, 2025, 47: 1863--1876

\bibitem{yu2025sshnet}
Yu J, Cao S~Y, Zhang R,  et~al.
\newblock {SSHNet}: unsupervised cross-modal homography estimation via problem
  reformulation and split optimization.
\newblock In: Proc IEEE/CVF Conf Comput Vis Pattern Recognit, 2025.
\newblock 16685--16694

\bibitem{gao2026homodiff}
Gao Y, Zheng C, Xu J,  et~al.
\newblock {HomoDiff}: noise-robust multi-modal homography estimation via
  diffusion model.
\newblock IEEE Trans Pattern Anal Mach Intell, 2026, doi:
  10.1109/TPAMI.2026.3730469

\bibitem{you2026generalized}
You J, Cheng J, Zhang J,  et~al.
\newblock Towards generalized multimodal homography estimation.
\newblock In: Proc IEEE/CVF Conf Comput Vis Pattern Recognit, 2026.
\newblock 8408--8417

\bibitem{liu2026adverse}
Liu R, Zhang J, Zhang Z,  et~al.
\newblock Toward reliable homography estimation under adverse degradations: an
  optimization-driven approach.
\newblock IEEE Trans Pattern Anal Mach Intell, 2026, doi:
  10.1109/TPAMI.2026.3720245

\bibitem{detone2016deep}
DeTone D, Malisiewicz T,  Rabinovich A.
\newblock Deep image homography estimation.
\newblock ArXiv:1606.03798, 2016

\bibitem{le2020deep}
Le H, Liu F, Zhang S,  et~al.
\newblock Deep homography estimation for dynamic scenes.
\newblock In: Proc IEEE/CVF Conf Comput Vis Pattern Recognit, 2020.
\newblock 7652--7661

\bibitem{cao2023recurrent}
Cao S~Y, Zhang R, Luo L,  et~al.
\newblock Recurrent homography estimation using homography-guided image warping
  and focus transformer.
\newblock In: Proc IEEE/CVF Conf Comput Vis Pattern Recognit, 2023.
\newblock 9833--9842

\bibitem{zhao2021deep}
Zhao Y, Huang X,  Zhang Z.
\newblock Deep {Lucas-Kanade} homography for multimodal image alignment.
\newblock In: Proc IEEE/CVF Conf Comput Vis Pattern Recognit, 2021.
\newblock 15950--15959

\bibitem{zhang2024sparse}
Zhang K,  Ma J.
\newblock Sparse-to-dense multimodal image registration via multi-task
  learning.
\newblock In: Proc Int Conf Mach Learn, 2024.
\newblock 59490--59504

\bibitem{baker2004lucas}
Baker S,  Matthews I.
\newblock {Lucas-Kanade} 20 years on: a unifying framework.
\newblock Int J Comput Vis, 2004, 56: 221--255

\bibitem{liu2022unsupervised}
Liu S, Lu Y, Jiang H,  et~al.
\newblock Unsupervised global and local homography estimation with motion basis
  learning.
\newblock IEEE Trans Pattern Anal Mach Intell, 2023, 45: 7885--7899

\bibitem{zhu2025edffdnet}
Zhu H, Qu B, Cao S~Y,  et~al.
\newblock {EDFFDNet}: towards accurate and efficient unsupervised multi-grid
  image registration.
\newblock In: Proc IEEE/CVF Int Conf Comput Vis, 2025.
\newblock 5102--5111

\bibitem{li2025camflow}
Li H, Zhou T, Yang Z,  et~al.
\newblock Estimating {2D} camera motion with hybrid motion basis.
\newblock In: Proc IEEE/CVF Int Conf Comput Vis, 2025.
\newblock 7624--7633

\bibitem{melekhov2017relative}
Melekhov I, Ylioinas J, Kannala J,  et~al.
\newblock Relative camera pose estimation using convolutional neural networks.
\newblock In: Proc Int Conf Adv Concepts Intell Vis Syst, 2017.
\newblock 675--687

\bibitem{zhou2014learning}
Zhou B, Lapedriza A, Xiao J,  et~al.
\newblock Learning deep features for scene recognition using {Places} database.
\newblock In: Adv Neural Inf Process Syst, 2014.
\newblock 487--495

\bibitem{en2018rpnet}
En S, Lechervy A,  Jurie F.
\newblock {RPNet}: an end-to-end network for relative camera pose estimation.
\newblock In: Proc Eur Conf Comput Vis Workshops, 2018.
\newblock 738--745

\bibitem{chen2021wide}
Chen K, Snavely N,  Makadia A.
\newblock Wide-baseline relative camera pose estimation with directional
  learning.
\newblock In: Proc IEEE/CVF Conf Comput Vis Pattern Recognit, 2021.
\newblock 3258--3268

\bibitem{schonemann1966generalized}
Sch{\"o}nemann P~H.
\newblock A generalized solution of the orthogonal {Procrustes} problem.
\newblock Psychometrika, 1966, 31: 1--10

\bibitem{arnold2022map}
Arnold E, Wynn J, Vicente S,  et~al.
\newblock Map-free visual relocalization: metric pose relative to a single
  image.
\newblock In: Proc Eur Conf Comput Vis, 2022.
\newblock 690--708

\bibitem{yin2025srpose}
Yin R, Zhang Y, Pan Z,  et~al.
\newblock {SRPose}: two-view relative pose estimation with sparse keypoints.
\newblock In: Proc Eur Conf Comput Vis, 2024.
\newblock 88--107

\bibitem{dong2025reloc3r}
Dong S, Wang S, Liu S,  et~al.
\newblock {Reloc3r}: large-scale training of relative camera pose regression
  for generalizable, fast, and accurate visual localization.
\newblock In: Proc IEEE/CVF Conf Comput Vis Pattern Recognit, 2025.
\newblock 16739--16752

\bibitem{loiseau2025alligat0r}
Loiseau T, Bourmaud G,  Lepetit V.
\newblock {Alligat0R}: pre-training through covisibility segmentation for
  relative camera pose regression.
\newblock In: Adv Neural Inf Process Syst, 2025.
\newblock 15404--15431

\bibitem{liu2025equipose}
Liu Y,  Dong Q.
\newblock {EquiPose}: exploiting permutation equivariance for relative camera
  pose estimation.
\newblock In: Proc IEEE/CVF Conf Comput Vis Pattern Recognit, 2025.
\newblock 1127--1137

\bibitem{zhou2019continuity}
Zhou Y, Barnes C, Lu J,  et~al.
\newblock On the continuity of rotation representations in neural networks.
\newblock In: Proc IEEE/CVF Conf Comput Vis Pattern Recognit, 2019.
\newblock 5745--5753

\bibitem{levinson2020analysis}
Levinson J, Esteves C, Chen K,  et~al.
\newblock An analysis of {SVD} for deep rotation estimation.
\newblock In: Adv Neural Inf Process Syst, 2020.
\newblock 22554--22565

\bibitem{cai2021extreme}
Cai R, Hariharan B, Snavely N,  et~al.
\newblock Extreme rotation estimation using dense correlation volumes.
\newblock In: Proc IEEE/CVF Conf Comput Vis Pattern Recognit, 2021.
\newblock 14566--14575

\bibitem{bezalel2025extreme}
Bezalel H, Ankri D, Cai R,  et~al.
\newblock Extreme rotation estimation in the wild.
\newblock In: Proc IEEE/CVF Conf Comput Vis Pattern Recognit, 2025.
\newblock 1061--1070

\bibitem{poursaeed2018deep}
Poursaeed O, Yang G, Prakash A,  et~al.
\newblock Deep fundamental matrix estimation without correspondences.
\newblock In: Proc Eur Conf Comput Vis Workshops, 2018.
\newblock 485--497

\bibitem{zhou2020learn}
Zhou Q, Sattler T, Pollefeys M,  et~al.
\newblock To learn or not to learn: visual localization from essential
  matrices.
\newblock In: Proc IEEE Int Conf Robot Autom, 2020.
\newblock 3319--3326

\bibitem{thomee2016yfcc100m}
Thomee B, Shamma D~A, Friedland G,  et~al.
\newblock {YFCC100M}: the new data in multimedia research.
\newblock Commun ACM, 2016, 59: 64--73

\bibitem{jin2021image}
Jin Y, Mishkin D, Mishchuk A,  et~al.
\newblock Image matching across wide baselines: from paper to practice.
\newblock Int J Comput Vis, 2021, 129: 517--547

\bibitem{brachmann2021limits}
Brachmann E,  Rother C.
\newblock On the limits of pseudo ground truth in visual camera
  re-localisation.
\newblock In: Proc IEEE/CVF Int Conf Comput Vis, 2021.
\newblock 6218--6228

\bibitem{xiao2013sun3d}
Xiao J, Owens A,  Torralba A.
\newblock {SUN3D}: a database of big spaces reconstructed using {SfM} and
  object labels.
\newblock In: Proc IEEE Int Conf Comput Vis, 2013.
\newblock 1625--1632

\bibitem{li2018megadepth}
Li Z,  Snavely N.
\newblock {MegaDepth}: learning single-view depth prediction from internet
  photos.
\newblock In: Proc IEEE Conf Comput Vis Pattern Recognit, 2018.
\newblock 2041--2050

\bibitem{schonberger2016pixelwise}
Sch{\"o}nberger J~L, Zheng E, Frahm J~M,  et~al.
\newblock Pixelwise view selection for unstructured multi-view stereo.
\newblock In: Proc Eur Conf Comput Vis, 2016.
\newblock 501--518

\bibitem{dai2017scannet}
Dai A, Chang A~X, Savva M,  et~al.
\newblock {ScanNet}: richly-annotated {3D} reconstructions of indoor scenes.
\newblock In: Proc IEEE Conf Comput Vis Pattern Recognit, 2017.
\newblock 5828--5839

\bibitem{taira2018inloc}
Taira H, Okutomi M, Sattler T,  et~al.
\newblock {InLoc}: indoor visual localization with dense matching and view
  synthesis.
\newblock In: Proc IEEE Conf Comput Vis Pattern Recognit, 2018.
\newblock 7199--7209

\bibitem{schops2017multi}
Sch{\"o}ps T, Sch{\"o}nberger J~L, Galliani S,  et~al.
\newblock A multi-view stereo benchmark with high-resolution images and
  multi-camera videos.
\newblock In: Proc IEEE Conf Comput Vis Pattern Recognit, 2017.
\newblock 3260--3269

\bibitem{arandjelovic2016netvlad}
Arandjelovi{\'c} R, Gronat P, Torii A,  et~al.
\newblock {NetVLAD}: {CNN} architecture for weakly supervised place
  recognition.
\newblock In: Proc IEEE Conf Comput Vis Pattern Recognit, 2016.
\newblock 5297--5307

\bibitem{liu2024ncmnet}
Liu X, Qin R, Yan J,  et~al.
\newblock {NCMNet}: neighbor consistency mining network for two-view
  correspondence pruning.
\newblock IEEE Trans Pattern Anal Mach Intell, 2024, 46: 11254--11272

\bibitem{shen2022semi}
Shen Z, Sun J, Wang Y,  et~al.
\newblock Semi-dense feature matching with transformers and its applications in
  multiple-view geometry.
\newblock IEEE Trans Pattern Anal Mach Intell, 2023, 45: 7726--7738

\bibitem{shen2024gim}
Shen X, Cai Z, Yin W,  et~al.
\newblock {GIM}: learning generalizable image matcher from internet videos.
\newblock In: Proc Int Conf Learn Represent, 2024

\bibitem{liang2025l2m}
Liang Y, Hu Y, Shao W,  et~al.
\newblock Learning dense feature matching via lifting single {2D} image to {3D}
  space.
\newblock In: Proc IEEE/CVF Int Conf Comput Vis, 2025.
\newblock 6621--6631

\bibitem{yang2026anymatch}
Yang M, Li Z, Tang L,  et~al.
\newblock {AnyMatch}: supercharging universal multi-modal image matching with
  large-scale single-view images.
\newblock In: Proc Eur Conf Comput Vis, 2026.
\newblock 188--206

\bibitem{vuong2025aerialmegadepth}
Vuong K, Ghosh A, Ramanan D,  et~al.
\newblock {AerialMegaDepth}: learning aerial-ground reconstruction and view
  synthesis.
\newblock In: Proc IEEE/CVF Conf Comput Vis Pattern Recognit, 2025.
\newblock 21674--21684

\bibitem{cheng2026airzoo}
Cheng X, Wu R, Liu X,  et~al.
\newblock {AirZoo}: a unified large-scale dataset for grounding aerial
  geometric {3D} vision.
\newblock In: Proc Eur Conf Comput Vis, 2026.
\newblock 656--674

\bibitem{mishkin2015wxbs}
Mishkin D, Matas J, Perdoch M,  et~al.
\newblock {WxBS}: wide baseline stereo generalizations.
\newblock In: Proc Br Mach Vis Conf, 2015.
\newblock 12.1--12.12

\bibitem{sarlin2022lamar}
Sarlin P~E, Dusmanu M, Sch{\"o}nberger J~L,  et~al.
\newblock {LaMAR}: benchmarking localization and mapping for augmented reality.
\newblock In: Proc Eur Conf Comput Vis, 2022.
\newblock 686--704

\bibitem{barath2023homography}
Barath D, Mishkin D, Eichhardt I,  et~al.
\newblock A large-scale homography benchmark.
\newblock In: Proc IEEE/CVF Conf Comput Vis Pattern Recognit, 2023.
\newblock 21360--21370

\bibitem{loiseau2025rubik}
Loiseau T,  Bourmaud G.
\newblock {RUBIK}: a structured benchmark for image matching across geometric
  challenges.
\newblock In: Proc IEEE/CVF Conf Comput Vis Pattern Recognit, 2025.
\newblock 27070--27080

\bibitem{wu2026aerialextrematch}
Wu R, Huang Z, He X,  et~al.
\newblock {AerialExtreMatch}: a benchmark for extreme-view image matching and
  localization.
\newblock IEEE Robot Autom Lett, 2026, 11: 6951--6958

\bibitem{hoang2026semidense}
Hoang K, Nguyen H~T, Ninh H,  et~al.
\newblock Semi-dense matching uncertainty is not just local confidence.
\newblock ArXiv:2608.08685, 2026

\bibitem{chen2022oetr}
Chen Y, Huang D, Xu S,  et~al.
\newblock Guide local feature matching by overlap estimation.
\newblock In: Proc AAAI Conf Artif Intell, 2022.
\newblock 365--373

\bibitem{zhang2026area}
Zhang Y,  Zhao X.
\newblock Searching from area to point: a semantic guided framework with
  geometric consistency for accurate feature matching.
\newblock Pattern Recognit, 2026, 179: 113920

\bibitem{zhang2024mesa}
Zhang Y,  Zhao X.
\newblock {MESA}: matching everything by segmenting anything.
\newblock In: Proc IEEE/CVF Conf Comput Vis Pattern Recognit, 2024.
\newblock 20217--20226

\bibitem{zhang2026mesa}
Zhang Y, Shen S,  Zhao X.
\newblock {MESA}: effective matching redundancy reduction by semantic area
  segmentation.
\newblock IEEE Trans Pattern Anal Mach Intell, 2026, 48: 4454--4472

\bibitem{liu2025sgad}
Liu X, Wang C, Shi G,  et~al.
\newblock {SGAD}: semantic and geometric-aware descriptor for local feature
  matching.
\newblock In: Proc IEEE/CVF Int Conf Comput Vis, 2025.
\newblock 27095--27104

\bibitem{wang2025vggt}
Wang J, Chen M, Karaev N,  et~al.
\newblock {VGGT}: visual geometry grounded transformer.
\newblock In: Proc IEEE/CVF Conf Comput Vis Pattern Recognit, 2025.
\newblock 5294--5306

\bibitem{wang2026pi3}
Wang Y, Zhou J, Zhu H,  et~al.
\newblock $\pi^3$: permutation-equivariant visual geometry learning.
\newblock In: Proc Int Conf Learn Represent, 2026

\bibitem{lin2026depth}
Lin H, Chen S, Liew J~H,  et~al.
\newblock {Depth Anything 3}: recovering the visual space from any views.
\newblock In: Proc Int Conf Learn Represent, 2026

\bibitem{nordstrom2026mum}
Nordstr{\"o}m D, Edstedt J, Kahl F,  et~al.
\newblock {MuM}: multi-view masked image modeling for {3D} vision.
\newblock In: Proc IEEE/CVF Conf Comput Vis Pattern Recognit, 2026.
\newblock 21736--21747

\bibitem{yu2025madpose}
Yu Y, Liu S, Pautrat R,  et~al.
\newblock Relative pose estimation through affine corrections of monocular
  depth priors.
\newblock In: Proc IEEE/CVF Conf Comput Vis Pattern Recognit, 2025.
\newblock 16706--16716

\bibitem{ding2025reposed}
Ding Y, Kocur V, V{\'a}vra V,  et~al.
\newblock {RePoseD}: efficient relative pose estimation with known depth
  information.
\newblock In: Proc IEEE/CVF Int Conf Comput Vis, 2025.
\newblock 14876--14886

\bibitem{weinzaepfel2023croco}
Weinzaepfel P, Lucas T, Leroy V,  et~al.
\newblock {CroCo} v2: improved cross-view completion pre-training for stereo
  matching and optical flow.
\newblock In: Proc IEEE/CVF Int Conf Comput Vis, 2023.
\newblock 17969--17980

\bibitem{an2025zeroco}
An H, Kim J~H, Park S,  et~al.
\newblock Cross-view completion models are zero-shot correspondence estimators.
\newblock In: Proc IEEE/CVF Conf Comput Vis Pattern Recognit, 2025.
\newblock 1103--1115

\bibitem{gallego2020event}
Gallego G, Delbr{\"u}ck T, Orchard G,  et~al.
\newblock Event-based vision: a survey.
\newblock IEEE Trans Pattern Anal Mach Intell, 2022, 44: 154--180

\bibitem{zhang2021image}
Zhang H, Xu H, Tian X,  et~al.
\newblock Image fusion meets deep learning: a survey and perspective.
\newblock Inf Fusion, 2021, 76: 323--336

\bibitem{li2025deep}
Li L, Han L, Ye Y,  et~al.
\newblock Deep learning in remote sensing image matching: a survey.
\newblock ISPRS J Photogramm Remote Sens, 2025, 225: 88--112

\bibitem{jiang2021review}
Jiang X, Ma J, Xiao G,  et~al.
\newblock A review of multimodal image matching: methods and applications.
\newblock Inf Fusion, 2021, 73: 22--71

\bibitem{qiu2024spatiotemporal}
Qiu X, Zhu D~Y, Lu Y,  et~al.
\newblock Spatiotemporal modeling of molecular holograms.
\newblock Cell, 2024, 187: 7351--7373

\bibitem{ren2025minima}
Ren J, Jiang X, Li Z,  et~al.
\newblock {MINIMA}: modality invariant image matching.
\newblock In: Proc IEEE/CVF Conf Comput Vis Pattern Recognit, 2025.
\newblock 23059--23068

\bibitem{frolov2026pym}
Frolov A,  Rodehorst V.
\newblock Patch your matcher: correspondence-aware image-to-image translation
  unlocks cross-modal matching via single-modality priors.
\newblock In: Proc IEEE/CVF Winter Conf Appl Comput Vis, 2026.
\newblock 7913--7924

\bibitem{shrivastava2025selfsupervised}
Shrivastava A,  Owens A.
\newblock Self-supervised spatial correspondence across modalities.
\newblock In: Proc IEEE/CVF Conf Comput Vis Pattern Recognit, 2025.
\newblock 6383--6393

\bibitem{wang2024dgc}
Wang S, Kannala J,  Barath D.
\newblock {DGC-GNN}: leveraging geometry and color cues for visual
  descriptor-free {2D-3D} matching.
\newblock In: Proc IEEE/CVF Conf Comput Vis Pattern Recognit, 2024.
\newblock 20881--20891

\end{thebibliography}

\end{document}